\documentclass[11pt,a4paper]{report}

\usepackage{graphicx,float,subfig,multirow,setspace,parskip,verbatim,slashbox,array,rotating}
\usepackage[numbers]{natbib}
\usepackage[table]{xcolor}
\usepackage[centertags]{amsmath}
\usepackage{amssymb}
\usepackage{titlesec}
\usepackage{colortbl}
\usepackage{listings}
\usepackage{adjustbox}
\usepackage{epsfig}
\usepackage[]{algorithm2e}
\usepackage{hyperref}
\hypersetup{
    colorlinks=true,
    linkcolor=blue,
    filecolor=magenta,      
    urlcolor=cyan,
}
\restylefloat{figure}
\definecolor{kugray5}{RGB}{224,224,224}

\newcommand{\R}{{\cal R}}

\newcommand{\SKIP}[1]{} 

\newcommand{\mbegin} {\left [ \begin{array}}

\newcommand{\mend}   {\end{array} \right ]}

\newcommand{\vbegin} {\left ( \begin{array}{c}}

\newcommand{\vend} {\end{array}\right )}

\newcommand{\bm}[1]{\mbox{\boldmath $#1$}}
\def\squareforqed{\hbox{\rlap{$\sqcap$}$\sqcup$}}

\def\qed{\ifmmode\squareforqed\else{\unskip\nobreak\hfil
	\penalty50\hskip1em\null\nobreak\hfil\squareforqed
	\parfillskip=0pt\finalhyphendemerits=0\endgraf}\fi}

\def\vec#1{\mathchoice%
	{\mbox{\boldmath $\displaystyle\bf#1$}}
	{\mbox{\boldmath $\textstyle\bf#1$}}
	{\mbox{\boldmath $\scriptstyle\bf#1$}}
	{\mbox{\boldmath $\scriptscriptstyle\bf#1$}}}
\def\v#1{\protect\vec #1}

\newcommand{\showeqnlabel}{
	\hbox to 0pt{\quad\quad\relax\fbox{\scriptsize\rm\eqnlblx}%
	\gdef\eqnlblx{xxxx}}} \newcommand{\eqnlblx}{}

\def\@eqnnum{\rm (\theequation)\showeqnlabel}

\newcommand{\nofig}[1]{\centerline{\bf Figure here}}

\newcommand{\impK}{8.5}
\newcommand{\impB}{6.5}

\newcommand{\beq}{\begin{equation}}
\newcommand{\eeq}{\end{equation}} 
\newcommand{\bea}{\begin{eqnarray}}
\newcommand{\eea}{\end{eqnarray}}

\DeclareMathOperator*{\argmax}{arg\,max}

\newcommand{\lam}{{\lambda}}

\newcommand{\calD}{{\cal D}}

\newcommand{\vvcvpr}{{\bf v}} \newcommand{\lb}{{\langle}}
\newcommand{\rb}{{\rangle}} 

\def\boldf#1{\hbox{\rlap{$#1$}\kern.4pt{$#1$}}}

 \newcommand{\trans}{^{\scriptscriptstyle
\top}}

\newcommand{\x}{{\bf x}} 
\newcommand{\w}{{\bf w}}

\titleformat{\chapter}[display]
  {\normalfont\Large\itshape}
  {\chaptertitlename\ \thechapter}{0pt}
  {\normalfont\huge\bfseries}
\titlespacing*{\chapter}{0pt}{10pt}{20pt}

\titleformat{\part}[display]{}{} {3em} {\sffamily\huge}

\newcommand{\qsection}[1]{\vspace{5pt} \noindent \textbf{#1:}}

\newcommand{\ignore}[2]{\hspace{0in}#2}

\begin{document}
\addcontentsline{toc}{chapter}{\scshape{Title}}
\begin{titlepage}
\pagenumbering{arabic}
\begin{center}

\begin{tabular}{cp{1cm}c}
{\includegraphics[height=2cm]{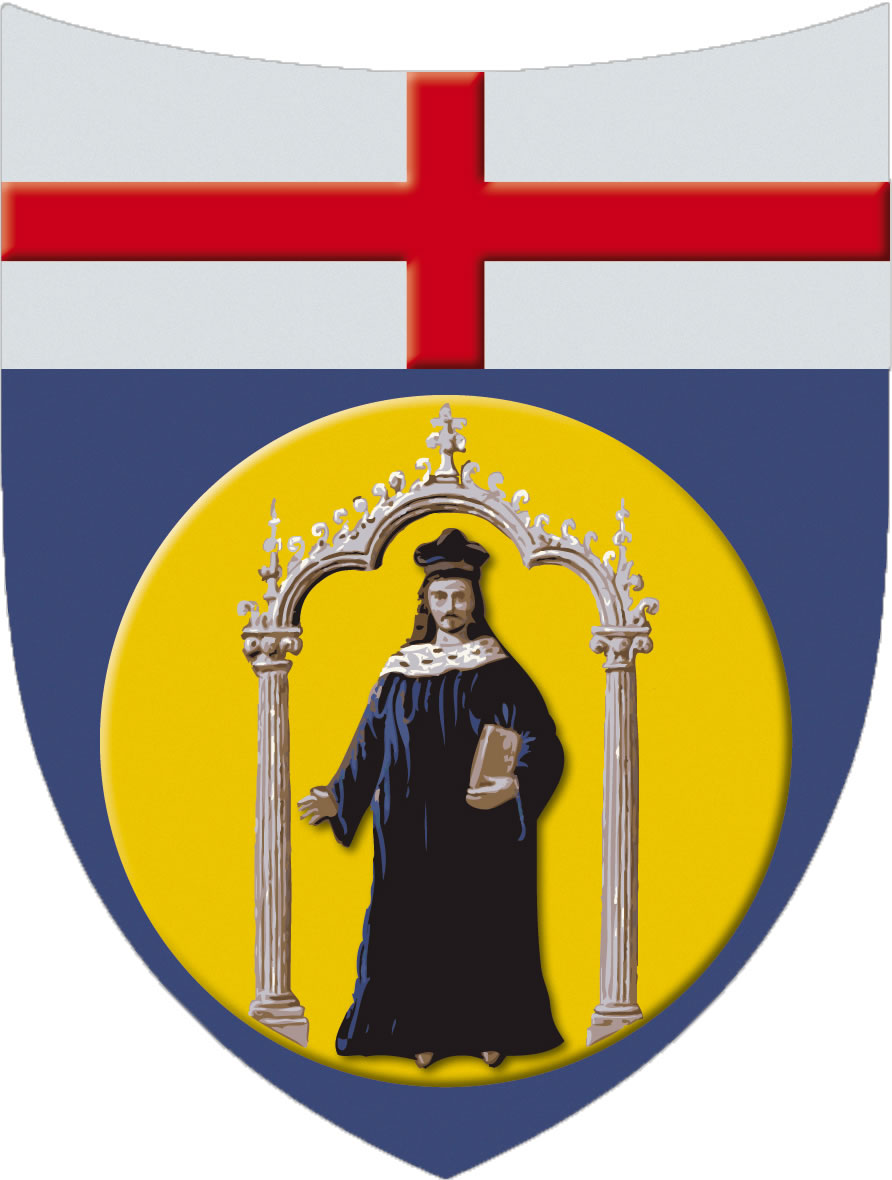}} && {\includegraphics[height=2cm]{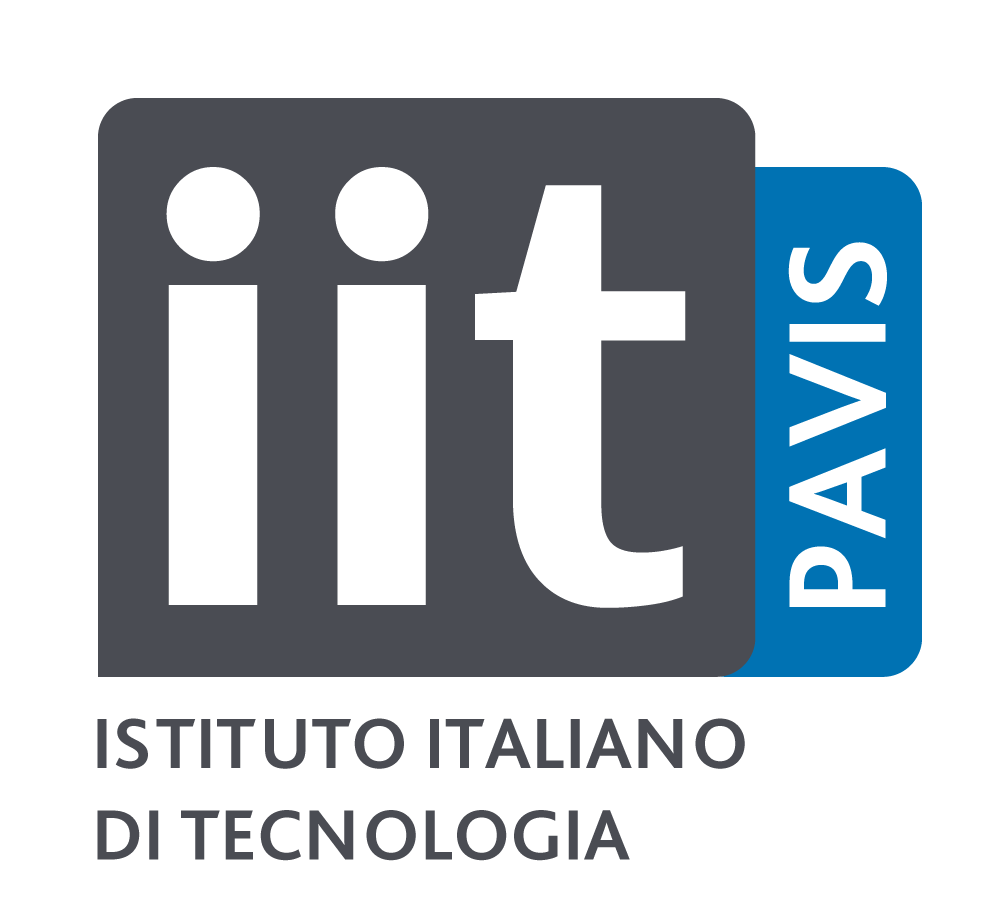}} \\
&&\\
\rmfamily{Universit\`a degli Studi di Genova} && \rmfamily{Istituto Italiano di Tecnologia}\\[3cm]
\end{tabular}
Doctoral Thesis\\[0.8cm] 

\begin{center}
	\huge{\textsc{Mid-level Representation\\
	for Visual Recognition}}\\[0.2cm] 
	{\LARGE Lessons from image and video understanding}\\
	\vspace{4cm}
\end{center}

\begin{center} 
\begin{tabular*}{\textwidth}{@{}l@{\extracolsep{\fill}}r@{}}
\emph{Author:}  & \emph{Supervisor:} \\
\textbf{Moin Nabi}   & \textbf{Prof. Vittorio Murino} \\[0.3cm]
\end{tabular*}
\end{center}
\vspace{2cm}

\large \textit{Submitted in fulfilment of the requirements\\ for the degree of Doctor of Philosophy in Computer Vision}\\[1cm] 
\rmfamily{\large Doctoral School on ``\textit{Life and Humanoid Technologies}"}\\[0.2cm]
\rmfamily{\large Doctoral Course on ``\textit{Nanosciences}" - ``\textit{Computer Vision}"} \\[1cm]
 
{\large {March 2015 - XXVII Cycle}}\\ 
\vfill
\end{center}
\end{titlepage}
\clearpage
\renewcommand{\baselinestretch}{2} 

\newpage
\thispagestyle{empty}
\mbox{}

\vspace{20cm}

\newpage
\thispagestyle{empty}
\mbox{}


\vspace{5cm}
\begin{center}
To my parents and teachers.\\
\emph{(without whom nothing would have been possible)}
\end{center}


\addcontentsline{toc}{chapter}{\scshape{Abstract}}
\begin{abstract}

Visual Recognition is one of the fundamental challenges in AI, where the goal is to understand the semantics of visual data. Employing mid-level representation, in particular, shifted the paradigm in visual recognition. The mid-level image/video representation involves discovering and training a set of mid-level visual patterns (e.g., parts and attributes) and represent a given image/video utilizing them. The mid-level patterns can be extracted from images and videos using the motion and appearance information of visual phenomenas. 

This thesis targets employing mid-level representations for different high-level visual recognition tasks, namely (i) \emph{image understanding} and (ii) \emph{video understanding}.


In the case of image understanding, we focus on object detection/recognition task. We investigate on discovering and learning a set of mid-level patches to be used for representing the images of an object category. We specifically employ the discriminative patches in a subcategory-aware webly-supervised fashion. We, additionally, study the outcomes provided by employing the subcategory-based models for undoing dataset bias.

In the case of video understanding, we first study acquiring a mid-level motion-based representation (i.e., tracklet) to capture the commotion of a crowd motion for the task of abnormality detection in crowd. Next, we study exploiting dynamics of a dictionary of appearance-based models (i.e. Poselets) as human motion representation for activity recognition.

Our empirical studies show that employing richer mid-level representations can provide significant benefits for visual recognition (both in image and video understanding). Our main contributions are as follows: (a) introducing a method for undoing dataset bias using subcategory-based models, (b) discovering and training webly-supervised subcategory-aware discriminative patches for object recognition, (c) proposing tracklet-based commotion measure for abnormality detection in crowd, (d) introducing Temporal Poselet a descriptor for group activity recognition.

\ignore{Mid-level represeantions have become an essential tool in visual recognition. Much of the progress in computer vision over the past decade has built on simple, local representations such as SIFT or HOG. Part-based models in particular shifted the paradigm in mid-level representation. Subsequent works have often focused on improving either computational efficiency, or invariance properties. This thesis belongs to the latter group. Invariance is a particularly relevant aspect if we intend to work with dense features. The traditional approach to sparse matching is to rely on stable interest points, such as corners, where scale and orientation can be reliably estimated, enforcing invariance; dense features need to be computed on arbitrary points. Dense features have been shown to outperform sparse matching techniques in many recognition problems, and form the bulk of our work. In this thesis we present strategies to enhance low-level, local features with mid-level, global cues. We devise techniques to construct better features, and use them to handle complex ambiguities, occlusions and background changes. To deal with ambiguities, we explore the use of motion to enforce temporal consistency with optical flow priors. We also introduce a novel technique to exploit segmentation cues, and use it to extract features invariant to background variability. For this, we downplay image measurements most likely to belong to a region different from that where the descriptor is computed. In both cases we follow the same strategy: we incorporate mid-level, "big picture" information into the construction of local features, and proceed to use them in the same manner as we would the baseline features. We apply these techniques to different feature representations, including SIFT and HOG, and use them to address canonical vision problems such as stereo and object detection, demonstrating that the introduction of global cues yields consistent improvements. We prioritize solutions that are simple, general, and efficient. Our main contributions are as follows: (a) An approach to dense stereo reconstruction with spatiotemporal features, which unlike existing works remains applicable to wide baselines. (b) A technique to exploit segmentation cues to construct dense descriptors invariant to background variability, such as occlusions or background motion. (c) A technique to integrate bottom-up segmentation with recognition efficiently, amenable to sliding window detectors.}

\end{abstract}

\newpage
\thispagestyle{empty}
\mbox{}


\section*{Acknowledgments}

First of all, I would like to thank my family~(Malih, Nasrollah, Mahsa, Meraj, Mostaan); it is because of their never ending support that I have had the chance to progress in life. Their dedication to my education provided the foundation for my studies.

I started working on computer vision at Institute for Research in Fundamental Sciences (IPM). For this, I am grateful to Professor Mehrdad Shahshahani; for the past years he has had a critical role in conducting computer vision researches at Iran. With no doubt, without his support and guidance, I could not come this far. He inspired me by the way he looks at research and life.

I would like to express my deepest gratitude to my advisor, Professor Vittorio Murino for providing me with an excellent environment for conducting research. Professor Murino has always amazed me by his intuitive way of thinking and supervisory skills. He has been a great support for me and a nice person to talk to.

I am especially indebted to Dr. Ali Farhadi who has been a great mentor for me during my visit to the GRAIL lab at the University of Washington. I am grateful for his support, advice, and friendship. I also express my sincere gratitude to Dr. Alessandro Perina, Dr. Santosh Divvala and Professor Massimiliano Pontil for sharing their experiences and supporting me with their advices.


I have compiled a list of authors I have cited the most in this thesis. According to that list, the work presented here has been mostly inspired by (in alphabetical order): Alyosha Efros, Jitendra Malik, Deva Ramanan, and Andrew Zisserman.

Last but not least, I'd like to acknowledge funders who supported my PhD studies. The works in this thesis were funded by Italian Institute of Technology (IIT).

\newpage
\thispagestyle{empty}
\mbox{}

\tableofcontents

\listoffigures	


\listoftables  

\newpage
\thispagestyle{empty}
\mbox{}
\chapter{Introduction}

Visual recognition, as the core component of Artificial Intelligence systems, is to process, perceive and interpret the semantic of visual contents such as images and videos. Visual [re]cognition for human beings, thanks to a billion years of evolution, is so trivial, thus human vision is exceptionally powerful even without any conscious effort. Visual recognition for a computer, however, is extremely complex, insofar as it can be viewed as the key hinder factor of computer vision research. To give this ability to computers, many researchers from neuroscientists and biologists to psychologists, engineers and mathematicians work closely together. Recent advancements improved visual recognition with lots of astonishing abilities, as a result of employing rich mathematical models and the inspiration from the nature. Visual recognition consists of many tasks in both images and video level, including object category detection and recognition, action and activity recognition, scene understanding and motion-based crowd analysis. Broadly speaking, visual recognition is:
\begin{center}
				\emph{`` pattern recognition techniques applied to visual patterns.}''
\end{center}
\vspace{.5cm}

Visual recognition became so popular for both academic purposes and industrial AI. This growing trend is mainly due to increasing collections of visual data as well as processing power. The former is a consequence of Internet and the latter by employing high-performance computing. By the help of these two, industry is taking serious steps toward including the visual recognition in the next generation of AI systems.

\vspace{-2.5mm}

\section{Motivations}

During the last decades, using local interest points (such as SIFT\cite{lowe1999object}) has proven to noticeably tackle visual recognition. However, there has been recently a growing number of approaches suggesting that using only very local low-level features may be insufficient, and that improvement can be achieved through the use of \emph{mid-level representations} capturing higher-level concepts. There is always a vast semantic gap between low-level image pixels and high-level semantics in images or videos. The mid-level representations gained much attention specifically because they can fill-in this gap by shifting the goal of recognition from naming to describing. Going beyond low-level representations, allows us to describe, compare, and more easily recognize visual concepts. 

Besides that, visual recognition systems based on low-level representations usually require a lot of manually labeled training data, typically several thousands of examples. Employing mid-level representations, however, can reduce the number of necessary training examples by serving an intermediate layer in a recognition cascade. This enables the system to recognize visual concepts, for which it had limited numbers of seen training examples (in order of hundreds or less).

Using mid-level representation for visual recognition is very well-storied in image and video understanding, but recently due to the growing trend of representation learning (e.g. deep learning) for visual recognition, investigating the capabilities of the \emph{mid-level image and video representation} became a hot topic again. Methodologically speaking, the mid-level representation for visual recognition covers a wide spectrum of frameworks in computer vision including part-based models \cite{felzenszwalb2013visual}, attribute-based representations \cite{farhadi2009describing,lampert2009learning} and hierarchical mid-level representations (e.g. Convolutional Neural Nets \cite{krizhevsky2012imagenet}).

Pursuing the goal of approaching to a more semantic visual recognition, this thesis is devoted to study mid-level representation in visual recognition. In this thesis, we seek computational techniques for the recognition (as oppose to \emph{biological approaches}) of visual patterns. Specifically, we address visual recognition by using computational models capturing the statistics of motion and appearance. To show the capabilities gained by employing mid-level representations, we peculiarly focus on three important visual recognition high-level tasks and study utilizing mid-level image/video representations for them. The tasks are: \emph{object detection, human activity recognition, and motion-based crowd analysis.}

\begin{figure*}
\begin{center}
\includegraphics[width=14cm]{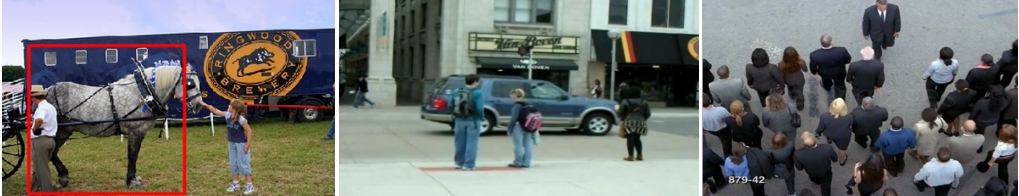} \\
\end{center}
\hspace{2cm} (a) \hspace{4cm} (b) \hspace{4cm} (c)
   \caption{Visual Recognition tasks: (a) object detection, (b) activity recognition, (c) crowd analysis.}
\label{fig:intro}
\end{figure*}

In the following, we introduce briefly each task:

\qsection{Object detection} The goal of object detection is to localize all instances of specific object category in a given image. Typically only a small number of instances of the object is present in the image, but there is a very large number of possible locations and scales at which they can occur and that need to be explored \cite{amit2014object}. Computational algorithms of object detection usually build a statistical model for object categories given a set of training examples and prior knowledge (i.e., human supervision). Such prior knowledge is provided as annotation for images in terms of bounding boxes around object instances or the parts (see Fig.~\ref{fig:intro}a).

\qsection{Human activity recognition} The goal of this task is to recognize the activity of one (or more) people acting in a video sequence. Activity recognition can be done in different settings: action of individuals, human interactions and group activities (see Fig.~\ref{fig:intro}b). This task has been formalized in the literature as classification problem where a label corresponding to an activity has to be assigned to a specific video frame or a video clip, possibly also identifying the spatial location in the video sequence where such activities occur (i.e., activity detection).

\qsection{Crowd behavior analysis} The goal of crowd behavior analysis is to detect abnormal behavior in crowded scenes based on the crowd \emph{motion characteristics}. The main reason for using the motion information for this task is the low-quality of videos captured by surveillance camera and the real-time nature of this task, where high-level appearance-based models normally fail. Abnormal behaviors can be localized in three different levels: \emph{pixel-level}, \emph{frame-level} and \emph{video-level} (see Fig.~\ref{fig:intro}c).\\

In this thesis, we address each of these tasks of visual recognition in the context of mid-level representation. The methods discussed in this dissertation are based on some new components developed for these tasks and some existing state-of-the-art components that are extended or adapted to fit in the system. 

\vspace{-2.5mm}

\section{Main Contributions}

To form a comprehensive analysis, we study different aspects of mid-level representations each evaluated in an aforementioned tasks of visual recognition. 

For a methodological investigation, we organize this thesis in two parts: \emph{(i) mid-level representation for image understanding}, and \emph{(ii) mid-level representation for video understanding.}

\begin{itemize}
\item  In part (i), we focus on employing mid-level image representation for object recognition problem. After an extensive systematic review on subcategory-based models, we study the outcomes provided by employing this model for the task of object recognition. We first investigate on \emph{undoing dataset bias} and introduce a subcategory-based approach to tackle this problem. We, next, study discovering and learning discriminative patch models to build a mid-level image representations. For this purpose, we have introduced a fully-automated system that learns visual concepts by processing lots of books and images on the Internet (i.e., webly-supervised statistical model). We evaluated the proposed method for object recognition and detection.
\item In part (ii), we explore the significance of utilizing mid-level techniques for motion representation. For this purpose, we first focus on pure motion statistics, then complement it by appearance information in order to build a richer representation. We specifically introduce a tracklet-based representation to model commotion in crowd and a detector-based motion representation for group activity detection and recognition in the wild.
\end{itemize}
\vspace{.7cm}
The contributions are well described in Chapters \ref{chap:image} and \ref{chap:video}, but the main contributions of this thesis is shortened below with a brief description.

\begin{enumerate}
\item \qsection{The subcategory-based undoing dataset bias} We proposed a subcategory-based framework for tackling database bias problem, combining new ideas from multitask learning. Our method leverages the availability of multiple biased datasets to tackle a common classification task in a principled way. 
\item \qsection{ The Webly-supervised subcategory-aware discriminative patch}  We introduced a subcategory-aware method for discovering and training the webly-supervised discriminative patches. We aim at mining these patch models in a discriminative way to form a richer representation for visual object recognition.

\item \qsection{The Commotion Measure for abnormality detection in crowd}
We employed the tracklet as an intermediate-level motion representation, and proposed a new unsupervised measure to evaluate the commotion of a crowd scene in pixel, frame and video levels.

\item \qsection{ The Temporal Poselet as a descriptor for group activity recognition} We introduced a novel semantic descriptor for human activity by analyzing the activation correlation of a bank of patch models (e.g., poselet \cite{bourdev2009poselets}) over time. This provides a video representation composed by joint human detections and activity characterization using the basis of the detection activation patterns of the poselets.

\end{enumerate}
\vspace{-2.5mm}

\subsection{PhD Publications}

\begin{itemize}
\item "Learning with Dataset Bias in Latent Subcategory Models", D. Stamos, S. Martelli, M. Nabi, A. McDonald, V. Murino, M. Pontil, International Conference on Computer Vision and Pattern Recognition (CVPR), 2015.~\cite{Stamos_2015_CVPR}

\item "Temporal Poselets for Collective Activity Detection and Recognition", M. Nabi, A. Del Bue and V. Murino. IEEE International Conference on Computer Vision Workshops (ICCVW 2013), Sydney, Australia.~\cite{6755938}

\item "Crowd Motion Monitoring Using Tracklet-based Commotion Measure", M. Nabi, H. Mousavi, H. Kiani, A. Perina and V. Murino, International Conference on Image Processing (ICIP), 2015.~\cite{mousavi2015crowd}

\item "Abnormality Detection with Improved Histogram of Oriented Tracklets", H. Mousavi, M. Nabi, H. Kiani, A. Perina and V. Murino, International Conference on Image Analysis and Processing (ICIAP), 2015.~\cite{mousavi2015abnormality}

\end{itemize}

\vspace{-2.5mm}
\section{Thesis Overview}

The rest of this thesis is organized as follows.

In Chapter \ref{chap:back}, we briefly review the recent successful architectures for visual recognition. We also present an extended analysis of unsupervised visual subcategory as a key component of state-of-the-art visual recognition models. We note that the previous works related to each visual recognition task (e.g., object recognition, crowd analysis and activity recognition) is extensively described in the corresponding chapters (\ref{chap:image} and \ref{chap:video}).

In Chapter \ref{chap:image}, we first study subcategory-based models and address dataset bias problem by employing them. Then, we investigate utilizing patch-based models for mid-level image representations. We specifically introduce an approach to discover and learn subcategory-aware discriminative patch in a webly-supervised fashion.

In Chapter \ref{chap:video}, we first present a motion-based video representation, employing tracklet as an intermediate-level representation. A novel measure is also introduced to capture the commotion of a crowd motion for the task of abnormality detection in crowd. Next, we introduce an appearance-based video representation (Temporal-Poselets), exploiting the dynamics of a given dictionary of patch-based models (Poselets) in time for detecting/recognizing the activity of a group of people in crowd environments.

Both chapter \ref{chap:image} and chapter \ref{chap:video} are supported with experimental results of different visual recognition tasks on several data sets.

Finally, Chapter \ref{chap:con} draws some concluding remarks and summarizes the most valuable results obtained in this work, as well as future perspectives.

\newpage
\thispagestyle{empty}
\mbox{}
\chapter{Concepts and Backgrounds} \label{chap:back}

\section{Backgrounds on Visual Recognition}
In this section we briefly explain a set of well-known architectures which extensively have been used for visual recognition (especially for object detection). All of these frameworks will be exploited in the next following chapters. 

\subsection{HOG \& SVM Framework} \label{sec:hogsvm} 
Histogram of Oriented Gradients (HOG) \cite{dalal2005histograms} is a descriptor which is built on the gradient information of a given image by counting the local orientations. This description is so powerful, specifically because it can handle little noise and small variations (due to illumination and viewpoint) by local quantization, while keeping the global structure of the image. To extract HOG, we simply find the number of gradient in various orientations in each 8x8 neighborhood, and then compute the histogram of them locally. The final HOG feature is created by stacking all local features to create a global feature map. 

Now, For training a model for a particular object category (e.g. pedestrian), we extract HOG from the positive bounding boxes (i.e., belong to images of pedestrian category) as well as negative ones (i.e., extracted randomly from background images). Finally we feed all the positive and negative feature vectors to train SVM classifier (see Figure \ref{fig:dpm}(a)). 
In detection phase, we will sweep the model over the image in a sliding window manner. For each hypothesis bounding box, we compute the inner product between parameter vector and HOG of that box. The final detect results are the bounding boxes where the product is higher than a threshold.

More formally, consider a classification problem where we observe a dataset of $n$ labeled examples $D = (<x_1,y_1>,...,<x_n,x_n>)$ with $y_i \in \{-1,1\}$. For standard binary classification problem, a commonly used approach is to minimize the trade-off between the $l_2$ regularization term and the hinge loss on the training data:
\begin{align}
\arg \min_{w} \frac{1}{2} \parallel\omega\parallel^2 +~C\sum_{i=1}^{n} \epsilon_i, \label{eq:svm}\\
y_i \cdot s_i > 1 - \epsilon_i,~\epsilon_i > 0,\\
s_i = \omega \cdot \phi(x_i)+b.
\end{align}
The parameter $C$ controls the relative weight of the hinge-loss term, $\omega\in$$R^D$ is the vector of model parameters and $\phi(.)\in$$R^D$ denotes the feature representation for sample $x_i$ (Histogram of Oriented Gradient).

\subsection{Deformable Part-based Model (DPMs)} \label{sec:dpm}

\begin{figure}[t]
\begin{center}
\includegraphics[width=6.0cm]{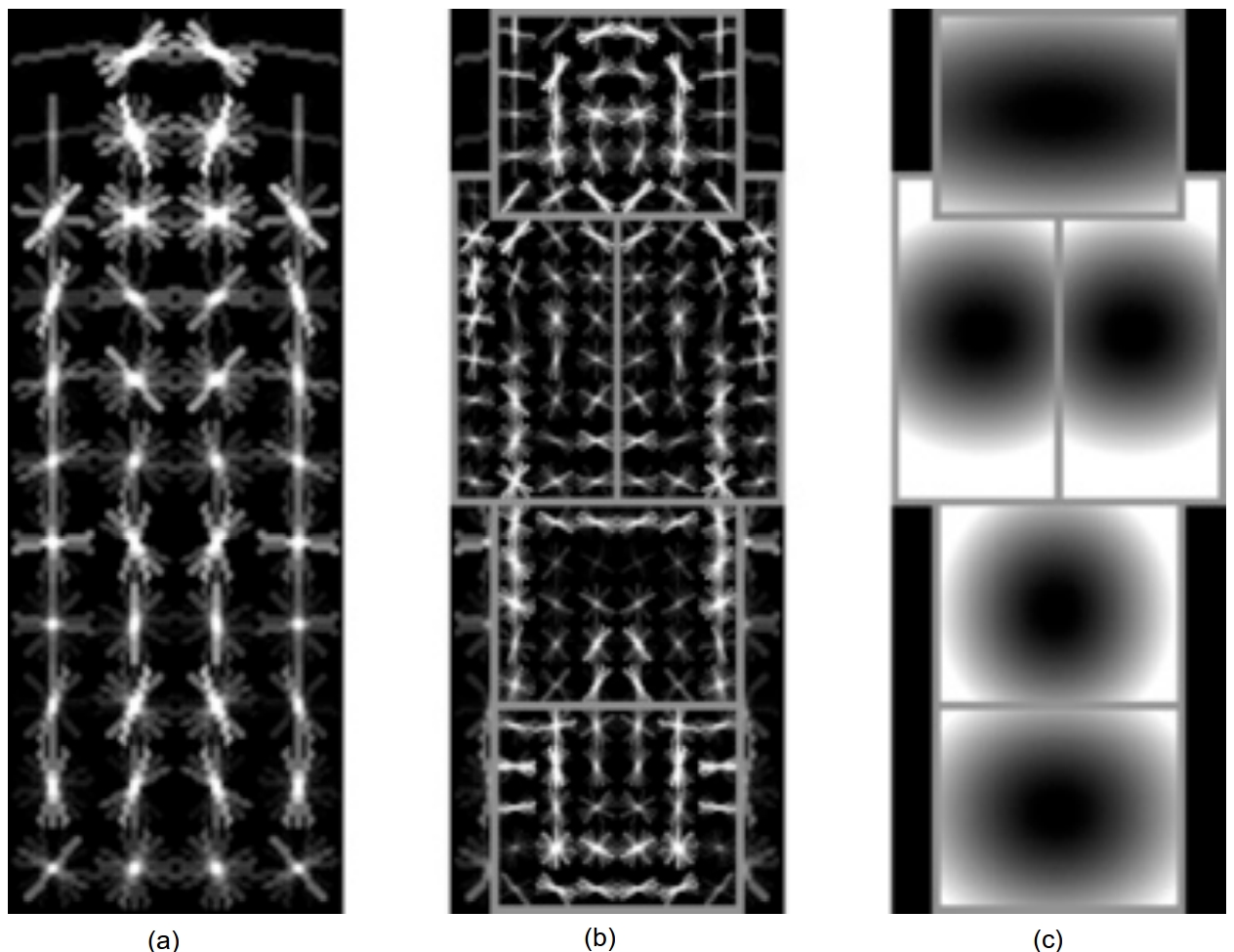} \\
\end{center}
\caption{The part-based feature extraction process in \cite{felzenszwalb2010object}. (a) The holistic feature $\varphi_0(x)$ extracted from the whole sliding window $x$ ($x$ here represents a standing up human body). (b) $P$ ($P$ = 5) different parts extracted from corresponding patches in a fixed configuration. The $j$-th part corresponds to $\varphi_0(x_j)$, where $p_j$ is the offset from the top left corner of $x$. (c) A representation of the deformation parameters $\beta$ (see the text for details). In few words, each part is associated with specific deformation parameters included in $\beta$. These parameters are used at testing time in order to penalize parts localized far from the expected configuration. In (c), the darker is the pixel, the less is the penalization.}
\label{fig:dpm}
\end{figure}

In this section, we will have a deep look at the Deformable Part Model (DPM) paradigm introduced by Felzenszwalb {\em et al.} \cite{felzenszwalb2010object}, which is one of the most successful and widely used state-of-the-art object detection approaches. One rational solution to deal with the high visual diversity in the object  categories is to take the HOG \& SVM model, cut it up to little pieces and now let the little pieces move around. Different arrangement of such pieces can model diverse appearance of an object category (see Figure \ref{fig:dpm}(b)).
It can be seen as a way to enumerate a large set of global templates in a   small set of local tokens (parts). This general idea of modeling an object category using local appearances is so well-storied in the computer vision literature. It was first introduced in \cite{fischler1973representation} as \emph{pictorial structures} and then revised in \cite{felzenszwalb2005pictorial}.

Deformable Part-based Models (DPMs) \cite{felzenszwalb2010object} are collection of local templates, each correspond to an object part. 
Here we define the model mathematically in inference time in terms of a scoring function:
\begin{align}
score(x,z) = \sum_{i}^{n} \omega_i~.~\phi(x,z_i) + \sum_{i,j\in E} \omega_{ij}~.~\varPsi(z_i,z_j) \label{eq:dpm}\\
\varPsi(z_i,z_j) = [dx ~~ dx^2 ~~ dy ~~ dy^2]
\end{align}

For a given image $x$, and a configuration of $n$ parts parametrized with pixel location of each part in x,y coordinate: $z_i = (x_i,y_i)$, we define a score associated to that configuration. The score function in equation \ref{eq:dpm} has two terms: 
\begin{itemize}
\item \qsection{Part template score} This term takes into account the appearance of the parts and it is simply formulated as the sum over the score of all detected parts. For each part we extract HOG from the particular location on the image and computed the inner product of it with the template tuned for that part. (first term in scoring function \ref{eq:dpm})

\item \qsection{Spring deformation score} The second term models the consistency in the overall locations of parts. It is modeled by looking at each pair of parts that are connected by a spring and score the relative position (i.e., deformation) of them. This scoring is done simply by a quadratic function. In practice, we look at the difference of locations in x and y coordinate, and compute the score by dot product of these four numbers   with four trained deformation parameters. Intuitively, it can be seen as a mass spring system because of the quadratic cost. The deformation parameters which define the maximum of the quadratic function can be interpreted as the rest position of the springs. It also defines shape of the quadratic ball which is the rigidity of the spring and shows how quickly the cost increases when parts pull away (see Figure \ref{fig:dpm}(c)).
\end{itemize}
A deeper look at the Deformable Part Model (DPM) paradigm reveals that different ingredients contribute to make the detector impressively robust (especially for articulated object classes such as the human body). In order to proceed with a critical analysis of this system, we introduce The ingredients of the DPM detection approach:

\textbf{\emph{1- Histogram of Oriented Gradient (HOG):}} The first ingredient is the use of HOG features \cite{dalal2005histograms}, which have been extensively proved to be accurate object’s visual appearance descriptors together with linear SVMs (as explained in section \ref{sec:hogsvm}).

\textbf{\emph{2- Deformable Parts:}} The second ingredient is the use of parts. Contrary to \cite{dalal2005histograms}, in fact, in \cite{felzenszwalb2010object} the content of a sliding window is not described using a single holistic HOG feature, extracted from the whole window, but it is based on a combination of different parts. More specifically, if $x$ is the current sliding window and $l = (p_1 , ..., p_P )$ is the configuration of the object parts (where $p_j$ is a point indicating the relative position of the $j$-th part in $x$), then $x$ is represented by a vector $\phi(x)$ obtained by the concatenation the $P + 1$ HOG-like descriptors extracted from different rectangular patches of $x$:
\begin{align}
\phi(x_i) = (\varphi_0(x)^T;\varphi(x,p_1)^T;...;\varphi(x,p_P)^T).
\end{align}

In (2.4), $\varphi_0(x)$ is an holistic feature, i.e., it is extracted from the whole window $x$ and it is equivalent to the representation used in \cite{dalal2005histograms} (see Figure \ref{fig:dpm}(a)). Conversely, $\varphi(x,p_j)$ is a HOG feature extracted from a rectangular patch whose top left corner is $p_j$ (Figure \ref{fig:dpm}(b)). Width and height values of all the parts are fixed in the heuristic initialization step, while for simplicity we skip here the details concerning the feature resolution. Moreover, a DPM includes a vector of deformation parameters $\beta$ which defines the cost function associated to possible spatial perturbations of the points in $l$ at testing time (Figure \ref{fig:dpm} (c)).

\label{sec:LSM} \textbf{\emph{3- Latent Subcategory Model (LSM)}}: Finally, the third ingredient are the components or sub-categories. In order to deal with large variations in the appearance of a given object category (e.g., a pedestrian), the positive training samples are partitioned (clustered) in subcategories and a separate linear SVM is trained for each component. The membership of sample $x$ to the $k$-th component, as well as the deformation parameter vector $\beta$ are treated as latent variables $z$ in the optimization process. In practice, this means that the minimization is performed iteratively. In the first step, $z$ is initialized using some heuristics and kept fixed, minimizing with respect to $w$ (the set of SVM margins corresponding to all the components). Then $w$ is kept fixed and the minimization is solved with respect to $z$ (and the process is iterated). In the latent variable minimization step, the initial values of $\beta$ can vary. But also the membership value of a sample $x$ can vary: this is called \emph{latent sub-categorization}, which means that an element $x$ is assigned to the component $k$ whose current hyperplane $(w_k , b_k )$ scores the maximum value for $x$.

More formally, consider learning a classification problem where examples are clustered into $K$ separate sub-classes (sub-categories), and a separate classifier is trained per subcategory. The assignment of instances to subcategories is modeled as a latent variable $z$. This binary classification task is formulated as the following (\emph{latent-SVM}) optimization problem that minimizes the trade-off between the $l_2$ regularization term and the hinge loss on the training data:

\begin{align}
\arg \min_{w} \frac{1}{2} \sum_{k=1}^{K} \parallel\omega_k\parallel^2 +~C\sum_{i=1}^{n} \epsilon_i, \label{eq:lsvm-1}\\
y_i~.~s_{i,z_i} \geqslant 1 - \epsilon_i,~\epsilon_i \geqslant 0, \label{eq:lsvm-2}\\
z_i = \arg \max_{k} s_{i,k}, \label{eq:lsvm-3}\\
s_{i,k} = \omega_k~.~\phi_k(x_i)+b_k. \label{eq:lsvm-4}
\end{align}

$\omega_k$ denotes the separating hyperplane for the $k$-th subclass, and $\phi_k$(.) indicates the corresponding feature representation. Since the minimization is semi-convex, the model parameters $\omega_k$ and the latent variable $z$ are learned using an iterative approach \cite{felzenszwalb2010object}. Note that given the latent assignment of examples to the subcategories $z_i$ , the optimization problem in equation (\ref{eq:lsvm-1}) boils down to solving $K$ separate one vs. all (binary) classification problems (corresponding to each subcategory). The latent relabeling step in equation (\ref{eq:lsvm-3}) acts as a simple yet important discriminative re-clustering step that deals with noisy assignments and fragmentation of the data (which occurs because the “optimal” number of clusters is not known in advance).

For simplicity, in formulas (\ref{eq:lsvm-1} - \ref{eq:lsvm-4}) no part decomposition and no model deformation is expressed, thus the only latent variables $z$ concern the sample membership. The value $z_i\in\{1, ..., K\}$ is assigned using (\ref{eq:lsvm-3} - \ref{eq:lsvm-4}) and depends on the classifiers $\{(w_k , b_k )\}$ computed in the previous iteration. The values $z_i$ are initialized using a simple clustering of the positive samples based on the aspect ratio of their ground truth bounding boxes. These aspect ratios are kept fixed for the whole minimization process and the image representation $\phi_k(x_i)$ in (\ref{eq:lsvm-4}) is an holistic representation which only depends on the specific sliding window aspect ratio associated with the $k$-th component. Finally, the spatial configuration of parts ($l$), as well as the number of parts ($P$) and the part rectangular shapes are not included into the latent variables and are fixed in the initialization step, using simple heuristics (see Section 5.2 in \cite{felzenszwalb2010object}).

\begin{figure}[t]
\begin{center}
\includegraphics[width=13cm]{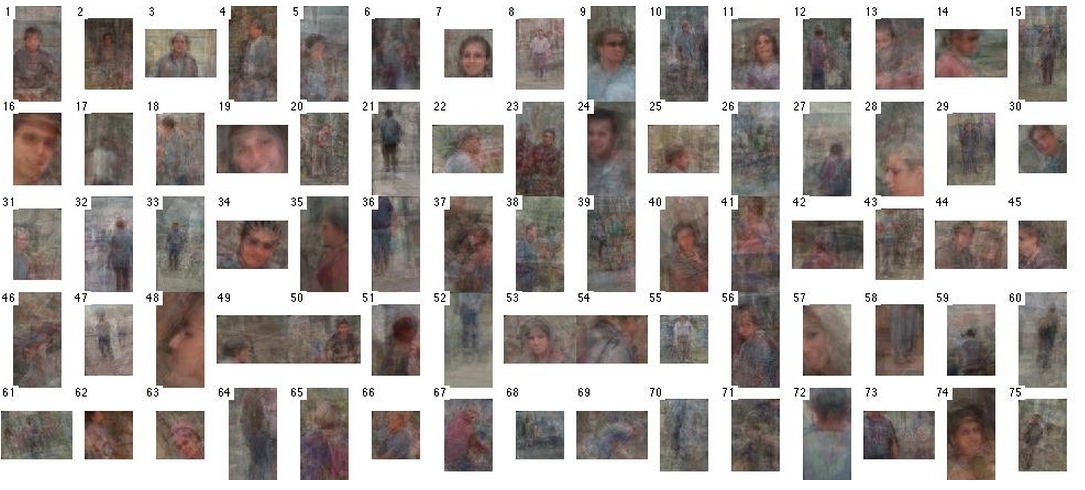} \\
\end{center}
\caption{Poselet for pedestrian category \cite{bourdev2009poselets}.}
\label{fig:poselet-top}
\end{figure}

\subsection{Poselets}

A poselet is novel definition of a "part" that capture a part of human pose under a given viewpoint. Poselets are tightly clustered in both appearance space (and thus are easy to detect) as well as in configuration space (and thus are helpful for localization and segmentation) \cite{bourdev2009poselets}. The key difference of poselet with other parts (e.g. latent parts in DPMs) is that a poselet is constructed in a strongly-supervised fashion, not from the raw pixel values of the images. As a result examples of poselets may vary very significantly in appearance and yet they capture common underlying semantics.

For a given part of human pose, finding similar pose configuration in the training set is very challenging. Poselet framework addresses this problem by employing extra supervision in terms of key-point annotation of human body joints. Poselets can effectively capture at the local configuration of key-points for each initial patch. Then, we search over all other training examples, to find the optimal geometric transformation which minimizes the discrepancies. The extracted patches are the corresponding patches to the given patch and the quality of match for each can be measured as the residual error after subtracting the similarity transform. Practically speaking, given a seed patch, for every other example we find the closest patch. We compute the residual error, sort patches by their residual error. A set of inconsistent examples would be thrown away by selecting a predefined threshold on the residual error. The remaining patches are positive training examples for training Poselet. Training poselet is done by extracting HOG feature from examples and fitting to a SVM model (similar to \ref{eq:svm}). The same strategy of scanning window and bootstrapping is done like other architectures.

This is the pipeline explained for an individual poselet. In practice we repeat this algorithm for thousands of poselet by picking the random windows, generate poselet candidates, and train linear SVMs. By this regime, we end up with a set of poselet classifiers each capture a specific pose of a particular part of human body (see Figure \ref{fig:poselet-top}(c)).

\begin{figure}[t]
\begin{center}
\includegraphics[width=8cm]{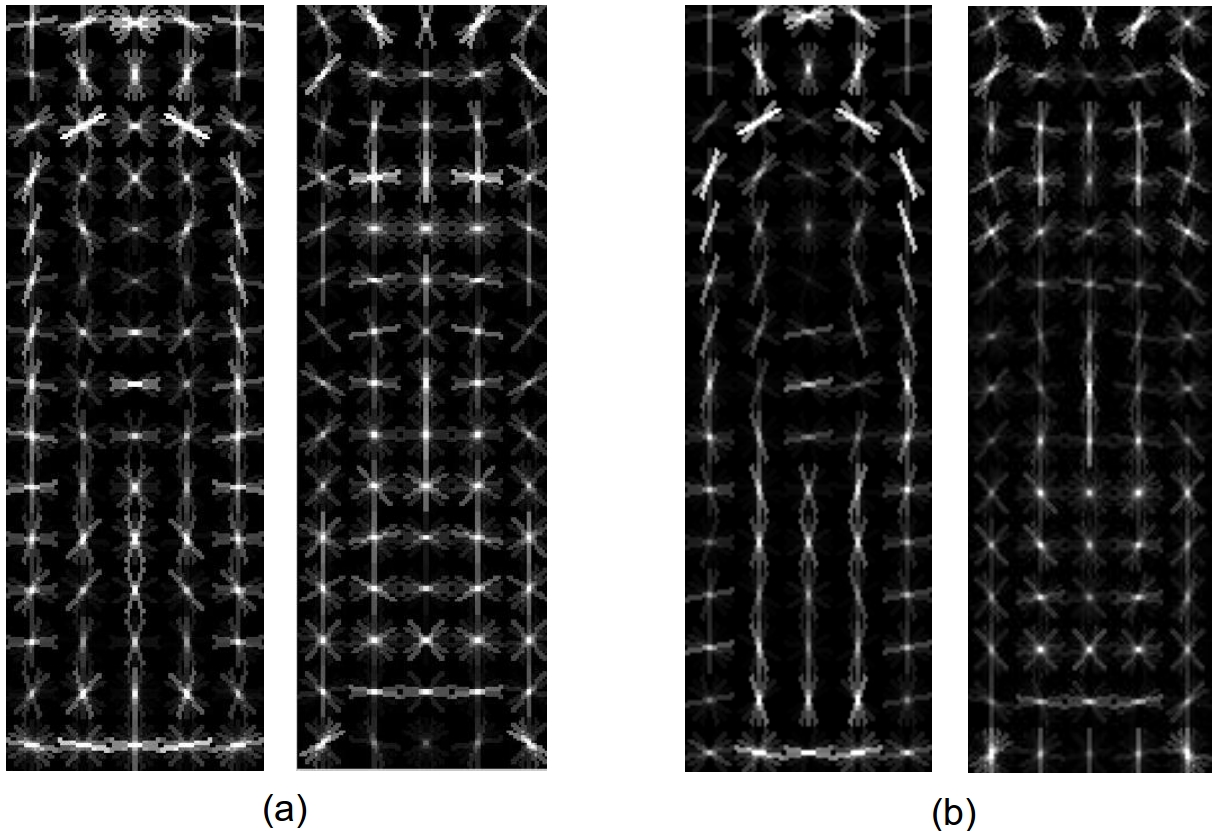} \\
\end{center}
\caption{HOG template trained for Pedestrian in \cite{hariharan2012discriminative}: (a) SVM, (b) LDA.}
\label{fig:lda}
\end{figure}

\subsection{Exemplar-SVMs}

The Ensemble of Exemplar-SVM is actually a collection of many binary SVM "sub-classifier". Each "sub-classifier" is trained using a single positive example for one class and an enormous collection of negative examples for the other. For example instead of discriminating horses vs. non-horse samples (standard two-class SVM), each "sub-classifier" discriminates between specific examples of horse (e.g., "jumping horse", "horse head") and many non-horse examples. Individual "sub-classifier" SVMs are trained for each example of the positive class, so you would have one SVM for "jumping horse", another for "horse head", yet another for "side-view barrel horse", and so on. The outputs of these "sub-classifier" SVMs are calibrated and combined to determine whether a horse, not just one of the specific exemplars, appears in the test data.

\subsection{Exemplar-LDA} \label{sec:Exemplar-LDA}
\qsection{Linear Discriminant Analysis (Fisher \cite{fisher1936use})} Although most of the object models train in discriminative ways, the simplest generative model that can be used to build a object classifier probably is the Gaussian model.
Assuming that both positive and negative sets come from a Gaussian distribution, log-likelihood ratio of the two Gaussian is the optimal   classifier. Particularly if two Gaussians have the same covariance matrix, then log-likelihood ratio can be computed with a linear operation.
\begin{center}
$\omega = \sum^{-1} (\mu_P-\mu_N)$
\end{center}

As shown in LDA formulation, the optimal boundary is in fact the linear function which is computed simply by multiplying overall average covariance and mean difference \cite{duda2012pattern}. In \cite{hariharan2012discriminative}, authors showed that this simple model surprisingly provides very similar parameters with SVM model (see Figure \ref{fig:lda}) as well as a close performance in detection (Average Precision equal to 75\% vs. 80\% on INRIA pedestrian dataset \cite{dalal2005histograms}). In LDA model, training covariance matrix is done by computing the weighted average of two class-specific covariances where the weights are given by the prior probabilities.

\qsection{LDA accelerated version of Exemplar-SVM} Since in E-SVM (and in general in detection task) the negative set is very huge, so the LDA covariance matrix is basically the negative covariance. So, this covariance matrix has nothing to do with any object specific notion. So in practice it can be trained once and reused for all exemplar classifiers (and object categories). In training a new exemplar classifier all needed to be done is to extract the HOG of that exemplar, subtract it from the (pre-computed) average of negatives, then multiple by the (pre-computed) negative covariance.

Note that once we store the mean and covariance of negative samples somewhere, training Exemplar-LDA can be done efficiently as opposed to the very time-consuming step of hard-negative mining for Exemplar-SVM in large-scale datasets.

\subsection{Latent-LDA} \label{sec:latent_LDA}

The training efficiency of LDA over SVM, as a result of removing hard-negative mining step, is further adopted for training DPMs in \cite{girshick2013training}. This modification accelerates the learning time of DPM without any significant drop in the performance.
The authors in  \cite{girshick2013training} also modify the LSVM optimization problem so that it involves only positive examples and coarse background statistics. Optimizing the new optimization problem also involves alternating between latent label and model parameters. However, when the latent labels are fixed, the optimal parameter vector can be solved for in closed form. It has been shown that if the input features are whitened, then the optimal parameters form a set of LDA classifiers \cite{hariharan2012discriminative}. Unlike latent SVM, latent-LDA uses efficient closed-form updates and does not require an expensive search for hard negative examples \cite{girshick2013training}.


\section{Unsupervised Visual Subcategory}
Image understanding tasks are often formulated as binary classification problems, where the positive examples are category images/bounding boxes, and the negative examples are background images/patches. Although HOG + SVM paradigm \cite{dalal2005histograms} was effectively applicable for old datasets (e.g. Caltech \cite{fei2007learning}), it works poorly in new ones (e.g. PASCAL \cite{everingham2010pascal}). This substantial drop in performance is mainly due to the simple appearance variations in old-fashion datasets compare to the rich diversity in modern ones. Recent approaches address this problem by introducing diversity into the classification model by partitioning the category into smaller clusters, i.e. \emph{subcategories} and learning multiple subcategory classifiers. The less diversity in smaller cluster leads to break down the difficult non-linear learning problem to a set of simpler linear problems. However, splitting a category into subcategories itself is a computationally expensive problem with the exponential complexity (equal to Bell number \cite{BellNum}). There is also an infinite number of ways to split a basic-level category into subcategories. For example, one can define car subcategories by its poses/viewpoints (e.g., left-facing, right-facing, frontal), while others can do it based on car manufacturer (e.g., Fiat, Honda, Toyota), or even some functional attribute (e.g., sport cars, public transportation, utility vehicle).

In the literature, apart from the core methodological aspects, a substantial debate has been posed over the level of supervision used for the sub-categorization. In this context, two classes of approaches can be identified, which are related to the different level of used supervision:
On one hand, \emph{supervised} sub-categorization approaches have considered reorganizing the data into groups using either extra ground-truth annotations (e.g., viewpoint \cite{chum2007exemplar}, pose \cite{bourdev2009poselets}, taxonomy \cite{deng2009imagenet}), or empirical heuristics (e.g., aspect-ratio \cite{felzenszwalb2010object}). Having such supervision may not be available for many modern large datasets. Heuristics are also often brittle and fail to generalize to a large number of categories. On the other hand, \emph{unsupervised} sub-categorization approaches directly use appearance for building the `visually homogeneous' subcategories instead of using semantics or heuristics. It is interesting mainly because providing supervision is so costly for big data, and heuristics are often biased \cite{divvala-eccvw12}.

Recently, hybrid \emph{semi-supervised} sub-categorization approaches became popular, due to existence of several billions of unlabeled images available on the web. This type of approaches automatically benefit from this wealth of information can help alleviating the labeled data barrier. Recently-introduced notion of webly-supervised \cite{santosh2014web,chen2013neil} methods can be interpreted as such approaches.

There are some recent works which have shown the benefits of using unsupervised subcategories in improving categorization performance \cite{aghazadeh2012mixture,gu2012multi,zhu2014capturing,hoai2013discriminative,azizpour2014self}.

We mainly address unsupervised sub-categorization approaches in this section. A more comprehensive study on visual subcategory is available \cite{divvala2011visual,divvala2012context}.


\subsection{State of the art: Parts vs. Subcategories} \label{sec:partvssubcat}
In the vision community there is a debate on \emph{what are the most important elements in the DPM framework} and how this approach can possibly be extended in order to improve its performance?\\ 
There are basically two schools of thought: one believes that the strength of the model is the multi-component structure, while the other is more prone to trust in the part-based representation and the deformable architecture.\\
In \cite{divvala-eccvw12}, Divvala et al. experimentally show that using a simple holistic representation but splitting positive training data in different, more or less homogeneous subcategories, it is almost enough to reach the same performance of the whole Felzenszwalb’s framework. This is reasonable because, since linear SVMs are used as classifiers, they are probably not able to cope with the whole appearance variations of real object categories. Thus, subcategories, based, for instance, on specific viewpoints (e.g., a component for frontal pedestrians, another component for profile views, etc.) make the learning process easier and the learned templates $w_k$ cleaner. This results are confirmed by \cite{girshick2011object}, where Felzenszwalb et al. achieve a higher accuracy with respect to \cite{felzenszwalb2010object} increasing the number of components. Divalla and colleagues further exploit the subcategory principle in \cite{divvala2011object}, where they extend the discriminative clustering approach introducing a soft sample assignment, which allows the positive image $x$ to be shared among different components.\\
Somehow in contrast to the previous authors, the second school of thought believes that the deformable part structure is the secret of the success. In \cite{zhu2012we} Zhu, Ramanan et al. on the one hand recognize that clean data, partitioned in subcategories help improving accuracy. But, on the other hand, claim that the compositional model of DPMs is necessary to reach very high performance. To prove this, they use a DPM framework similar to \cite{Yang2013pami,zhu2012face}, in which the parameters of the object's part classifiers can be shared among different mixture components. As shown by their experimental results, sharing classifiers among different components makes it possible to decrease the total number of necessary positive training samples. In fact, an object part (e.g., the 'nose' in face detection task) can be common in different subcategories of the categories (face). In that case, once we used a sufficient amount of samples of a specific subcategory (e.g., images of frontal faces) for learning the parameters of the part's classifier, the same classifier can be used without retraining in other subcategories (e.g., different but similar face profiles). Anyway, this must be done carefully, since the appearance of the same part can vary in different subcategories. For instance, the appearance of a human hand usually largely differs in frontal and profile views of the body due to deformations produced by 3D rotations. This is confirmed by the experimental results obtained by the same authors in \cite{zhu2012we,zhu2012face}, where part sharing is useful especially when no sufficient training samples are given for all the subcategories, while \emph{component-specific part types} \cite{zhu2012we,zhu2012face,Yang2013pami} are chosen instead of part sharing when sufficient training data are available.\\
Finally, Ramanan and colleagues in \cite{zhu2012we} experimentally show that the matching cost function based on the “springs and templates” model \cite{fischler1973representation}, which is embedded in typical DPMs (see the deformation parameters $\beta$ in Figure \ref{fig:dpm} (c)), implicitly allows to synthesize unseen spatial configurations of parts (i.e., unseen subcategories) at training time, thus improving the generalization abilities of the system, especially with few training data. See \cite{Yang2013pami} for more details.

\subsection{Why Unsupervised Subcategories?}
\qsection{Classifier generalization in high visual diversity}\\
Images in a object categories are visually diverse and classifier generalization is so critical in such scenario. The generalization is achieved by either using sufficient labeled training data or enforcing prior knowledge as a regularizer to prevent overfitting to the given training samples \cite{friedman2001elements}. Even with a good regularization prior, learning a single generalizable classifier for a category is, however, challenging due to the high intra-class variations which often lead to multi-modality of training samples distributions. Typically, multi-modality is addressed by a non-linear classifier using a kernel. Having only a few training samples with high dimensional features, however, is not a desirable situation for kernel methods due to overfitting. The subcategory-based approaches address this problem by introducing diversity into the classification model by dividing the category into several subcategories and learning linear subcategory classifiers. The linear models are preferred to nonlinear SVMs also for their computational efficiency.

\qsection{Alleviating Human Supervision}\\
The high cost of providing human supervision for sub-categorization (especially for modern large-scale datasets) in one hand, the subjective nature of such supervision on the other hand grow popularity of unsupervised sub-categorization. Beside that, using the intermediate subcategory models directly for fine-grained recognition is also a plus we can have for free.
Furthermore, there are some works in the literature which observed that the unsupervised initialization performed on par with taxonomy-based initialization on a subset of IMAGENET dataset \cite{deselaers2011visual} and viewpoint-based clustering with the same $K$ on the PASCAL VOC 2007 “car” object category \cite{gu2010discriminative}. These results indicates that human supervision for creating the fine-grained semantic subcategories in order to train a basic-level category classifier may not even be of great benefit compared to the unsupervised visual subcategories.

\begin{figure}[t]
\begin{center}
\includegraphics[width=12cm]{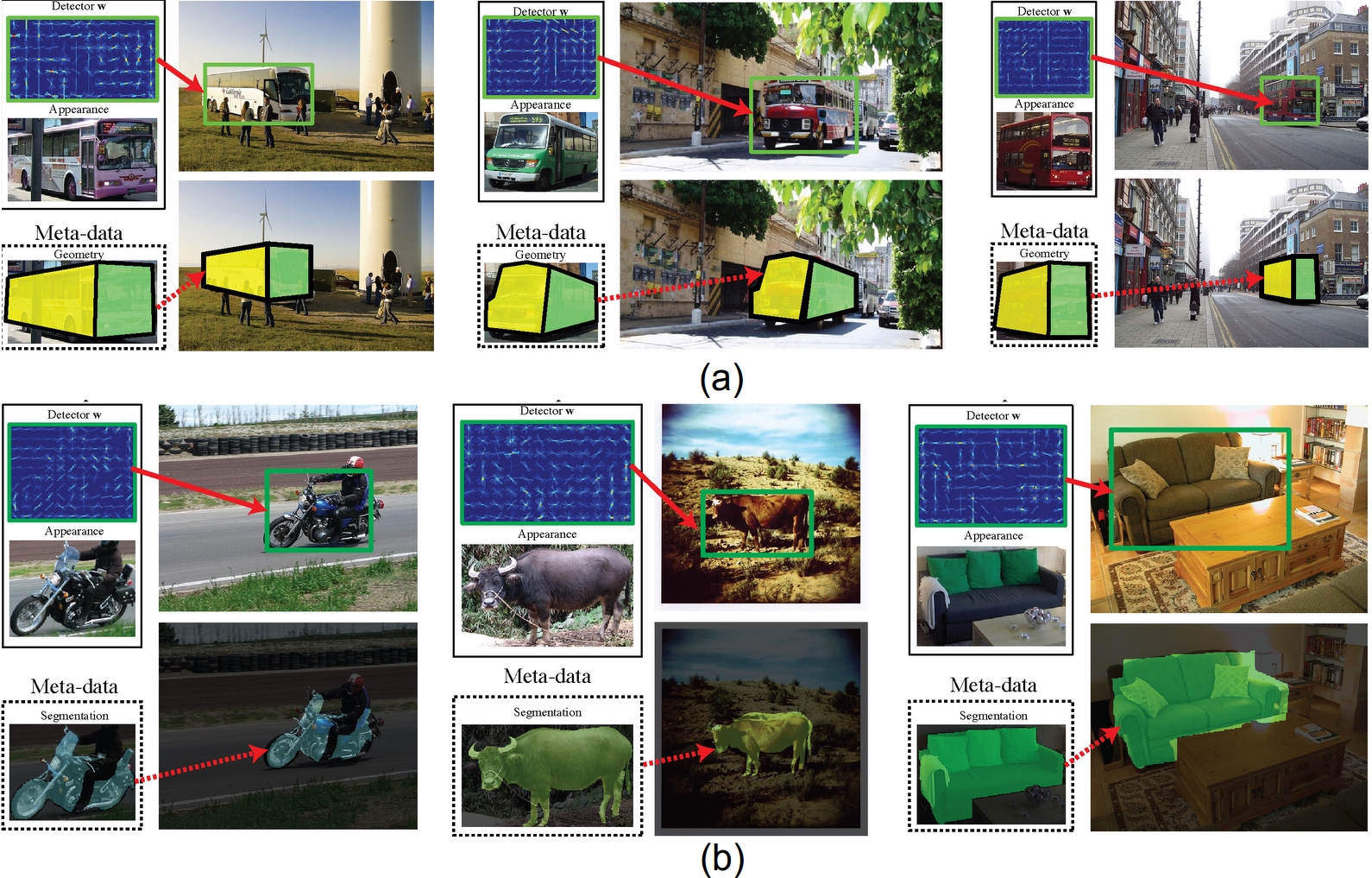} \\
\end{center}
\caption{Meta-data Transfer: (a) 3D geometry, (b) Segmentation (\cite{malisiewicz2011ensemble}).}
\label{fig:esvm}
\end{figure}
\qsection{Semantic Interpretablity}\\
Beyond the categorization task, subcategories offer richer information that could be useful in arriving at a deeper interpretation of the image. For example, rather than simply labeling a bounding box as ‘horse’, it will now be possible to provide more information about the detection as it being a ‘front-facing horse’ or a ‘horse jumping over a fence’, or describe it by transferring any meta-data information (part locations, segmentations, 3D geometry, object interactions(visual phrases\cite{farhadi2013phrasal}), etc) associated with the subcategory. It must be emphasized that such an interpretation is not possible in case of monolithic classifiers (e.g., single global linear or nonlinear SVM). A bunch of such applications have been introduced for Poselet (segmentation \cite{brox2011object}) and Exemplar-SVMs and can be extended to subcategories similarly. (Figure \ref{fig:esvm} from \cite{malisiewicz2011ensemble}).

\subsection{Challenges and Limitations} \label{sec:sub_limit}
Here is a list of key aspects/limitations of current methods on subcategory-based object detection systems have:

\qsection{Number of Subcategories}\\
An important parameter to be set is the number of subcategories $K$. This is unknown apriori and depends on the category. For example, a large rigid object (e.g car) would need fewer subcategories, while a deformable object (e.g cat) is expected to require a larger number. To deal with this problem, in \cite{divvala-eccvw12} they initialize the number of subcategories to a large number and subsequently prune the noisy subcategories via a calibration step. Differently with previous works, they found the optimal number of subcategories as 10-15.

\qsection{Initializing Subcategories}\\
Proper initialization of clusters is a key requirement for the success of latent variable models. In \cite{divvala-eccvw12,ye2013human} authors analyzed the importance of the appearance-based initialization and compare it with the aspect-ratio based
initialization of Felzenszwalb et al. \cite{felzenszwalb2010object}. They found the simple K-means clustering algorithm using Euclidean distance function can provide a good initialization of clusters. See \cite{zhu2012we} for more considerations.

\qsection{Subcategory Ensembling}\\
The subcategory based method is an ensemble classification approach that uses multiple classification models to obtain better predictive performance \cite{rokach2010ensemble}. Existing methods include ‘cross-validation committee’, ‘bagging’ and ‘boosting’ \cite{rokach2010ensemble}. Each classifier in the ensemble captures traits of a random subset of the training set. The ensemble of classifiers is usually learned independently. Matikainen et al. modified AdaBoost to learn a good ensemble of classifiers for recommendation tasks \cite{matikainen2012classifier}. In latent subcategory models \cite{felzenszwalb2010object,andrews2002support}, ensemble classifiers are learned jointly.

\qsection{Calibrating of Subcategory Classifiers}\\
Calibration of individual classifiers is another key step for the success of visual subcategory classifiers. It plays a dual role as it not only suppresses the influence of noisy clusters but also maps the scores of
individual classifiers to a common range. This problem is usually tackled by transforming the output of each of the individual classifier by a sigmoid to yield comparable score distributions (i.e., Platt\cite{platt1999probabilistic}). A major difficulty in estimating the parameters is that we do not have the ground truth cluster labels for instances in the training set. To circumvent this issue, usually we make an assumption that the highly-confident classifications of a particular classifier are instances belonging to that particular subcategory (i.e., most of the classifiers do demonstrate a low-recall, high-precision behavior). A common approach to calibrate the sub-classifiers in to apply all sub-classifier to all the training images and compute the overlap score (area of intersection over union) with the ground-truth. So, treat all detections with overlap greater than 0.5 as positives and all detections with overlap lower than 0.2 as negatives \cite{malisiewicz2011ensemble,divvala-eccvw12}

\qsection{Evaporation Effect}\\
Latent subcategory model (LSM) formulations suffer from the \emph{Evaporation Effect}\cite{girshick2013training}. (also known as \emph{Cluster degeneration}\cite{hoai2013discriminative}] , i.e., the situation where a few clusters dominate and claim all the points, leading to many empty clusters. This is problematic for sub-category discovery because the number of sub-categories obtained is then smaller than the sought number. Cluster degeneration has been pointed out to be an inherent problem of LSM. where has been resolved in \cite{hoai2013discriminative} by incorporating negative samples into the learning framework.

\label{sec:long-tail} \qsection{Long-tail Distribusion of Subcategories}\\
A recent argument in the object recognition community is that the frequency of samples in subcategories of an object category follow a long-tail distribution: a few subcategories are common, while many are rare, but rare cases collectively cover a large collection of the population. In other words, increasing the number of subcategories leads to improved homogeneity at the cost of leaving very few samples per subcategory, which might be insufficient to learn a robust classifier. This phenomenon leads the model to work well for usual cases of the object category e.g. \emph{upright human body} by capturing iconic object appearance, but fail in the case of abnormal appearance due to inadequate modeling of the tail e.g. \emph{Yuga twist poses} (see Figure \ref{fig:long-tail-1}). Many approaches have been proposed to come up with minimizing this phenomenon with different types of solutions. We observed three main trends in the literature to cope with this problem: 

\begin{figure}[t]
\begin{center}
\includegraphics[width=6.0cm]{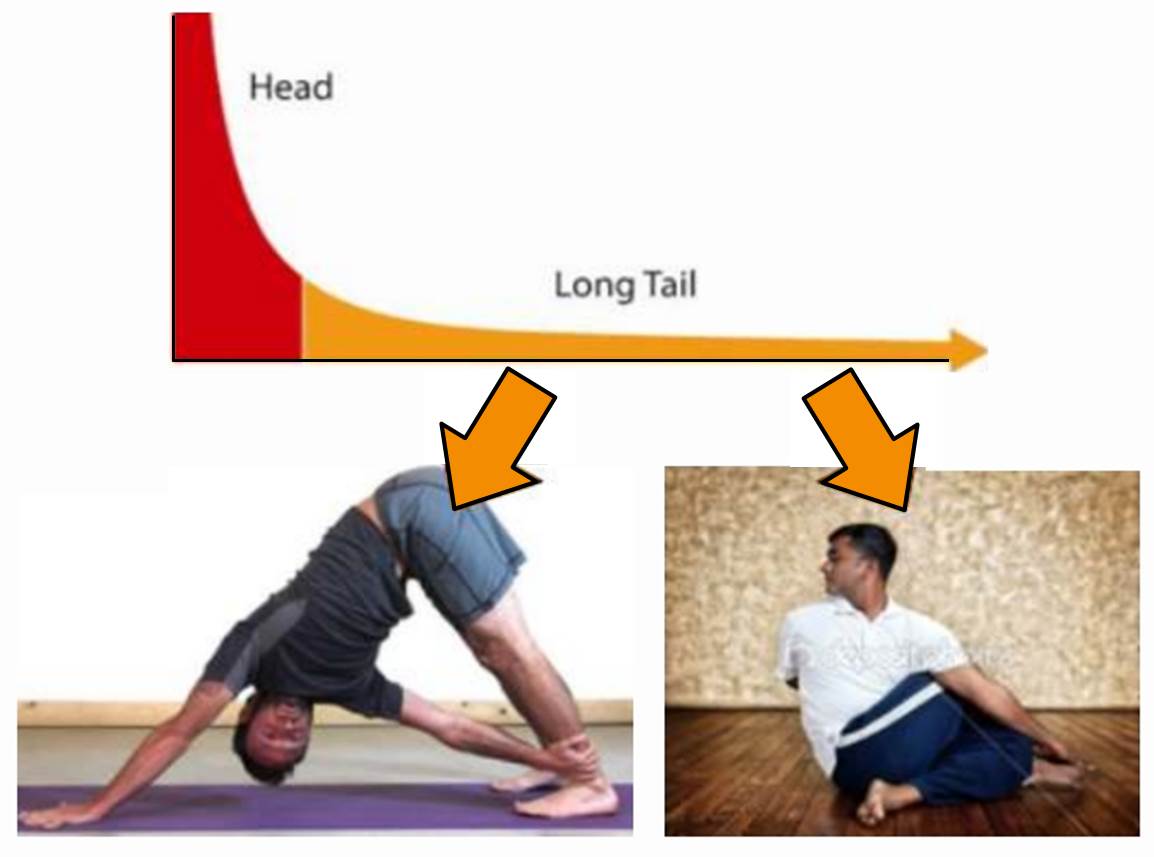} \\
\end{center}
\caption{Long-tail distribution of subcategories of a category (human body pose).}
\label{fig:long-tail-1}
\end{figure}

\begin{enumerate}
\item 
\emph{Sample sharing across subcategories} aims at assigning samples not just to one subcategory, but sharing it across subcategories in a soft assignment fashion. In this way, we can come up with subcategories with more samples inherited from other subcategories. 
Zhu et al. proposed learning overlapping subcategory classifiers to overcome information insufficiency \cite{zhucapturing} and Divalla et al. in 
\cite{divvala2011object} formulate an objective function to learn an ensemble of subcategory classifiers that share information among the subcategories.

\item 
\emph{Part sharing across subcategories}, on the other hand, aims at finding a set of mid-level patches (parts) and share them across subcategories of an object category. This provides us with a model which not only can represent observed object subcategories but also synthesize unseen configurations to form rare subcategories. Mid-level patches (parts) can be discovered and trained in supervised\cite{Yang2013pami,bourdev2009poselets}, weakly supervised\cite{felzenszwalb2010object} and unsupervised fashion\cite{singh2012unsupervised,juneja2013blocks,doersch2013mid,endres2013learning}.

\item 
\emph{Adding unlabeled samples to impoverished subcategories} the final solution to cope with the long-tail distribution of examples is by enriching the subcategories with few sample in a semi-supervised fashion. One of the successful trend in this regard is the recently introduced notion of \emph{web-supervision}. The general idea behind this school of thought is to leveraging the gigantic collection of unlabeled images available (e.g. on the web) for populating the impoverished subcategories.\cite{santosh2014web,chen2013neil}
\end{enumerate}

\qsection{Pooling detections from multiple subcategories}\\
Selecting the correct box from a pool of valid detections in a completely unsupervised setting is a challenging research problem. The greedy none-maximum suppression, co-occurrence matrix \cite{malisiewicz2011ensemble}, subcategory restoring \cite{santosh2014web} are some heuristic-based schemes introduced for this purpose. 

In \emph{none-maximum suppression (NMS)} scheme, given a set of detections, the detections is sorted by score, and greedily select the highest scoring ones while skipping detections with bounding boxes that are at least 50\% covered by a bounding box of a previously selected detection. In \emph{co-occurrence matrix} scheme, for each detection a context feature similar to \cite{bourdev2010detecting} is generated which pools in the SVM scores of nearby (overlapping) detections and generates the final detection score by a weighted sum of the local SVM score and the context score. After obtain the final detection score, the standard non-maximum suppression is used to create a final, sparse set of detections per image. And finally in \emph{subcategory restoring} scheme, we find a weight vector for restoring subcategory detections on a held-out dataset and utilize that weight vector for the given dataset as well.

\newpage
\thispagestyle{empty}
\mbox{}

\part*{Part I: Mid-level Representation for Image Understanding}
\chapter{Subcategory-based Discriminative Patches} \label{chap:image}
{\LARGE \qsection{Application}\emph{ Object Recognition}}\\[2cm]

In this chapter, we study the outcomes provided by employing the subcategories of an object category for \ignore{the task of} object recognition. The improvement of using subcategory is mainly provided as a result of breaking down the problem of [object] category recognition into a set of sub-problems (i.e., subcategory recognition) which is simpler to be solved directly. The solutions to the sub-problems are then combined to give a solution to the original problem. The simplicity of the sub-problems is due to the fact that the images within each subcategory have limited intra-class variance compared to an object category, so are easier to be modeled. Here, we specifically show two studies provided by using subcategory-based modeling for object recognition which are related to the \emph{undoing dataset bias}~\cite{stamos2015learning} and \emph{webly-supervised training of the discriminative patches} problems. 

As the first study, we analyze how the subcategory-based object modeling can tackle the dataset bias problem. To come up with a solution for long-tail distribution problem in subcategories (see section \ref{sec:long-tail}), a natural way is to take advantage of statistical power provided by merging different datasets. Training from combined sets, however, needs to be done with care due to the bias inherent in computer vision datasets. We address this problem from a multitask learning perspective, combining new ideas from subcategory-based object recognition. The proposed methodology leverages the availability of multiple biased datasets to tackle a common classification task (e.g., horse classification) in a principled way. 

For the second problem, we introduce a framework for discovering and training a set of mid-level features (i.e., local patches) for image representation. We aim at mining these features in a discriminative way to form a richer representation for visual object recognition. To this purpose, we find the discriminative patches in an object subcategory (as opposed to category) with less appearance variations (e.g different viewpoints, lighting conditions, types and etc. of an object). It will provide us with a richer discriminative patches, consequently better image representation for object recognition.

Following in this chapter, we first explain the \emph{subcategory discovery} using Latent Subcategory Model(LSM) as well as investigating the subcategory initialization issue in this model. For this purpose, we provide the theoretic proof as well as experimental results justifying a standard initialization heuristic which is advocated in the object recognition literature. We then study how this subcategory-based object modeling can tackle dataset bias. We, specifically, introduce a framework that simultaneously learn a set of biased LSMs as well as a compound LSM (visual world model) which is constrained to perform well on a concatenation of all datasets. We, finally, focus on the problem of discovering a set of discriminative patches for a particular object subcategory without almost any supervision except a set of unlabeled Internet images. We select the patches which are fixed position relative to the bounding box of the object subcategory. For each subcategory we initialize with a set of patches and then select a subset of them by an introduced criterion based on the spatial and visual consistency on positive and negative samples for that visual subcategory.

\begin{figure}[t]
\begin{center}
\includegraphics[width=6.0cm]{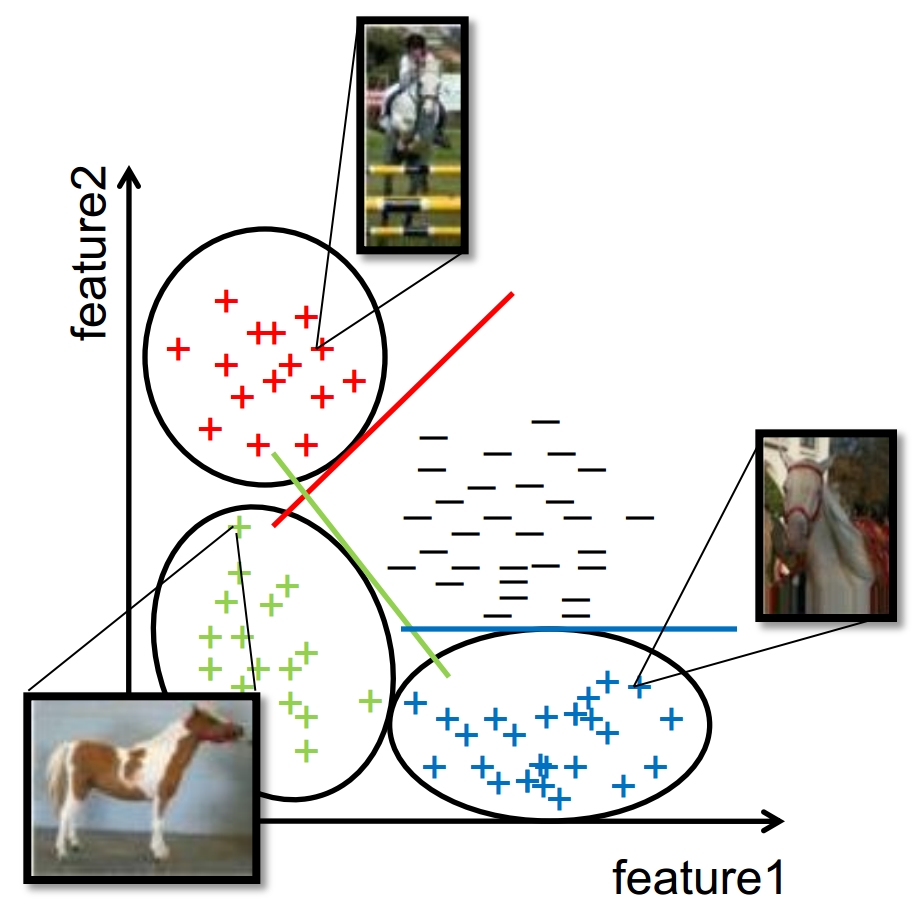} \\
\end{center}
\caption{Subcategories of a category (horse).}
\label{fig:subcategory}
\end{figure}

\section{Introduction} \label{sec:Intro}
The problem of object recognition has received much attention in computer vision. One of the most challenging problems in object recognition is the large inter-class appearance variability of a semantic object category(e.g., horse). This variations is mainly due to the diversity in object types, shapes, viewpoints, poses, illuminations and so on. In the case of {\it horse} as an object category, not only different kinds of horses exist, but also each one can be captured in different poses, viewpoints, or under different lightening conditions. This makes the problem of object recognition extremely difficult, especially for the class of objects which are variant, articulated and appearing in the wild contexts.

One of the most successful strategies for tackling this problem is the recently introduced notion of {\it subcategory}. This strategy is simply organizing examples of a category to the smaller groups (i.e., subcategories) with less visual diversity. It provides us with {\it cleaner} subset of examples so we can simply train separate models for each individual subcategory, then building a final category model by ensembling all subcategory models. This solution breaks down the difficult problem of finding a non-linear model to a finite set of linear models which collectively performs as an approximation of the former non-linear model (see Figure \ref{fig:subcategory} and section \ref{sec:subcat} for more details).  

One of the most successful \emph{subcategory-based} object recognition systems is based on deformable part-based models (DPMs), see \cite{Divvala2012,Felzenszwalb2010,Girshick2013,zhu12} and references therein. As mentioned in previous chapters, special case of Latent SVMs are latent subcategory models (LSM) \cite{gu2010,Divvala2012,zhu12}. It is generally acknowledged that these techniques provide a significant step forward over standard SVMs, because they can cope with finer object variations in a principled way. This approach has proved useful when the object we wish to classify or detect consists of multiple components (i.e., subcategories), each of which captures different characteristics of the object class. For example, components may be associated with different viewpoints, light conditions, etc.  Under these circumstances, training a single global classifier on the full dataset may result in a low complexity model which underfits the data. To address this, latent subcategory models train multiple subclassifiers simultaneously, each of which is associated with a specific linear classifier capturing specific characteristics of the object class. 

Training these models is a challenging task due to the presence of many local optima in the objective function and the increased model complexity which requires large training set sizes. An obvious way to have larger training set sizes is to merge datasets from different sources. However, it has been observed by \cite{ponce2006dataset} and later by \cite{torralba2011} that training from combined datasets needs to be done with care. Although we would expect training a classifier from all available data to be beneficial, it may in fact result in decreased performance because standard machine learning methods do not take into account the \emph{bias} inherent in each dataset. For example, the merged training set may be obtained by collected images of people from multiple cameras and in different working environments, the information of which camera and which location each image is associated being disclosed during learning. To address this problem several approaches have been considered, some of which we review in the next section. 

In general terms, the part-based object representation ,tackles the large visual diversity presented in an object category. This outcome is provided as the result of finding a set of mid-level parts (i.e., patches) representing that category with those parts. The mid-level patches can be discovered and trained in supervised~\cite{Yang2013pami,bourdev2009poselets}, weakly supervised~\cite{felzenszwalb2010object} and unsupervised fashion~\cite{singh2012unsupervised,juneja2013blocks,doersch2013mid,endres2013learning}. These patch-based object representation will provide us with an object model which not only can represent observed object instances but also synthesize unseen patch configurations of the object. 

\subsection{Discussion and Contribution}

As a first contribution of this work, we observe that if the positive examples admit a good $K$-means clustering, a good suboptimal solution can be obtained by simply clustering the positive class and then training independent SVMs to separate each cluster from the negatives. This result supports a commonly used heuristic for training subcategory models \cite{Divvala2012}. 

The second contribution of this work is to extend LSMs to deal with multiple biased datasets. We simultaneously learn a set of biased LSMs as well as a compound LSM (visual world model) which is constrained to perform well on a concatenation of all datasets. We describe a training procedure for our method and provide experimental analysis, which indicates that the method offers a significant improvement over both simply training a latent subcategory model from the concatenation of all datasets as well as the undoing bias method of \cite{Khosla2012}. Hence, our approach achieves the best of both worlds. 


As a third contribution of this work, we follow the regime of discovering and learning mid-level patch as an intermediate representation. However, instead of mining and training these patches based on semantic object categories, we model them based on visual subcategories, which are visually homogeneous clusters and have much less intra-class diversity. We argue that since examples in each subcategory are visually consistent, deformable patches can be substituted with fixed-position patches in the subcategory level. So, we particularly focus that the key ideas behind this contribution is to discover and train discriminative patches while they are fixed-position relative to the bounding box of the object. We also employ the powers of webly-supervised methods when we use the large amount of unlabeled images available on the Internet which have been put forward in \cite{santosh2014web}. We are related in nature to this work, but differently, we employ the patch-based mid-level representation for each subcategory. 

%

\section{Subcategory Discovery} 
\label{sec:lsm}
In this section, we revisit latent subcategory models for the task of subcategory discovery. We let $K$ be the number of linear subclassifiers and $(\w_1,b_1),\dots,(\w_K,b_K)$ be the corresponding parameters. A point $\x$ belongs to the subclass $k$ if $\lb \w_k, \x\rb +b_k > 0$, where $\lb \cdot,\cdot\rb$ denotes the standard scalar product between two vectors. For simplicity, throughout this section we drop the threshold $b_k$ since it can be incorporated in the weight vector using the input representation $(\x,1)$. A point $\x$ is classified as positive provided at least one subclassifier gives a positive output\footnote{The model does not exclude that a positive point belongs to more than one subclass. For example, this would be the case if the subclassifiers are associated to different nearby viewpoints.},
 or in short if $\max_k \lb \w_k, \x\rb >0$. 
The geometric interpretation of this classification rule is that the negative class is the intersection of $K$ half-spaces. Alternatively, the positive class is the union of half-spaces.

A standard way to learn the parameters is to minimize the objective function \cite{Divvala2012,Girshick2013,zhu12}
\beq
E_{K,\lambda}(\w) = \sum_{i=1}^m L(y_i\max_{k} \lb \w_k, \x_i\rb) + \lam \Omega(\w)
\label{eq:model}
\eeq
where $(\x_i,y_i)_{i=1}^m$ is a training sequence, $L$ is the loss function and, $\Omega(.)$ is the regularizer. 
With some abuse of notation, $\w$ denotes the concatenation of all the weight vectors. In this work, we restrict our attention to the hinge loss function, which is defined as 
$L(z) = \max(0,1-z)$. However, our observations extend to any convex loss function which is monotonic nonincreasing, such as  the logistic loss. 

We denote by $P$ and $N$ the index sets for the positive and negative examples, respectively, and decompose the error term as 
\beq
\sum_{i \in P} L(\max_k \lb \w_k, \x_i\rb)+\sum_{i \in N} L(-\max_k \lb \w_k, \x_i\rb).
\eeq
Unless $K=1$ or $K = |P|$, problem \eqref{eq:model} is typically nonconvex\footnote{If $\lam$ is large enough the objective function in \eqref{eq:model} is convex, but this choice yields a low complexity model which may perform poorly.} because the loss terms on the positive examples are nonconvex. 
To see this, note that 
$L(\max_k \lb\w_k, \x\rb) = \min_k L(\lb \w_k , \x\rb)$, which is neither 
convex nor concave\footnote{The function $L(\lb \w_k , \x\rb)$ is convex in $\w_k$ but the 
minimum of convex functions is neither convex or concave in general, see e.g., \cite{boyd2009convex}.}. On the other hand the 
negative terms are convex since $L(-\max_k \lb \w_k , \x\rb) = \max_k L(-\lb \w_k,  \x\rb)$, and  the maximum of convex functions remains convex. 

The most popular instance of \eqref{eq:model} is based on the regularizer $\Omega(\w)= \sum_k \|\w_k\|^2$ \cite{Divvala2012,gu2010,zhu12} and is a special case of standard DPMs \cite{Felzenszwalb2010}. 
Note that the case $K=1$ corresponds essentially to a standard SVM, whereas the case $K=|P|$ reduces to training $|P|$ linear SVMs, each of which separates one positive point from the negatives. The latter case is also known as exemplar SVMs \cite{Malisiewicz2011}. 

It has been noted that standard DPMs suffer from the ``evaporating effect'', see section \ref{sec:sub_limit} and \cite{Girshick2013} for a discussion. This means that some of the subclassifiers are redundant, because they never achieve the maximum output among all subclassifiers.
To overcome this problem, the regularizer has been modified to \cite{Girshick2013} 
\beq
\Omega(\w) = \max_k \|\w_k\|^2.
\label{eq:maxreg}
\eeq
This regularizer encourages weight vectors which have the same size at the optimum (that is, the same margin is sought for each component), thereby mitigating the evaporating effect. 
The corresponding optimization problem is slightly more involved since now the regularizer \eqref{eq:maxreg} is not differentiable. However, similar techniques to those described below can be applied to solve the optimization problem.

A common training procedure to solve \eqref{eq:model} is based on alternating minimization. We fix some starting value for $\w_k$ and compute the subclasses $P_k = \{i \in P : k = {\rm argmax}_\ell \lb \w_\ell,  \x_i\rb \}$. We then update the weights $\w_k$ by minimizing the convex objective function
\begin{eqnarray}
 \nonumber
F_{K,\lam}(\w) = \sum_{k=1}^K \sum_{i \in P_k} L(\lb \w_k , \x\rb) ~~~~~~~~~~~~~~~~~~~~~~~~~~~~~~~~~~~~& \\
~~~~~~~~+ \sum_{i \in N} L(-\max_k \lb \w_k, \x_i\rb) + \lam \Omega(\w_1,\dots,\w_K).& 
\label{eq:Kmeans}
\end{eqnarray}
This process is then iterated a number of times until some convergence criterion is satisfied. The objective function decreases at each step and in the limit the process converges to a local optimum. 

A variation to \eqref{eq:model} is to replace the error term associated with the negative examples~\cite{gu2010,zhucapturing} by 
\beq
\sum_k \sum_{i \in N} L(-\lb \w_k, \x_i \rb)
\label{eq:neg}
\eeq
This results in a simpler training procedure, in that the updating step reduces to solving $K$ independent SVMs, each of which separates one of the clusters from all the negatives. Each step can then be solved with standard SVM toolboxes. 
Often in practice problem \eqref{eq:model} is solved by stochastic subgradient methods, which avoid computations that require all the training data at once and are especially convenient for distributed optimization. Since the objective function is nonconvex, stochastic gradient descent (SGD) is applied to a convex upper bound to the objective, which uses a difference of convex (i.e., DC) decomposition: 
the objective function is first decomposed into a sum of a convex and a concave function and then the concave term is linearly approximated around the current solution. In this way we obtain an upper bound to the objective which we then seek to mininimize in the next step. One can refer to \cite{Girshick2013} for more information.

Finally we note that LSMs are a special case of DPMs without parts. Specifically, a DPM classifies an image $\x$ into one of two classes according to the sign of the function $\max_{k,h} \w_k\trans \bm{\phi}_k(\x,h)$. Here $k\in\{1,\dots,K\}$ is the latent component and $h$ is an additional latent variable which specifies the position and scale of prescribed parts in the object, represented by the feature vector $\bm{\phi}_k(\x,h)$. 
LSMs do not consider parts and hence they choose $\bm{\phi}_k(\x,h) =\x$ and discard the maximum over $h$. Our subcategory-based undoing bias methodology, presented in section \ref{sec:method} , extends to DPMs in a natural manner, however for simplicity in this work we focus on LSMs. 

\section{Subcategory Initialization}
\label{sec:Init}

As we noted above, the objective function of an LSM \eqref{eq:model} is difficult nonconvex optimization problem, which is prone to many local minima. The nonconvexity nature of the problem is due to the positive examples error term. Therefore, a good initialization heuristic is important in order to reach a good solution. In this section, we argue that if the positive points admit a good $K$-means clustering, then the minimizer of the function \eqref{eq:Kmeans} provides a good suboptimal solution to the problem of minimizing \eqref{eq:model}.  Our observations justify a standard initialization heuristic which was advocated in \cite{Divvala2012,Ye2013}. Moreover, we propose two other initialization schemes and experimentally evaluate their effectiveness compare to simple k-means initialization.

\subsection{Theoretic Justification of K-means Initialization}
We assume that we have found a good $K$-means clustering of the positive data, meaning that 
the average distortion error
$$
\sum_{i\in P} \min_{k} \|\mu_k - \x_i\|_2^2
$$
is small relative to the total variance of the data. In the above formula $\mu_1, \dots,\mu_K$ denote the $K$ means. We also let $k_i$ be cluster index of point $\x_i$, that is $k_i = {\rm argmin}_k \|\x_i -\mu_k\|$, we let $\delta_i = \x_i -\mu_{k_i}$ and $\epsilon = \sum_{i \in P} \|\delta_i\|$. Then we can show that
\beq
\min_\w F_{K,\lam'}(\w) \leq \min_{\w} E_{K,\lambda}(\w) \leq \min_\w F_{K,\lambda}(\w)
\label{eq:bound}
\eeq
where $\lam' = \lam - 2\epsilon$. In other words, if $\epsilon$ is much smaller than $\lambda$ then the gap between the upper and lower bounds is also small. In this case, the initialization induced by $K$-means clustering provides a good approximation to the solution of problem \eqref{eq:model}.

The right inequality in \eqref{eq:bound} holds since the objective function in problem \eqref{eq:Kmeans} specifies the assignment of each positive point to a subclassifier and
hence this objective is greater or equal to that in problem \eqref{eq:model}. The proof of the left inequality uses the fact that the hinge loss function is Lipschitz with constant $1$, namely $|L(\xi) - L(\xi')| \leq |\xi-\xi'|$. In particular this allows us to give a good approximation of the loss $\min_k (1-\lb \w_k, \x_i\rb)$ in terms of the loss of the corresponding mean, that is, $\min_k (1- \lb \w_k, \mu_{k_i}\rb)$.

The bound \eqref{eq:bound} has a number of implications. First, as $K$ increases, the gap between the upper and lower bound shrinks, hence the quality of the suboptimal solution improves. As $K$ decreases down to $K=2$ the initialization induced by $K$-means provides a more coarse approximation of problem \eqref{eq:model}, see also \cite{zhu12} for related considerations. Second, the bound suggests that a better initialization can be obtained by replacing $K$-means by $K$-medians, because the latter algorithm directly optimizes the quantity $\epsilon$ appearing in the bound.

We notice that a similar reasoning to the one presented in this section applies when the negative error term in \eqref{eq:model} is replaced by \eqref{eq:neg}. In this case, clustering the positive points, 
and subsequently training $K$ independent SVMs which separate each cluster from the set of all negative points 
yields a good suboptimal solution of the corresponding nonconvex problem, provided the distortion parameter $\epsilon$ is small relative to the regularization parameter $\lambda$. A detailed derivation is presented below:

\qsection{Derivation of Bound \eqref{eq:bound}} The right inequality readily follows by noting that the objective function in problem \eqref{eq:Kmeans} considers an {\em a-priori} assignment of each positive point to a subclassifier, hence the objective is greater or equal to that in \eqref{eq:model}. 

\begin{figure}
\begin{center}
\includegraphics[width=1\linewidth]{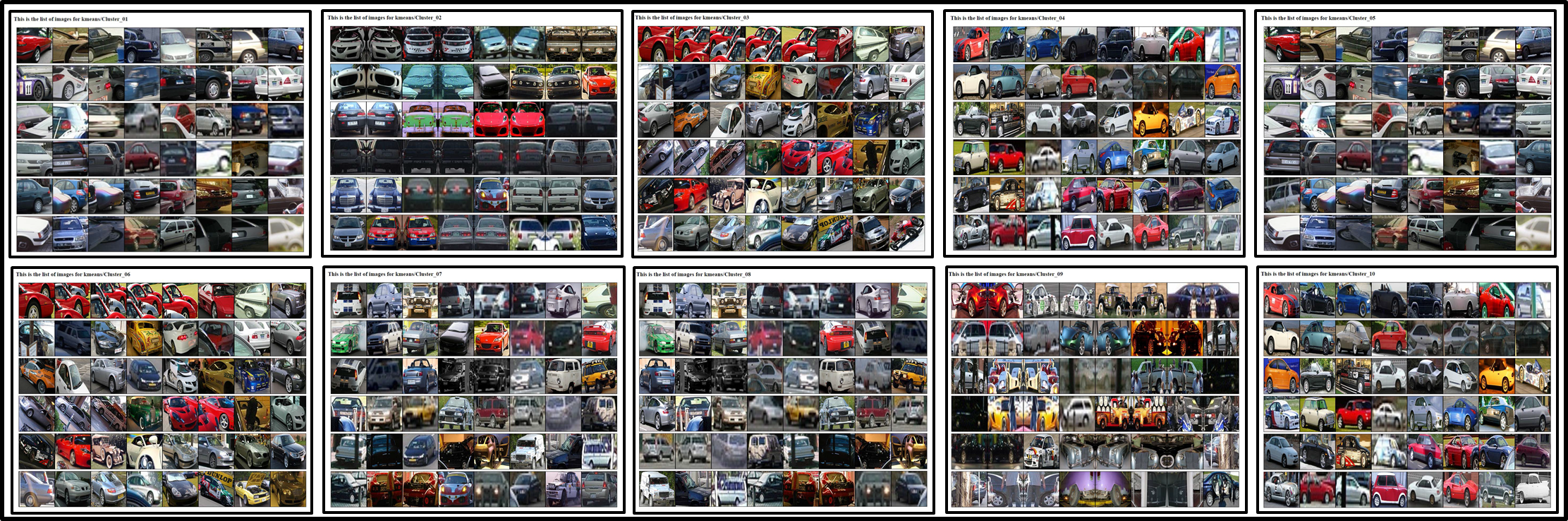}
\end{center}
  \caption{K-Means initialization.}
\label{fig:sub_kmeans}
\end{figure}

We now prove the left inequality. We consider a more general setting, in which the loss function $L$ is convex and Lipschitz. The latter property means that there exists a constant $\phi$ such that for 
every $\xi, \xi' \in \R$, $|L(\xi) - L(\xi')| \leq \phi|\xi-\xi'|$. For example the hinge loss is Lipschitz with constant $\phi=1$.

Choosing $\xi = \lb \w_k, \x_i\rb$, $\xi' = \lb \w_k ,\mu_{k_i}\rb$ and letting $\delta_i = \x_i -\mu_{k_i}$, we obtain
$$
|L(\lb\w_k, \mu_{k_i}\rb) -  L(\lb \w_k, \x_i\rb)| \leq \phi|\lb \w_k, \delta_i\rb| \leq \phi \|\w_k\|  \|\delta_i\|
$$
where the last step follows by Cauchy-Schwarz inequality.
Furthermore, using the property that, for every choice of functions $f_1,\dots,f_K$, it holds $|\min_k f_k(x) - \min_k f_k(x')| \leq \max_k |f_k(x) - f_k(x')|$, we have
$$
\max_k L(\lb\w_k, \mu_{k_i}\rb) - \max_k L(\lb \w_k, \x_i\rb) \leq \max_k | L(\lb \w_k, \mu_{k_i}\rb) - \max_k L(\lb \w_k, \x_i\rb)| \leq \phi \max_k \|\w_k\|  \|\delta_i\|.
$$
Letting $\epsilon = \phi \sum_{i \in P} \|\delta_i\|$, we conclude, for every choice of the weight vectors $\w_1,\dots,\w_K$, that
\beq
\sum_{k =1}^K 
p_k L(\lb\w_k, \mu_{k}\rb) - \epsilon
\max_k \|\w_k\|  \leq \sum_{i \in P} \min_k L(\lb \w_k, \x_i\rb) \leq \sum_{k =1}^K 
p_k L(\lb\w_k, \mu_{k}\rb) + \epsilon
\max_k \|\w_k\|  
\label{eq:pino}
\eeq
where $P$ is the set of positive points, $P_k = \{i \in P : k_i = k\}$ and $p_k = |P_k|$, that is the number of positive points in cluster $k$.

We define the surrogate convex function
\beq
S_{K,\lam}(\w) = \sum_{k=1}^K p_k L(\lb\w_k, \mu_{k}\rb) + \sum_{i \in N} L(-\max_k \lb \w_k, \x_i \rb) + \lam \max_k\|\w_k\|,
\label{eq:surrK}
\eeq
where $\w$ is a shorthand for the concatenation of $\w_1,\dots,\w_K$. Using equation \eqref{eq:pino} we obtain that
\beq
S_{K,\lam-\epsilon}(\w) \leq 
E_{K,\lam}(\w) \leq 
S_{K,\lam+\epsilon}(\w).
\label{kM-bound}
\eeq
Now using the fact that
$$
L(\lb\w_k, \mu_{k_i}\rb)  \geq L(\lb \w_k, \x_i\rb) - \|\w_k\| \|\delta_i\|
$$
and recalling equation \eqref{eq:Kmeans}, we conclude that
$$
 F_{K,\lam-2\epsilon}(\w)  =\sum_{k=1}^K \sum_{i \in P_k} L(\lb \w_k, \x_i\rb) + \sum_{i \in N} L(-\max_k \lb \w_k, \x_i\rb) + (\lam - 2\epsilon) \max_k \|\w_k\| \leq S_{K,\lambda-\epsilon}(\w).
$$
The result follows by combining the left inequality in \eqref{kM-bound} with the above inequality and minimizing over the weight vectors $\w_1,\dots,\w_K$.

\vspace{.2truecm}

\subsection{Experimental Evaluation on Subcategory Initialization} \label{sec:Init-exp}

\begin{figure}
\begin{center}
\includegraphics[width=1\linewidth]{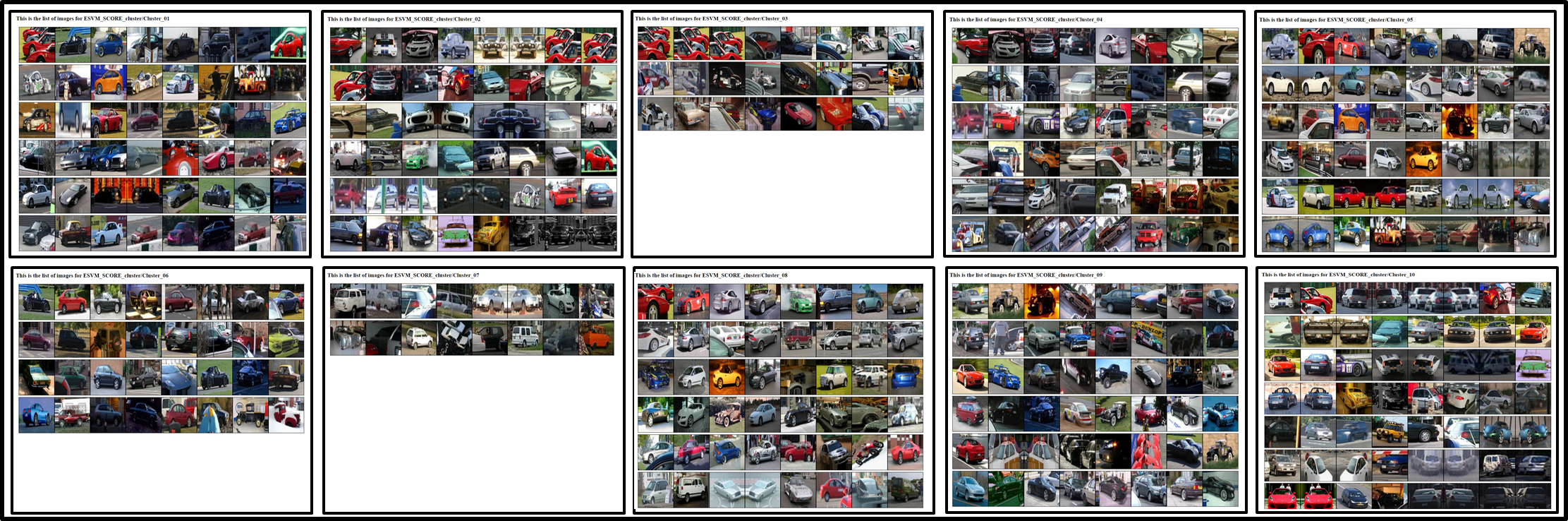}
\end{center}
  \caption{score-based initialization.}
\label{fig:sub_score}
\end{figure}

In this section, we empirically analysis the effective of different initialization approaches on the \emph{PAVIS multi-prototype dataset}. This dataset has been collected using the images and annotations of PASCAL VOC 2007 for the \emph{'car'} category. Images have been warped in order to have the same size. Then HOG features have been extracted, resulting in image descriptors of the same length. Here we report
and comment the results obtained with three different initialization approaches, respectively \emph{standard k-means}, \emph{score-based initialization} and \emph{ranking-based initialization}. The qualitative results show that k-means and ranking-based initialization are working effectively, however we cannot quantitatively evaluate them on PASCAL because the ground-truth subcategory annotation is not available for this dataset. 

\subsection{K-means Initialization}
Given 2500 positive images belonging to the car class extracted from the PASCAL VOC 2007 dataset, first we warped the images in order to have a fixed size (we adopted the experimental set-up of Divvala et al.\cite{divvala-eccvw12}). Then, we extracted the HOG features using a cell size of 8x8 and a block size of 16x16 pixels. Given the features for all the examples, we applied k-means (K=10) and then we visualized the results. Figure \ref{fig:sub_kmeans} shows some snapshots of the K-means initialization results.

\subsection{Exemplar-SVM Score-based Initialization}
Instead of using the standard Euclidean distance, here we use a similarity measure based on the output of a classifier. For each positive examples, we train an Exemplar LDA classifier (Training set
composed by one positive and all negatives samples). Then we compute a similarity matrix between the samples using the following procedure, inspired by T. Lan et al.\cite{Lan_2013_ICCV}. Given an exemplar detector, we apply it to all the other positive samples obtaining a classification score for
each one. Then, we create a directional graph using the classification scores as the weights of each edge. In order to create a non-directional graph, we compute the average of the edge scores
between each two nodes. This is used as a similarity measure between two examples in the clustering framework. The distance between two examples is 1-similarity normalized by the max of the all scores. Figure \ref{fig:sub_score} shows some snapshots of the E-SVM score-based initialization results.

\begin{figure}
\begin{center}
\includegraphics[width=1\linewidth]{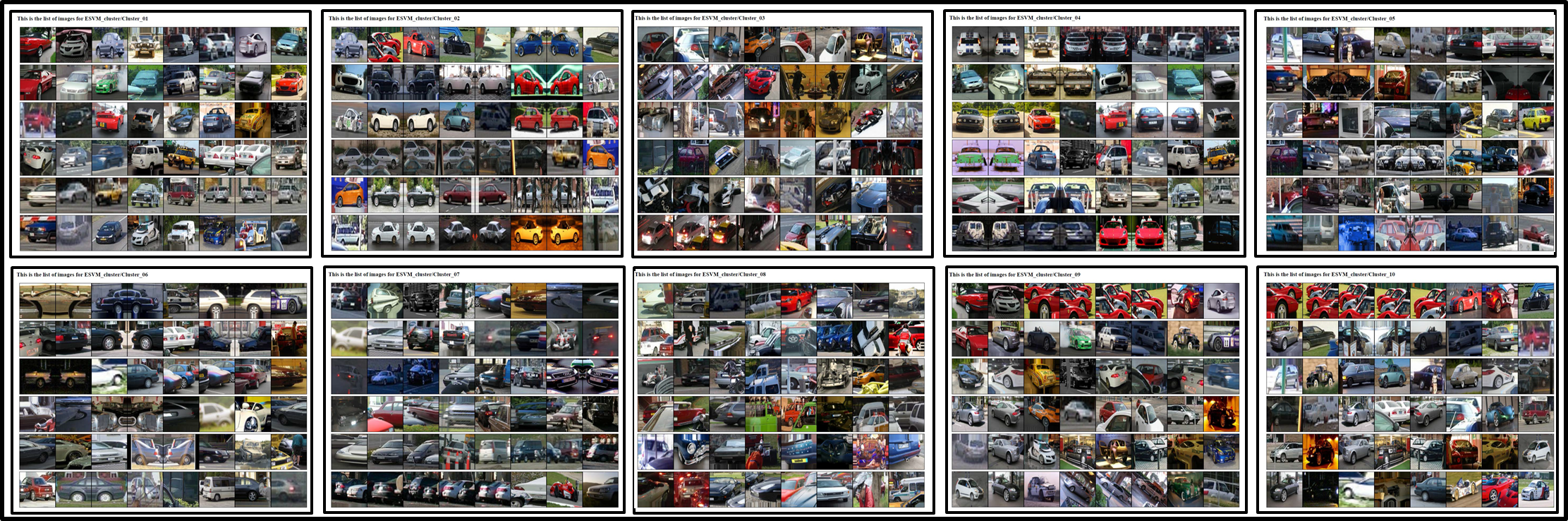}
\end{center}
  \caption{ranking-based clustering.}
\label{fig:sub_ranking}
\end{figure}

\subsection{Exemplar-SVM Ranking-based Initialization}
In this last experiment we tested another similarity measure. We train an exemplar classifier with the same procedure described above in the Score-Based Approach. The similarity measure we adopted here is different. Given an example and his exemplar classifier, we obtain the
classification scores for all the other positive examples and then we rank them in an decreasing order. We keep the first N highest scores (e.g. N = 20) and then we apply a ranking-based similarity measure we developed inspired by Divvala et al.\cite{santosh2014web}. Figure \ref{fig:sub_ranking} shows some snapshots of the E-SVM score-based initialization results.


\section{Subcategory-based Undoing Dataset Bias}

\begin{figure}[t]
\begin{center}
 \includegraphics[width=0.5\linewidth]{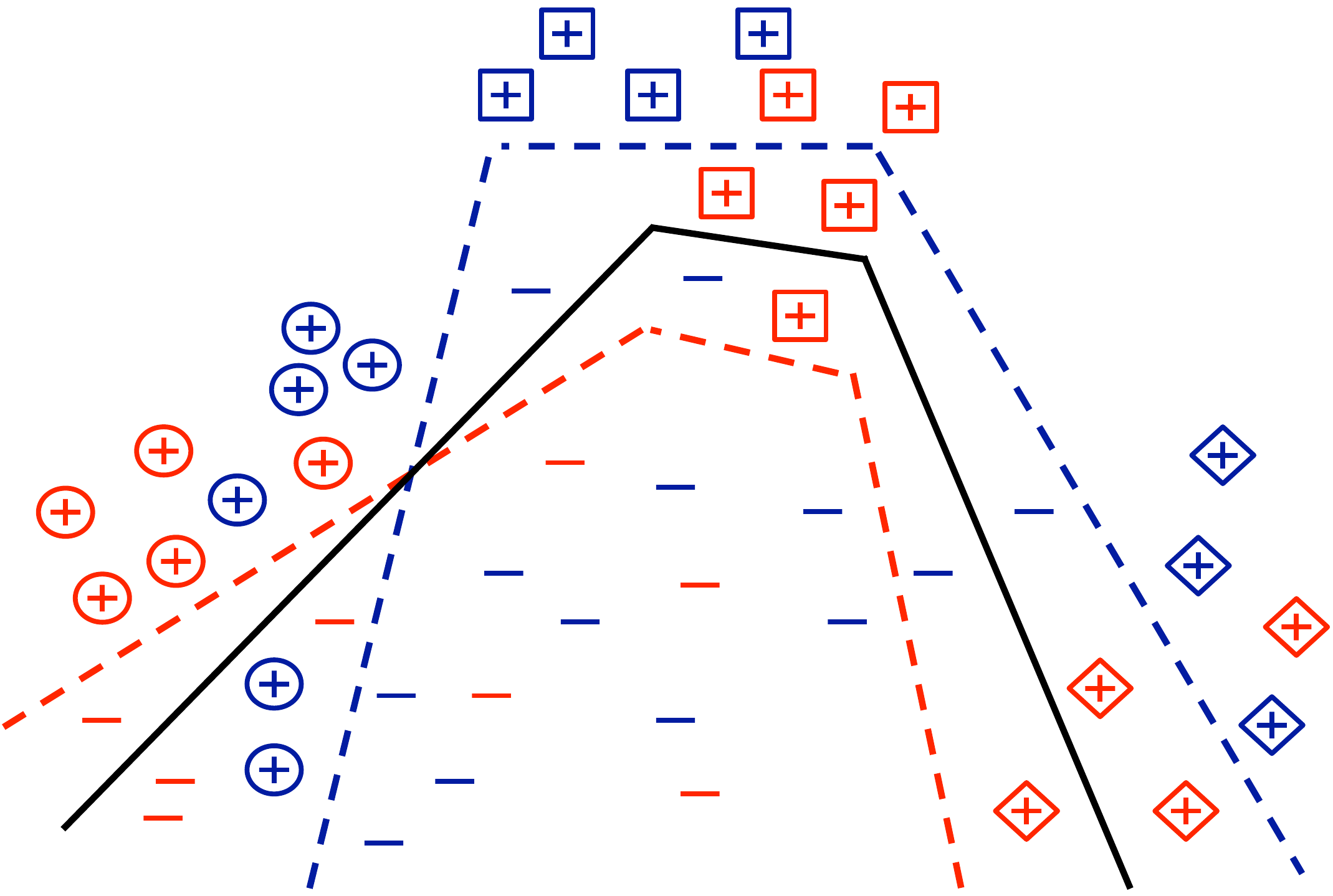}
 \vspace{.1truecm}
 \caption{Parameter sharing across datasets can help to train a better subcategory model of the visual world.  Here we have two datasets (red and blue) of a class (e.g. ``horse"), each of which is divided into three subcategories (e.g. viewpoints).  The red and blue classifiers are trained on their respective datasets. Our method, in black, both learns the subcategories and undoes the bias inherent in each dataset.}
\label{fig:1}
\end{center}
\end{figure}

Subcategory-based models using LSM offer significant improvements over training flat classifiers such as linear SVMs. Training LSMs is, however, a challenging task due to the potentially large number of 
local optima in the objective function and the increased model complexity which requires large training set sizes. Often, larger datasets are available as a collection of heterogeneous datasets. However, previous work has highlighted the possible danger of simply training a model from the combined datasets, due to the presence of bias. 
In this section, we present a model which 
jointly learns an LSM for each dataset as well as a compound LSM. The method provides a means to borrow statistical strength from the datasets while 
reducing their inherent bias. In experiments we demonstrate that the compound LSM, when tested on PASCAL, LabelMe, Caltech101 and SUN in a leave-one-dataset-out fashion, achieves an improvement over a previous SVM-based undoing bias approach \cite{Khosla2012} and a standard LSM trained on the concatenation of the datasets. This improvement provided mainly as a result of subcategory-based object modeling (as opposed to category-based model of \cite{Khosla2012}), and suggests that undoing dataset bias can be done effectively in subcategory-level (compared to category) due to presence of less visual diversity. See Figure \ref{fig:1} for an illustration of the method.

\subsection{Related Works}
Latent subcategory models (sometimes also called mixture of template models) \cite{gu2010,Divvala2012,zhu12} are a special case of DPMs \cite{Felzenszwalb2010,Girshick2013} and structured output learning \cite{tsochantaridis2005large}. Closely related methods have also been considered in machine learning under the name of multiprototype models or multiprototype support vector machines \cite{Aiolli2005}, as well as in optimization \cite{Magnani2009}. An important issue in training these models is the initialization of the subclassifier weight vectors. This issue has been addressed in \cite{gu2010,Divvala2012}, where clustering algorithms such as $K$-means 
is used to cluster the positive class and subsequently independent SVMs are trained to initialize the weight vectors of the subclassifiers. In Section \ref{sec:Init}, we observed that $K$-means clustering can be justified as a good initialization heuristic when the positive class admits a set of compact clusters. In this section, we, in addition, discuss how this initialization can be adapted to our undoing bias setting.

We note that other initialization heuristics are presented in section \ref{sec:Init-exp} and also discussed in \cite{Girshick2013}. Furthermore other interesting latent subcategory formulations are presented in \cite{hoai13} and \cite{zhucapturing}. While we do not follow these works here, our method could also be extended to those settings. 

Most related to our work is the work by Khosla {\em et al.} \cite{Khosla2012}, which consider jointly training multiple linear max-margin classifiers from corresponding biased datasets. The classifiers pertain to the same classification (or detection) task (e.g. ``horse classification'') but each is trained to perform well on a specific ``biased'' dataset. The framework they consider resembles the regularized multitask learning method of Evgeniou and Pontil \cite{Evgeniou2004} with the important difference that the common weight vectors (``visual world'' classifier) is directly constrained to fit the union of all the training datasets well. A key novelty of our approach is that we enhance these methods by allowing the common vector and bias vectors to be LSMs. 
We show experimentally that our method improves significantly over both \cite{Khosla2012} and a standard LSM trained on the concatenation of all datasets.


At last, we note that our model is different from the supervised domain adaptation methodology in \cite{saenko2010adapting,kulis2011you}, which focuses on learning transformations between a source and a target domain. A key difference compared to these methods is that they require labels in the target domain, whereas our setting can be tested on unseen datasets, see also \cite{Khosla2012} for a discussion. Other related works include \cite{hoffman2012discovering} and \cite{salakhutdinov2011learning}.

\subsection{The Proposed Framework}
\label{sec:method}
In this section, we extend LSMs described in Section \ref{sec:lsm} to a multitask learning setting. Although we focus on LSMs, the framework we present in this work extends in a natural manner to training multiple latent part-based models.

Following \cite{Khosla2012} we assume that we have several datasets pertaining to the same object classification or detection task. Each dataset is collected under specific conditions and so it provides a biased view of the object class (and possibly the negative class as well). For example, if the task is people classification one dataset may be obtained by labelling indoor images as people / not people, whereas another dataset may be compiled outdoors, and other datasets may be generated by crawling images from internet, etc. Although the classification task is the same across all datasets, the input data distribution changes significantly from one dataset to another. Therefore a classifier which performs well on one dataset may perform poorly on another dataset. Indeed, \cite{Khosla2012} empirically observed that training on the concatenation of all the datasets and testing on a particular dataset is outperformed by simply training and testing on the same dataset, despite the smaller training set size.

In the sequel, we let $T$ be the number of datasets and for $t \in \{1,\dots,T\}$, we let $m_t$ be the sample size in training dataset $t$ and let $\calD_t = \{(x_{t1},y_{t1}),\dots,(x_{tm_t},y_{tm_t}) \} \subset \R^d \times \{-1,1\}$ be the corresponding data examples. We assume that the images in all the datasets have the same representation so the weight vectors can be compared by simply looking at their pairwise Euclidean distance.

\qsection{Undoing Bias SVM}
In \cite{Khosla2012}, the authors proposed a modified version of the multitask learning framework in \cite{Evgeniou2004} in which the error term includes an additional term measuring the performance of a compound model (visual world classifier) on the concatenation of all the datasets. This term is especially useful when testing the compound model on an ``unseen'' dataset, a problem we return upon in the sequel. Specifically, in \cite{Khosla2012} the authors learn a set of linear max-margin classifiers, represented by weight vectors $\w_t \in \R^d$ for each dataset, under the assumption that the weights are related by the equation $\w_t = \w_0 + \vvcvpr_t$, where $\w_0$ is a compound weight vector (which is denoted as the {\em visual world} weight in \cite{Khosla2012}). The weights $\w_0$ and $\vvcvpr_1,\dots,\vvcvpr_T$ are then learned by minimizing a regularized objective function which leverages the error of the biased vectors on the corresponding dataset, the error of the visual world vector on the concatenation of the datasets and a regularization term which encourages small norm of all the weight vectors.

\qsection{Undoing Bias LSM}
\label{sec:UB-LSM}
We now extend the above framework to the latent subcategory setting. We let $\w_t^1,\dots,
\w_t^K \in \R^d$ be the weight vectors for the $t$-th dataset, for $t=1,\dots,T$. For simplicity, we assume that the number of subclassifiers is the same across the datasets, but the general case can be handled similarly. Following \cite{Evgeniou2004,Khosla2012}, we assume that the weight vectors representative of the $k$-th subcategory across the different datasets are related by the equation
\beq
\w_t^k = \w_0^k + \vvcvpr_t^k
\label{eq:krelat}
\eeq
for $k=1,\dots, K$ and $t=1,\dots,T$. The weights $\w_0^k$ are shared across the datasets and the weights $\vvcvpr_t^k$ capture the bias of the $k$-th weight vector in the $t$-th dataset. We learn all these weights by minimizing the objective function
\begin{eqnarray}
&&C_1 \sum_{t=1}^T \sum_{i=1}^{m_t} L(y_{ti} \max_{k} \lb \w^k_0+ \vvcvpr_t^k , \x _{ti}\rb) ~~~~~~~~
\label{eq:K-MTL1}
 \\
&+& C_2  \sum_{t=1}^T \sum_{i=1}^{m_t}  L(y_{ti} \max_{k} \lb \w^k_0, \x_{ti}\rb)
 \label{eq:K-MTL2} \\
&+&  \sum_{k=1}^K \Big(\|\w^k_0\|^2 + \rho \sum_{t=1}^T  \|\vvcvpr_t^k\|^2\Big).
\label{eq:K-MTL3}
\end{eqnarray}
In addition to the number of subclassifiers $K$, the method depends on three other nonnegative hyperparameters, namely $C_1$, $C_2$ and $\rho$, which can be tuned on a validation set. 
Note that the method reduces to that in \cite{Khosla2012} if $K=1$ and to the one in \cite{Evgeniou2004} if $K=1$ and $C_2=0$.  Furthermore our method reduces to training a single LSM on the concatenation of all datasets if $C_1 = 0$.

The parameter $\rho$ plays an especially important role: it controls the extent to which the datasets are similar, or in other words the degree of bias of the datasets. Taking the limit $\rho \rightarrow \infty$ (or in practice setting $\rho \gg 1$) eliminates the bias vectors $\vvcvpr_t^k$, so we simply learn a single LSM on the concatenation of all the datasets, ignoring any possible bias present in the individual datasets. Conversely, setting $\rho=0$ we learn the bias vectors and visual world model independently. The expectation is that a good model lies at an intermediate value of the parameter $\rho$, which encourages some sharing between the datasets is necessary.

\qsection{Implementation}
A common method used to optimize latent SVM and in particular LSMs is stochastic gradient descent (SGD), see for example \cite{Shalev-Shwartz2011}. 
At each iteration we randomly select a dataset $t$ and a point $\x_{ti}$ from that dataset and update the bias weight vector $\vvcvpr_{t}^k$ and $\w_0^k$ by subgradient descent. We either train the SGD method with a fixed number of epochs or set a convergence criterion that checks the maximum change of the weight vectors. Furthermore, we use the adapting cache trick: if a point is correctly classified by at least 
two base and bias pairs $(\w_0^k, \w_t^k)$ 
then we give it a long cooldown. This means that the next 5 or 10 times the point is selected, we instead skip it. A similar process is used in \cite{Felzenszwalb2010,Khosla2012} and we verified empirically that this results in improved training times, without any significant loss in accuracy.

\qsection{Initialization}\label{sec:initalization}
It is worth discussing how the weight vectors are initialized.  First, we group all the positive points across the different datasets and run $K$-means clustering. Let $P_k$ be the set of points in cluster $k$, let $P_{t,k}$ be the subset of such points from dataset $t$ and let $N_t$ be the set of negative examples in the dataset $t$. For each subcategory $k \in \{1,\dots,K\}$ we initialize the corresponding weight vectors $\w_0^k$ and $\vvcvpr_1^k,\dots,\vvcvpr_T^k$ as the solution obtained by running the undoing datasets' bias method from \cite{Khosla2012}, with training sets $\calD_t = \{(\x_{ti},y_{ti}) : i \in P_{t,k} \cup N_t\}$. We then iterate the process using SGD for a number of epochs (we use $100$ in our experiments below). 

Our observations in Section \ref{sec:Init} extend in a natural way to the undoing bias LSMs setting. The general idea is the same: if the data admit a good $K$-means clustering then the initialization induced by $K$-means provides a good suboptimal solution of the problem. We have experimentally verified that the improvement offered by this initialization over a random choice is large. Table \ref{tab:0} reports the performance of the our method after 100 epochs of SGD starting with or without the $K$-means initialization. Average performance and standard deviation are reported over $30$ trials. As it can be seen $K$-means initialization offers a substantial improvement.

\begin{table}[t]
\begin{center}
\begin{tabular}{|c||c|c|c|}
\hline 
Object & Bird &  Car  & People \tabularnewline
\hline 
\hline 
$K$-means & 33.8 $\pm 0.4$ & $65.8 \pm 0.4$ & $67.5\pm0.2$ \tabularnewline \hline
Random & 29.4 $\pm 0.6$ & $61.3 \pm 0.5$ &  $64.7\pm0.5$ \tabularnewline
\hline 
\end{tabular}{}
\end{center}
\caption{Average performance (over 30 runs) of our method with or without $K$-means initialisation for two object classification tasks.}
\label{tab:0}
\end{table}

\qsection{Effect of Initialization in Undoing Bias LSM} As we noted earlier in this section, the initialization induced by $K$-means clustering can be extended in a natural way to the undoing bias LSM setting. We first run $K$-means on the aggregate set of positive points from all datasets. We let $P_{t,k}$ be the subset of positive points in dataset $t$ which belong to cluster $k$ and let $N_t$ be the set of negative points in the same dataset. For each subcategory $k \in \{1,\dots,K\}$, we initialize the corresponding weight vectors $\w_0^k$ and $\vvcvpr_1^k,\dots,\vvcvpr_T^k$ as the solution obtained by running the undoing bias method in \cite{Khosla2012}, with training sets $\calD_t = \{(\x_{ti},y_{ti}) : i \in P_{t,k} \cup N_t\}$. Specifically, for each $k$, we solve the problem 

$$
\sum_{t=1}^T  \sum_{i \in P_{t,k} \cup N_t} \Big[  C_1 L(y_{ti} \lb \w_0+ \vvcvpr_t , \x _{ti}\rb)  + C_2 L(y_{ti} \lb \w_0, \x_{ti}\rb)\Big] +
\|\w_0\|^2 + \rho \sum_{t=1}^T  \|\vvcvpr_t\|^2.
$$
We then attempt to minimize the objective function formed by equations \eqref{eq:K-MTL1}--\eqref{eq:K-MTL3} with SGD for a number of epochs. The computation of a subgradient for this objective function is outlined in Algorithm \ref{algo:S} below.

Using arguments similar to those outlined above we can show that this initialization gives a good approximation to the minimum of the non-convex objective \eqref{eq:K-MTL1}--\eqref{eq:K-MTL3}, provided the average distortion error $\sum_{t}\sum_{i \in P} \|\delta_{ti}\|$ is small, where we let $\delta_{ti} = \min_{k} \|\x_{ti} - \mu_k\|$.


Table \ref{table:accuracy} illustrates the importance of this initialization process, using a fixed parameter setting over 30 runs, in a seen dataset setting.   
The first row shows the performance (average precision) of a random choice of $\w_0$ and $\vvcvpr_1, \dots, \vvcvpr_T$.  
The second row shows the performance of our method starting from this random initialization. 
The third row shows the performance of the $K$-means induced initilization reviewed above.
Finally, the fourth row is our method. As we see, $K$-means based initialization on its own already provides a fair solution. In particular, for ``chair'', ``dog'' and ``person'' there is a moderate gap between the performance of $K$-means based initialization and 
$K$-means followed by optimization. Furthermore, in all cases $K$-means followed by optimization provides a better solution than 
random initialization followed by optimization.

\begin{table*}[t]
\begin{adjustbox}{width=1\textwidth,center=\textwidth}
\begin{tabular}{c||c|c|c|c|c}
\hline 
Test & bird  & car & chair & dog & person   \tabularnewline
\hline 
\hline 
random & 3.9 (0.1) & 17.9 (0.1) & 7.3 (0.1)  & 3.6 (0.3) & 21.5 (0.1)\tabularnewline
\hline 
random followed by optimization & 29.4 (0.6) & 61.3 (0.5) & 34.5 (0.1) & 27.7 (0.8) & 64.7 (0.5) \tabularnewline
\hline
$K$-means & 18.3 (0.3)  & 51.2 (0.6)  & 30.2 (0.2)  & 24.3 (0.3)  & 61.2 (0.3) \tabularnewline
\hline
$K$-means followed by optimization & 33.8 (0.4)  & 65.8 (0.4)  & 35.2 (0.2)  & 31.4 (0.3) & 67.5 (0.2)  \tabularnewline
\hline
\end{tabular}{}
\end{adjustbox}
\caption{Effect of initialization on AP on different image classification problems. Top to bottom: random initialization, random initialization and optimization (100 epochs of SGD), $K$-means initialization, $K$-means initialization and optimization (100 epochs of SGD). }
\label{table:accuracy}
\end{table*}


\begin{center}
\fbox{\parbox{1\linewidth}{
\begin{algorithm}[H] 

 \caption{Computation of subgradient for the objective function \eqref{eq:K-MTL1}--\eqref{eq:K-MTL3}.}
\label{algo:S}
 	\SetAlgoLined

      \For{$k \leq K$ } {

      	\uIf{$k = \argmax_{j} \langle \w^j_0, \x_{ti} \rangle$ {\bf and} $k = \argmax_{j} \langle \w^j_0 + \vvcvpr_t^j, \x_{ti} \rangle$ {\bf and} $y_{ti} \langle \w_0^k, \x_{ti} \rangle \leq 1$ {\bf and} $y_{ti} \langle \w_0^k + \vvcvpr_t^k, \x_{ti} \rangle \leq 1$ }{


      	$\partial_{\w_0^k}J = - C_1 y_{ti} \x_{ti} - C_2 y_{ti} \x_{ti} + \w_0^k$

		}      	

      	\uElseIf{$k = \argmax_{j} \langle \w^j_0, \x_{ti} \rangle$  {\bf and} $y_{ti} \langle \w_0^k, \x_{ti} \rangle \leq 1$}{


      	$\partial_{\w_0^k}J = - C_2 y_{ti} \x_{ti} + \w_0^k$  

		}      	

      	\uElseIf{$k = \argmax_{j} \langle \w^j_0 + \vvcvpr_t^j, \x_{ti} \rangle$ {\bf and} $y_{ti} \langle \w_0^k + \vvcvpr_t^k, \x_{ti} \rangle \leq 1$ }{     


      	$\partial_{\w_0^k}J = - C_1 y_{ti} \x_{ti}  + \w_0^k$

      	}

      	\uElse{


      	$\partial_{\w_0^k}J = \w_0^k$      	

      	}

	  }

	  \mbox{}\\ 

	  \For{$k \leq K$} {

	  	\uIf{$k = \argmax_{j} \langle \w^j_0 + \vvcvpr_t^j, \x_{ti} \rangle$ {\bf and} $y_{ti} \langle \w_0^k + \vvcvpr_t^k, \x_{ti} \rangle \leq 1$ }{


	  	$\partial_{\vvcvpr_t^k} = - C_1 y_{ti} \x_{ti} + \rho \vvcvpr_t^k $

	  	}

	  	\uElse{	  	


	  	$\partial_{\vvcvpr_t^k} = \rho \vvcvpr_t^k$

	  	}
subsection
	  }
	  \end{algorithm}
}}
\end{center}

\subsection{Experiments}
\label{sec:exp}

In this section, we present an empirical study of the proposed method.
The goal of the experiments is twofold. On the one hand, we 
investigate the advantage offered by our method over standard LSMs trained on the union (concatenation) of all the datasets. Intuitively, we expect our method to learn a better set of visual world subclassifiers since it filters out datasets bias. On the other hand, we compare our method to the ``undoing bias'' method in \cite{Khosla2012}, where each dataset is modelled as a linear SVM classifier (so no subclassifiers are learned in this case). As we already noted, both methods are special cases of ours for a certain choice of the hyperparameters.

In the experiments, we focus on object classification tasks as this allows us to directly compare with the results in \cite{Khosla2012} using the publicly available features provided by the authors\footnote{See the link {\em http://undoingbias.csail.mit.edu/}.}. 
However the method can also be employed for detection experiments. 
Following the setting in \cite{Khosla2012} we employ four datasets: Caltech101  \cite{fei2007learning}, LabelMe  \cite{labelme}, PASCAL2007 \cite{pascal} and SUN09 \cite{choi2010}. 
We use the bag-of-words representation provided by \cite{Khosla2012}. It is obtained by extracting SIFT descriptors at multiple patches, followed by local linear coding and a 3-level spatial pyramid with linear kernel. Performance of the methods is evaluated by average precision (AP).

We use the same training and test splits provided in \cite{Khosla2012}. Furthermore, to tune the model parameter $C_1,C_2$ and $\rho$, we used 75\% of training data of each dataset for actual training and the remaining 25\% for validation. We use the following parameter range for validation: $\rho = 10^r$, for $r$ ranging from $-9$ to $4$ with a step of $1$ and $C_1,C_2 = 10^{r}$, for $r$ ranging from $-9$ to $4$ with a step of $.5$. 

In our experiments, the number of subclassifiers $K$ is regarded as a free hyperparameter chosen from the test set and we try values from $1$ and up to $10$.  Although related work by \cite{Divvala2012} recommends using values of $K$ up to $50$, they consider detection tasks. However, as we show below, smaller values of $K$ provide the best results for classification tasks, since in the features employed in this case are extracted from larger images which are often dominated by the background rather than the object itself. This makes it more difficult to learn finer subcategories. 

We test the methods in two different scenarios, following the ``seen dataset'' and ``unseen dataset''  settings outlined in \cite{Khosla2012}. In the first scenario we test on the same datasets used for training. The aim of this experiment is to demonstrate that the visual world model works better than a single model trained on the concatenation of the datasets, and it is competitive with a specific model trained only on the same domain. Furthermore, we show the advantage over setting $K=1$. 
In the second scenario, we test the model on a new dataset, which does not contribute any training points. 
Here our aim is to show that the visual world model improves over just training a model on the concatenation of the training datasets as well as the visual world model from \cite{Khosla2012}. We discuss the results in turn.

\begin{table*}[t]
\begin{adjustbox}{width=1\textwidth,center=\textwidth}
\begin{tabular}{c||c|c|c|c|c|c|c}
\hline 
Test & $\w_{Pas}$  & $\w_{Lab}$ & $\w_{Cal}$& $\w_{SUN}$ & $\w_{\rm vw}$ & ${\rm Aggregate}$ & ${\rm Independent}$ \tabularnewline
\hline 
\hline 
Pas &  ${66.8}~(64.8)$ &  $55.6~(50.5)$ &  $56.3~(54.2)$ &  $65.9~(51.2)$&  $66.5~(57.0)$&  $63.7~(57.4)$ & ${\bf 67.1}~(65.9)$\tabularnewline
\hline 
Lab &  $ 73.1~(68.8)$ &  ${\bf 75.2}~(72.9)$ &  $75.0~(71.2)$ &  $71.6~(73.3)$&  $75.1~(72.4)$&  $72.9~(72.9)$ & $72.4~(71.7)$ \tabularnewline
\hline 
Cal &  $ 96.5~(94.8)$ &  $97.5~(94.6)$  &  $98.2~({\bf 99.7})$&  $97.6~(95.6)$&  $98.0~(98.9)$&  $98.9~(97.0)$ & $98.8~(99.4)$  \tabularnewline
\hline 
SUN & $ 57.2~(40.1)$ &  $57.6~(46.5)$ &  $57.7~(50.2)$&  $58.0~({\bf 59.6})$&  $57.8~(54.1)$&  $53.9~(54.0)$ & $58.9~(55.3)$ \tabularnewline
\hline  \hline
Average& $73.4~(67.1)$ &  $71.2~(66.2)$ &  $71.8~(68.8)$&  $73.3~(69.9)$ &  ${\bf 74.5}~(70.6)$ &  $72.4~(70.3)$ & $74.3~(73.0)$  \tabularnewline
\hline 
\end{tabular}{}
\end{adjustbox}
\caption{Average precision (AP) of ``car classification'' on seen datasets for our method ($K=2$) and, within brackets, AP for the undoing bias method.}
\label{table:table1_cvpr}
\end{table*}

\begin{figure*}[t]
\begin{center}
 \includegraphics[width=1.0\linewidth]{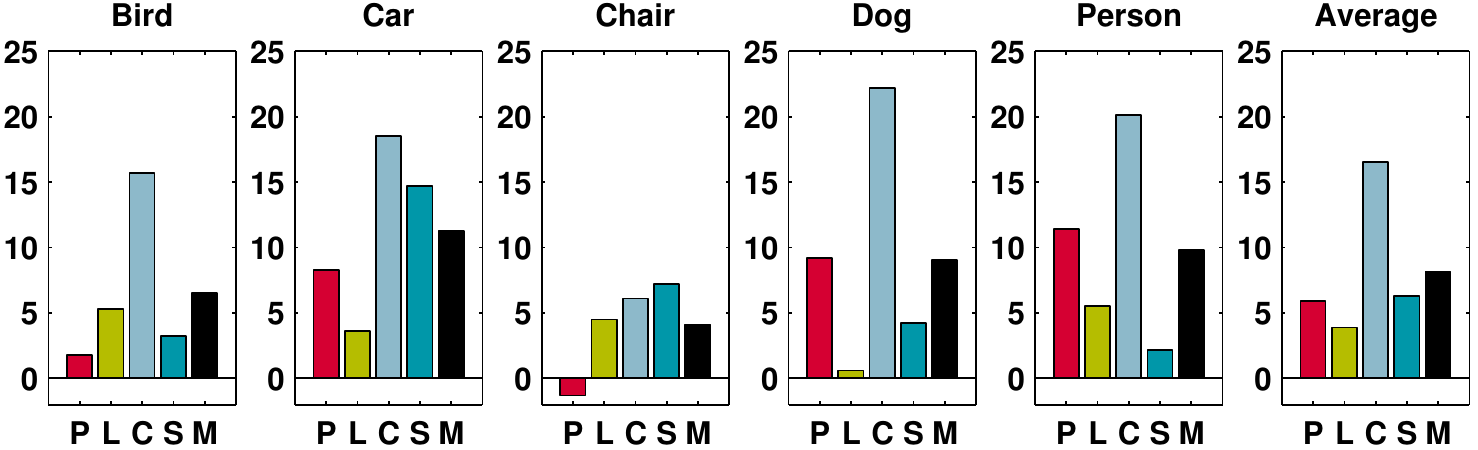}
\end{center}
\vspace{.1truecm}
  \caption{Relative improvement of undoing dataset bias LSM vs. the baseline LSM trained on all datasets at once (aggregated LSM).  On all datasets at once (P: PASCAL, L: LabelMe, C: Caltech101, SUN: M: mean).}
\label{fig:unseen}
\end{figure*}

\qsection{Testing on Seen Datasets}
In the first set of experiments, we test our method on ``car classification'' datasets.
Results are reported in Tables \ref{table:table1_cvpr}
The main numbers indicate the performance of our method, while within brackets we report performance for $K=1$, which corresponds to the undoing bias method in \cite{Khosla2012}. As noted above, in this set of experiments we test the method on the same datasets used during training 
(by this we mean that the training and test sets are selected within the same domain). For this reason and since the datasets are fairly large, we do not expect much improvement over training on each dataset independently (last column in the table). However the key point of the table is that training a single LSM model on the union of all datasets (we call this the aggregate model in the tables) yields a classifier which neither performs well on any specific dataset nor does it perform well on average. In particular, the performance on the ``visual world'' classifier is much better than that of the aggregated model. 
This finding, due to dataset bias, is in line with results in \cite{Khosla2012}, as are our results for $K=1$.
Our results indicate that, on average, using a LSM as the core classifier 
provides a significant advantage over using single max-margin linear classifier. The first case corresponds to a variable number of subclassifiers, the second case corresponds to $K=1$. This is particularly evident by comparing the performance of the two visual world classifiers (denoted by $\w_{vw}$ in the tables) in the two cases.



\begin{figure*}[t]
\begin{center}
 \includegraphics[width=1\linewidth]{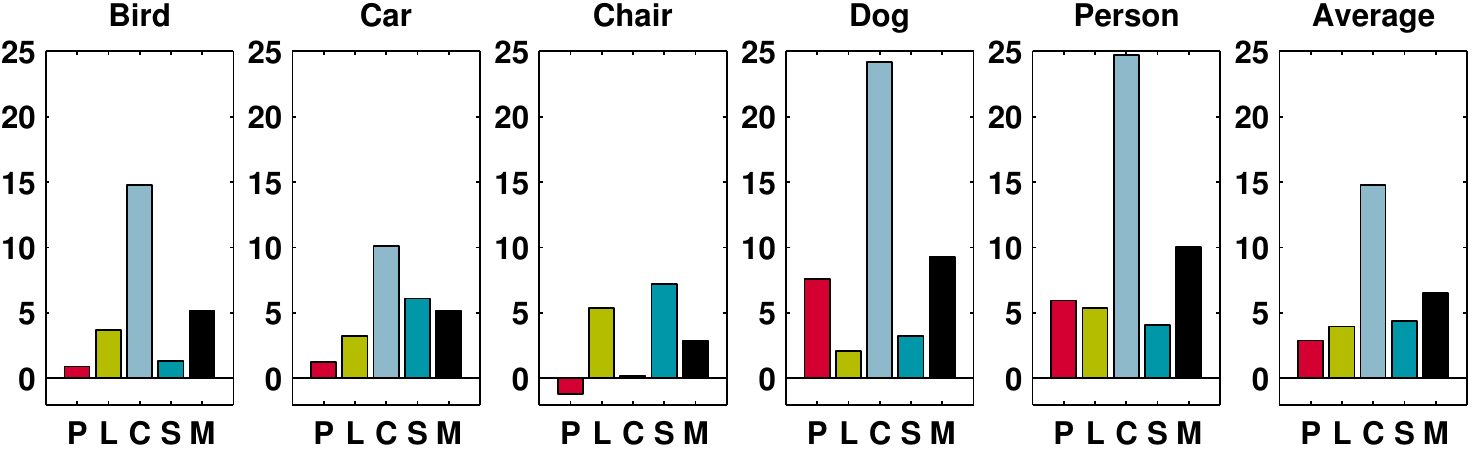}
\end{center}
\vspace{.1truecm}
  \caption{Relative improvement of undoing dataset bias LSM vs. undoing bias SVM \cite{Khosla2012}.}
\label{fig:unseen2}
\end{figure*}

\qsection{Testing on Unseen Datasets}
In the next set of experiments, we train our method on three out of four datasets, retain the visual world classifier and test it on the dataset left out during training. Results are reported in Figure \ref{fig:unseen}, where we show the relative improvement over training a single LSM on all datasets (aggregated LSM), and Figure \ref{fig:unseen2}, where we show the relative improvement of our method over the method in \cite{Khosla2012}. Overall our method gives an average improvement of more than $\impK\%$ over the aggregated LSMs and an average improvement of more than $\impB\%$ over \cite{Khosla2012}. On some datasets and objects the improvement is much more pronounced than others, although overall the method improves in all cases (with the exception of ``chair classification'' on the PASCAL dataset, where our method is slightly worse than the two baselines). Although our method tends to improve more over aggregated LSM than undoing bias SVMs, it is interesting to note that on the Caltech101 ``person'' or ``dog'' datasets, the trend reverses. Indeed, these object classes contain only one subcategory (for ``person'' a centered face image, for ``dogs'' only Dalmatian dogs) hence when a single subcategory model is trained on the remaining three datasets a more confused classifier is learned.

\begin{figure*}
\begin{center}
\includegraphics[width=140mm]{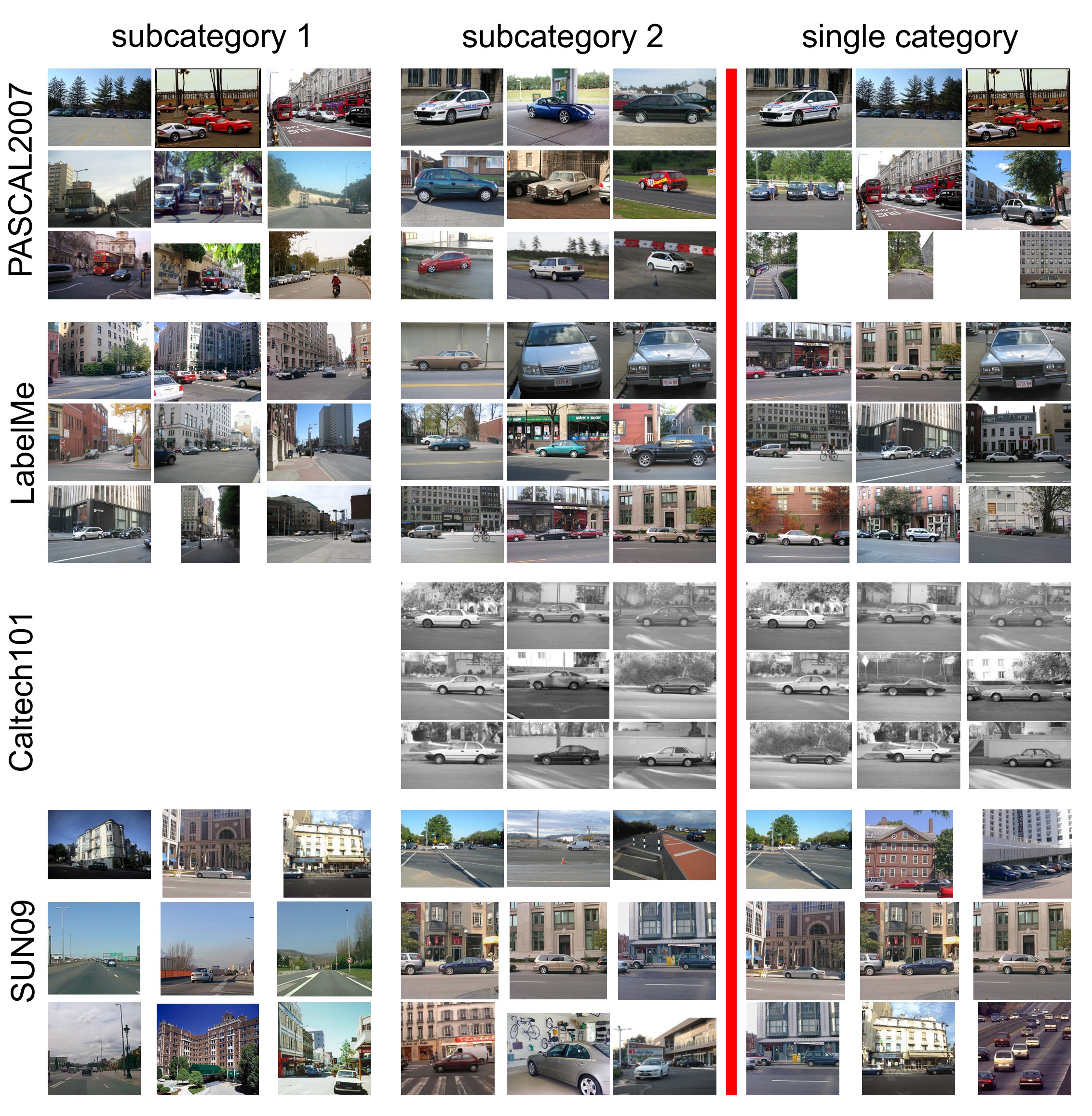}
\caption{Left and center, the top scoring images for visual world subclassifiers $\w_0^1$ and $\w_0^2$ using our method, respectively.  Right, the top scoring image for single category classifier $\w_0$ from \cite{Khosla2012}.  }
\label{fig:score}
\end{center}
\end{figure*}

%
To further 
illustrate the advantage offered by the new method, we display in Figure \ref{fig:score} the car images which achieved a top score on each of the four datasets for our method and the visual world classifier from \cite{Khosla2012}.
In our case we use $K=2$ subclassifiers because this gives the best performance on this object class. Note that among the two visual world subclassifiers, $\w_0^1$ and $\w_0^2$, 
the former tends to capture images containing a few cars of small size with a large portion of background, while the latter  concentrates on images which depict a single car occupying a larger portion of the image. 
This effect is especially evident on the PASCAL and LabelMe datasets. 
On Caltech101, the 
first subclassifier is empty, which is not surprising as this dataset contains only images with well centered objects, so no cars belong to the first discovered subcategory. Finally the SUN dataset has fewer images of large cars and contributes less to the second subcategory. Note, however, that we still 
find images of single cars although of smaller size. The left portion of Figure \ref{fig:score} reports similar scores for the visual world classifier trained in \cite{Khosla2012} $(K=1)$. In this case we see that images of the two different types are present among the top scores, which indicates that the model is too simple and underfits the data in this example.

\subsection{Discussion and Conclusion}
\label{sec:conc}

We presented a method for learning latent subcategories in presence of multiple biased datasets. Our approach is a natural extension of previous work on multitask learning to the setting of latent subcategory models (LSMs). In addition to the number of subclassifiers, the model depends upon two more hyperparameters, which control the fit of the visual world LSM to all the datasets and the fit of each biased LSM to the corresponding dataset.

In experiments, we demonstrated that our method provides significant improvement over both standard LSMs and previous undoing bias methods based on SVM. Both methods are included in our framework for a particular parameter choice and our empirical analysis indicate our model achieves the best of both worlds:  it mitigates the negative effect of dataset bias and still reaps the benefits of learning object subcategories.

In future work it would be valuable to extend ideas presented here to the setting of DMPs, in which the subclassifiers are part-based models. 
Furthermore, our observations on $K$-means initialization may be extended to other clustering schemes and other LSMs such as those described in \cite{hoai13} and \cite{zhucapturing}.
Finally, learning LSMs across both biased datasets and different object classes provides another important direction of research.


\section{Subcategory-aware Discriminative Patches} \label{sec:web_patch}
In this section, we study discovering and learning a set of patch models to be employed as mid-level image representation. For this purpose, we discover a set of discriminative patches for each \emph{subcategory} and train them in a \emph{webly-supervised} setting. The main reason for doing the patch discovery in a subcategory-aware fashion is to limit the model to be trained for a particular appearance of an object (as opposed to whole object category). Limiting the scope to subcategory enables us to be able to use the large amount of unlabeled images available on Internet in a webly-supervised way. We follow the same hypothesis we had for undoing datasets bias which, in the case of web-supervision, is based on the assumption that: \emph{the bias of Internet images are less for the subcategories compare to its category.}

\subsection{Overview} \label{sec:Overview}

Most related to our work is the work by Divvala {\em et al.} \cite{santosh2014web}, which introduced a webly-supervised framework for visual concept learning (we refer this work as LEVAN later in this thesis). It is a fully-automated system that learns everything visual about any concept, by processing lots of books and images on the web. Moreover, this method provides us with a set of Internet images for each subcategory. All images are annotated with the bounding box generated by the initial model. This gigantic amount of data (in order of ten million images) is collected thanks to the web crawler of LEVAN. We refer to this set of images as PASCAL 9990 Dataset later in this section. Different from this method which trains a single monolithic model for each object subcategories, we on the other hand look at the inside the object bounding box, and train a set of discriminative patches which can be shared across object subcategories. It provides us with a better subcategory models and a more efficient inference.\ignore{We employ LEVAN to discover the subcategories of an object category as well as train an initial monolithic models for that.}

In this section, we aim at discovering and training discriminative patches for $g$ object categories of $C = \left \{ c_{1},...,c_{g} \right \}$. For each category $c_i$, we start by defining a set of object subcategories $S = \left \{ s_{1},...,s_{m} \right \}$ which have been discovered and initialized in a webly-supervised fashion. Then, for each subcategory $s_j$ belongs to category $c_i$, we find a set of discriminative patch $P = \left \{ p_{1},...,p_{n} \right \}$, where n is the number of patches per each subcategory. In this section, we first explain briefly how we adopt LEVAN to discover and train the initial subcategory models (Section \ref{sec:levan}). The pipeline is then explained in the subcategory-aware fashion by introducing the details of patch discovery and training (Section \ref{sec:Subcategory-aware}). Next, we describe merging monolithic and discriminative patch models to create the final subcategory model. Finally, we explain the model inference and how the whole pipeline has been used for the task of unsupervised object detection. (Section \ref{sec:merge})

 \begin{figure*}[t]
 \begin{center}
   \includegraphics[width=1\linewidth]{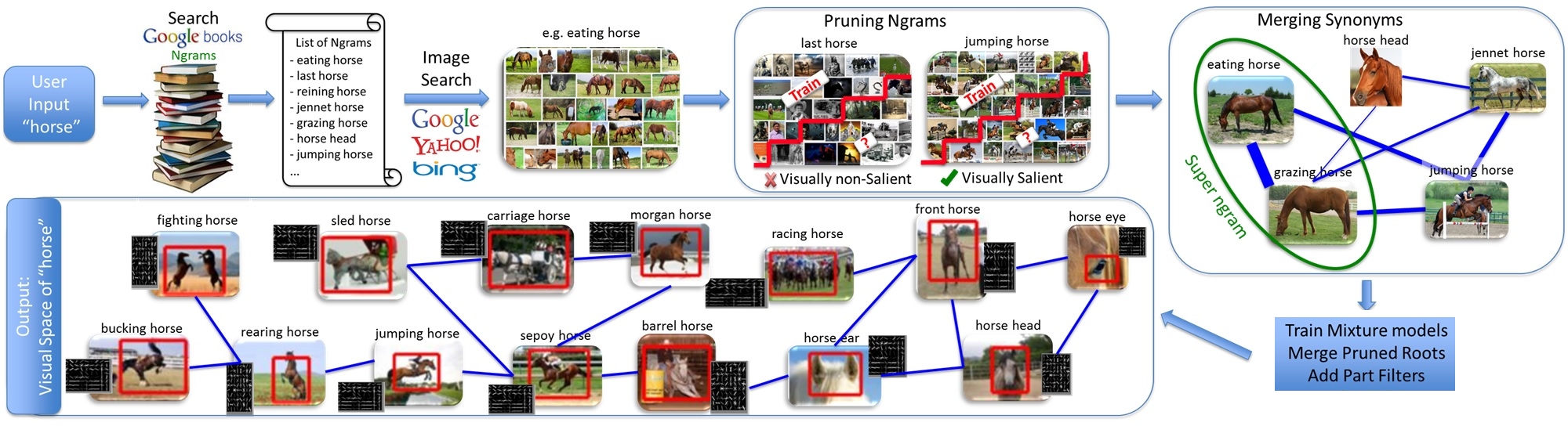}
 \end{center}
    \caption{Overview of LEVAN \cite{santosh2014web}.}
 \label{fig:levan}
 \end{figure*}
\begin{figure*}[t]
\begin{center}
\includegraphics[width=1\linewidth]{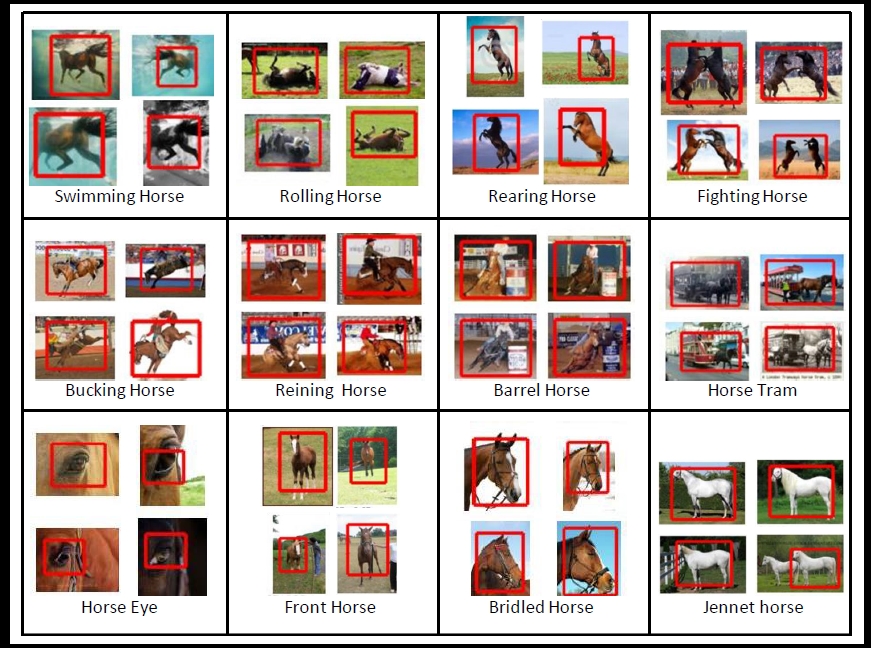} \\
\end{center}
   \caption{Webly-supervised discovered subcategories (for horse category)}
\label{fig:subcategory_LEVAN}
\end{figure*}
\begin{figure}[t]
\begin{center}
\includegraphics[width=9.0cm]{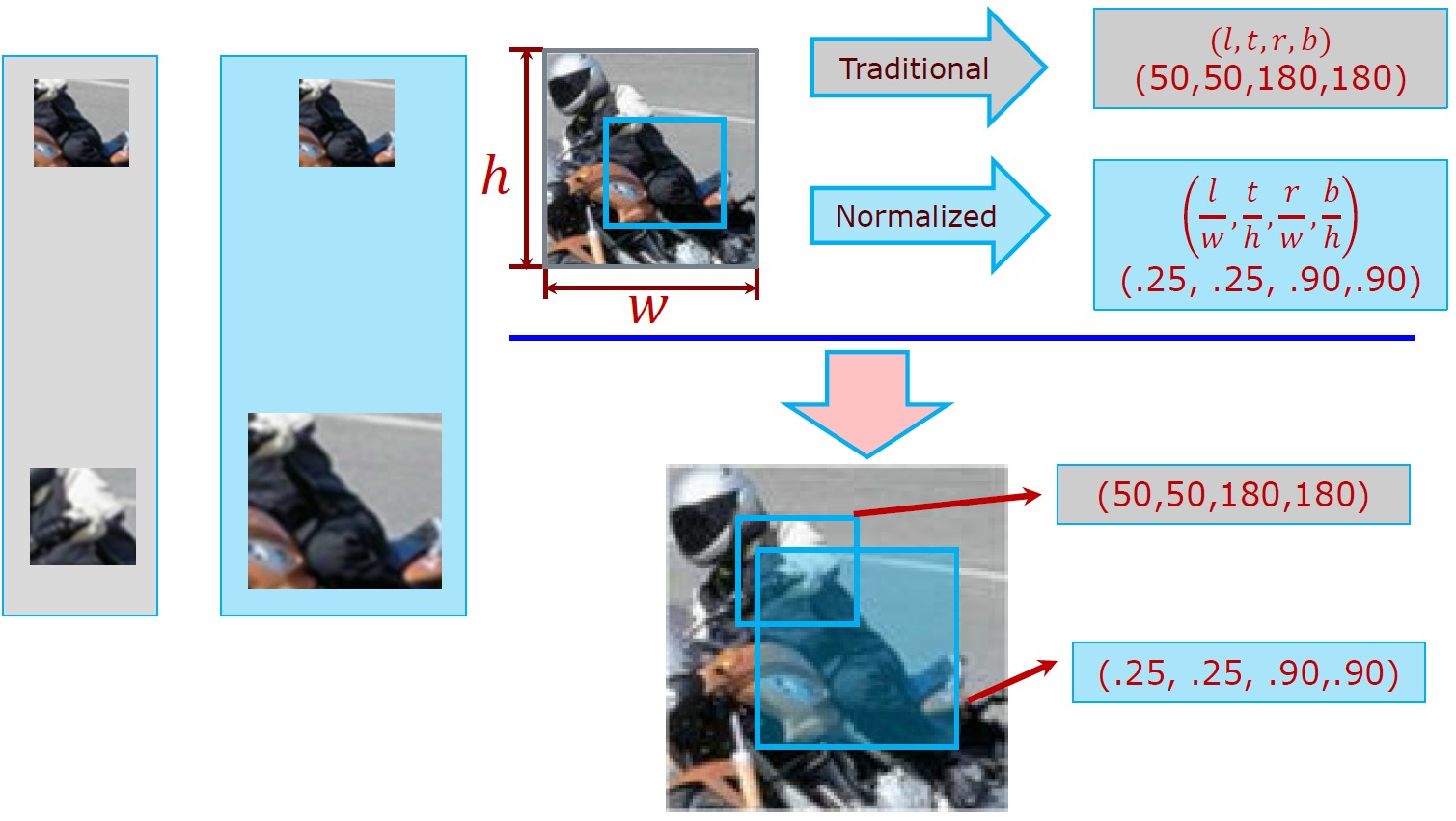} \\
\end{center}
\caption{Regionlet \cite{wang2013regionlets}: normalized feature extraction region.}
\label{fig:regionlet}
\end{figure}
\begin{figure*}[t]
\begin{center}
\includegraphics[width=12cm,height=13cm]{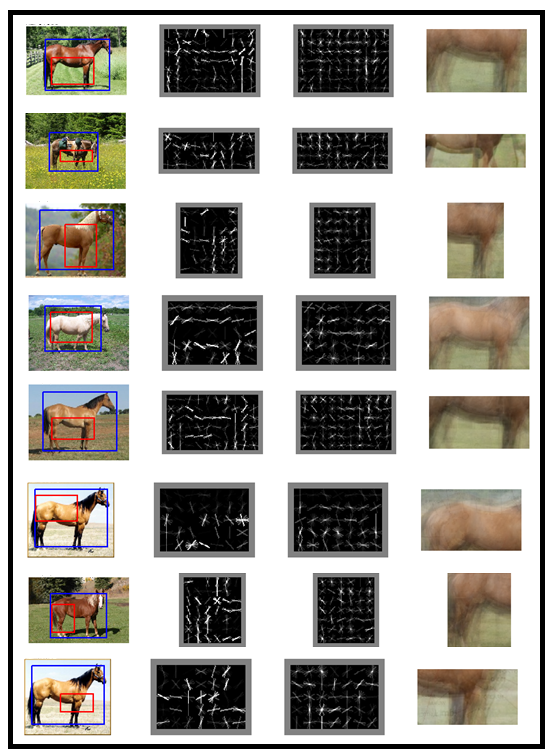}
\end{center}
\hspace{2.2cm} a) \hspace{2.6cm} b) \hspace{2.4cm} c) \hspace{2.4cm} d)
   \caption{Fixed-Position Patch Model: {\bf(a) }relative position, {\bf(b,c) }HOG visualization of model, and {\bf(d) }average image over the detected patches on other images of the subcategory.}
\label{fig:Patch}
\end{figure*}
\begin{figure*} [t]
\begin{center}
\includegraphics[width=1\linewidth]{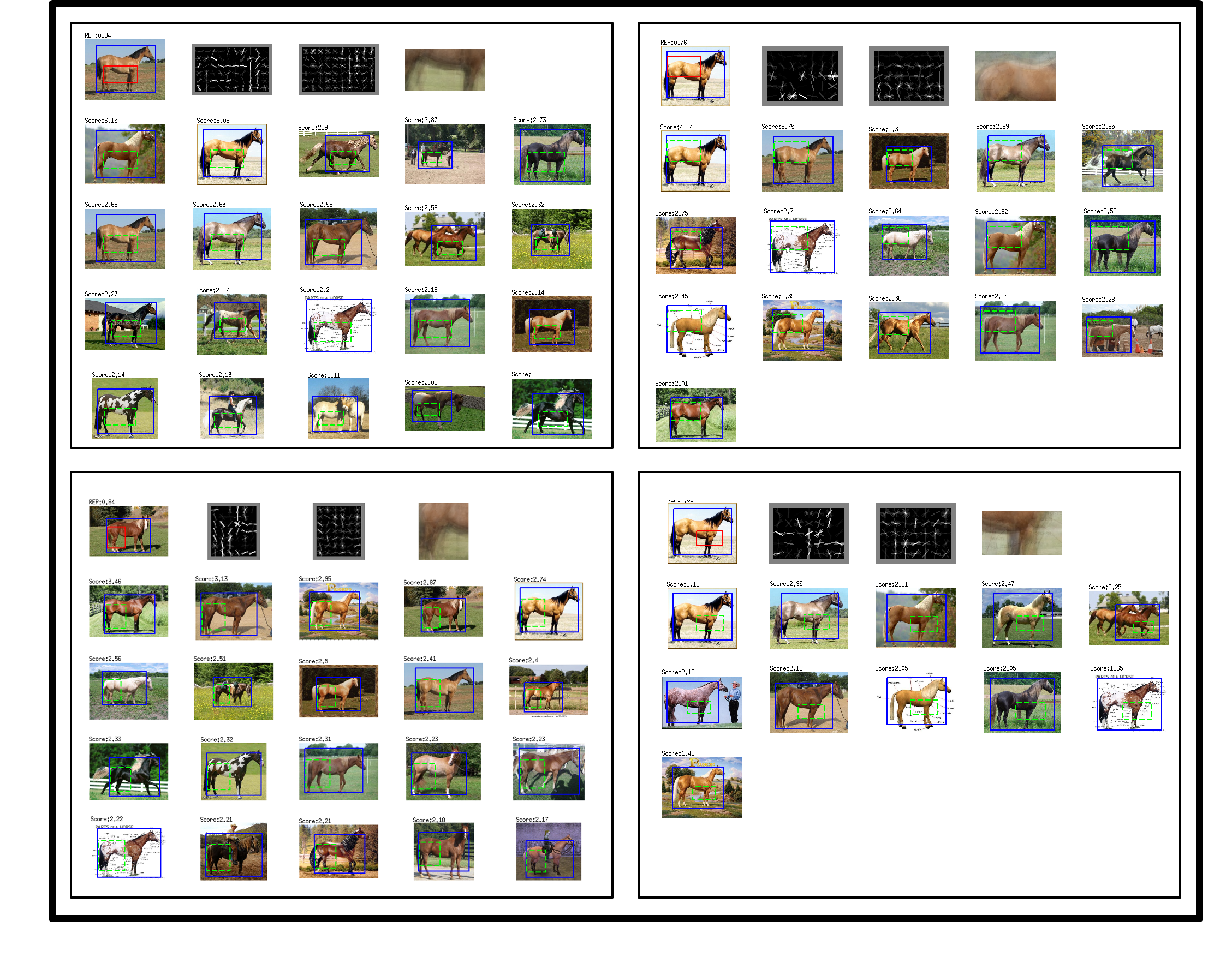}
\end{center}
   \caption{Patch Retrieval}
\label{fig:retrieval}
\end{figure*}

\subsection{Webly-supervised Subcategory Discovery} \label{sec:levan}

We adopt LEVAN for discovering the visual subcategories of an object category as well as training an initial model for each subcategory. \ignore{For each object category, we first determine the visual diversity of that object by discovering the subcategories of that category, so train a model for each subcategory.} The visual diversity of a concept is discovered by looking up over thousands of documents available on-line on the Google books. A sort of query expansion schema has been used for this purpose to find the n-grams linked to that concept. In the case of object category as an input concept, each n-gram corresponds to a visual subcategory of that object class of interest, which reflects a particular semantic of that object category. But not all of those n-grams are visually informative, so a pruning step employed for this purpose. To come up with semantic double counting visual n-grams, an extra post-processing stage is also utilized, which applied for merging synonym visual n-grams and form super n-grams. (See Figure \ref{fig:levan})

For each visual subcategory corresponding to each visual n-gram, a set of images have been downloaded from Internet to create \emph{PASCAL 9990} and a modified version of {\it unsupervised DPM}\cite{pandey2011scene} has been used for training the initial subcategory model. Since images on each subcategory are so homogeneous, an object subcategory model can be trained even without the bounding box of that image. The final detection will be done by ensembling the available subcategory models for that object category. Figure \ref{fig:subcategory_LEVAN} shows a set of subcategories discovered for the horse category (Top 4 downloaded images for some subcategories and the corresponding visual n-grams at bottom).

\subsection{Subcategory-aware Patch Discovery and Training}  \label{sec:Subcategory-aware}

The key novelty of this work is to discover and train the patches in a subcategory-aware fashion. It helps us not only to form a richer object representation, but also to be performed efficiently. The former is provided as a result of having less visual diversity in subcategory-level, and the latter by computing separately per each subcategory and in parallel.

In this section, we first define in detail what we mean by fixed-position patches, then explain how to train the model and finally the criteria we used to select a subset of them for each subcategory. Please note that when we talk about the images of a visual subcategory, we mean samples of the web-downloaded images collected in PASCAL 9990.

\vspace{.1cm} 
\emph{\textbf{Fixed-position Patch - }}We define a fixed-position patch as spatially consistent image parts inside the bounding box of the object. More precisely, a fixed-position patch is a patch which has a fixed visual appearance in the same relative position of the object bounding box. To define the spatial relativity, we follow the idea of Regionlet \cite{wang2013regionlets}, which is using normalized relative position between bounding box of the object and bounding box of the patch. Figure \ref{fig:regionlet} shows the difference of a normalized region and a traditional region in two examples (the original window and a transformed one with double resolution)

\qsection{Initial Patch Model} \label{sec:Initial_Patch}
For each subcategory, we aim at defining a set of patches which can be used later for representing that subcategory. We start by randomly selecting a large set of fixed-position image patches inside the hypothesis bounding box generated by the initial subcategory model. Next, we train a peculiar appearance model for each patch by taking advantage of a tremendously huge set of negative samples. We follow the trend of exemplar-based models and more specifically we use the LDA accelerated version of Exemplar SVM introduced in \cite{hariharan2012discriminative} (see Sec. \ref{sec:Exemplar-LDA}). So, for each patch we do not only have a relative position parameter reflecting the position of the patch to the bounding box, but also an appearance model trained for that patch in a discriminative way. We call the former patch's relative position and the latter patch appearance model. In Figure \ref{fig:Patch}, a set of fixed position patches is shown (Patch in red, object bounding box in blue).

In our experiments, we observed that the defined fixed-position patches might still have a very slight deformation and this deformability is different from patch to patch. We take into account this fact and add patch deformability as a parameter to our patch model. To this purpose we first run each patch's appearance model inside the bounding box of the object on the same image, then capture the patch jitter by discarding the Non-Maximum Suppression (NMS) post-process on the detection responses. The bounding boxes generated as a result of multiple detection in this process is used then to compute the deformability parameter for each patch.

\label{sec:Patch_selection} \qsection{Patch Selection}
We start with a large number of random patch models which trained using a single patch. But, not all of these patches are useful to be selected in final subcategory model: some are not visually or spatially consistent for all the images belong to that subcategory, and some other not discriminative enough for the subcategory. We refer the former property as the {\it representation measure} and the latter as the {\it discrimination measure}. To select a more representative and discriminative subset of patches, we define a scoring function based on the stability of each patch in terms of spatial and visual consistency.

In other words, we aim at select a set of patches which are not only repetitive pattern in all images of the subcategory, but also fixed-position relative to the object bounding box. For this purpose, we define two types of scores for each patch on the image: {\it appearance consistency} and {\it spatial consistency} scores. (shown as two terms $\phi$ and $\psi$ in following equation \ref{eq:patch_score_1})

\begin{align}
score(x,p) = \omega\cdot\phi(i,z) + \psi(z,l), \label{eq:patch_score_1} ~~~~~~~~ \\
z = \argmax_k~(\omega\cdot\phi(i,k)),\label{eq:patch_score_2}~~~~~~~~~~~~~\\
\psi(z,l) = \frac{(z \cap l)}{(z \cup l)}, \label{eq:patch_score_3}~~~~~~~~~~~~~~~~~~\\
l = \varphi(b,r),\label{eq:patch_score_4}~~~~~~~~~~~~~~~~~~~~~~\\
x = (i,b),~~~~~~~~~~~~~~~~~~~~~~~\\
p = (\omega,r).~~~~~~~~~~~~~~~~~~~~~~~
\end{align}

The {\it appearance consistency score} $\phi$ reflects the confidence of a patch model $p$ to be detected in a given bounding box $b$ inside image $i$. We simply use the confidence score of the LDA model $\omega$ for this purpose and pick the best detection $z$ in that example $x$. We note that, $\phi(.)$ denotes the feature representation for a bounding box inside an image (Histogram of Oriented Gradient \cite{dalal2005histograms}).

The {\it spatial consistency score} $\psi$ shows how much that patch happens in the fixed position relative to the bounding box of the object in other images of that subcategory. For each patch on each image we compute the score of spatial consistency $\psi$ as the intersection over union between the detected bounding box and the potential bounding box of the patch. The detected bounding box $z$ is the region on the image in which our patch model get the maximum detection score, and the potential bounding box $l$ is where the relative position parameter $r$ in our patch model select as the best location of that patch inside the object bounding box (shown as $\varphi$ in equation \ref{eq:patch_score_4}).\\ 
\\Based of these two scores, we define two measures:\\
\emph{\textbf{Representation Measure:}} This measure shows how representative a patch is for the samples belongs to that subcategory. If this measure is high for a particular patch it means that patch is a repetitive pattern in the subcategory, so it probably worths to be involved in the final model.

For each patch, we run that patch appearance model on all images $I_s$ belong to the subcategory $s$, and collect appearance consistency as well as spatial consistency score for each image. Next, we compute the representation measure on each image of the subcategory by summing over appearance score and spatial score of it (similar to equation \ref{eq:patch_score_1}). We do this summation after normalization per/across patches. Finally, to find an overall representative measure for each patch we simply computing average over all images belong to that subcategory.
\begin{align}
rep (p,I_s) = \frac{1}{I_s} \sum_{i=1}^{\rvert I_s \rvert} score(x_i,p)
\end{align}
\emph{\textbf{Discrimination Measure:}} A good patch should be not only representative for a subcategory but also discriminative for that object category. 
For this purpose we built a mixed set of both images of the subcategory $I_s$ and a huge negative set of PASCAL training images $\bar{I}$. To find hard negatives, we run the patch detectors inside the hypothesis bounding boxes detected by initial subcategory model. We measure the discriminativity of a patch as median rank of retrieved images from the mixed set. The similarity measure defined for the retrieval on the mixed set was the same as the one has been used already for the inter-subcategory retrieval purpose: \emph{summation over spatial and appearance consistency.} In another word, a patch is discriminative if the top retrieved images in the mixed set belong to the subcategory in PASCAL 9990 and not to the PASCAL 2007 negatives.
\begin{align}
disc (p,I_s,\bar{I}) = \frac{median(rank(p,I_s,\bar{I}))}{I_s},\\
rank(p,I_s,\bar{I})~:~\mathbb{R}^{\rvert I_s \cup \bar{I} \rvert} \mapsto \mathbb{N}^{\rvert I_s \rvert}~~~~~
\end{align}
\\The final measure for selecting a good subset of patches takes into account both the representation and discrimination measures and is simply defined as the weighted sum of them (weights are equal in current implementation).
In Figure \ref{fig:retrieval} we selected four patch models of a subcategory, and showed the top retrieved images for each of the patches. The patches are so consistent both visually and spatially across samples.
 
\qsection{Patch Re-training} \label{sec:Patch_Re-training}
Our initial patch model is trained only on a single positive sample and a fixed set of negative data. So, the LDA model by itself can not perform very well, but as \cite{ahmed2014knowing} observed in the case of Poselet~\cite{bourdev2009poselets}, the performance of the LDA models and that of the SVM models are highly correlated. So, we follow the same regime, doing patch selection based on efficient (yet poor) LDA models, then do re-training an expensive (yet accurate) models only for the selected Patches, providing a further acceleration in training time. So, after selecting the best subset of initial patches, we retrain patch models by employing all retrieved samples for that patch using Latent-LDA\cite{girshick2013training} which is shown to perform efficiently as well as accurately.

\subsection{Patch-based Subcategory Model and Inference} \label{sec:merge}
Given an initial subcategory model and a set of patch models, we should build a final subcategory model. For this purpose, we first stack up the selected patch models and calibrate them all. We introduce a context-rescoring scheme for patch calibration. In inference time, we take advantage of patch models in a voting-based fashion \cite{farhadi2010attribute}. In the following, each of these steps will be explained in detail.

\qsection{Patch Calibration}  \label{sec:Patch_Calibration}
The patches trained for each subcategory have different sizes, so the detection scores of different patch models are not directly comparable. To tackle this problem, we did a patch calibration post-processing. We used a context-rescoring like scheme applied to a held-out set of images. This set consists of some images of that subcategory as well as some negatives samples (which were not used in the patch-based training step). This held-out set is provided in PASCAL 9990 for validation purposes and we used it for calibration in our work. In more detail, we first run all patch models on the samples of the validation set, then train an SVM on the detection score of them. It provides us with a weight vector reflects the contribution of each Patch model in the final detection stage. It is worth pointing out that we use the monolithic subcategory model along with all patch models in the calibration step.

\begin{figure*}[t]
\begin{center}
\includegraphics[width=13cm,height=14cm]{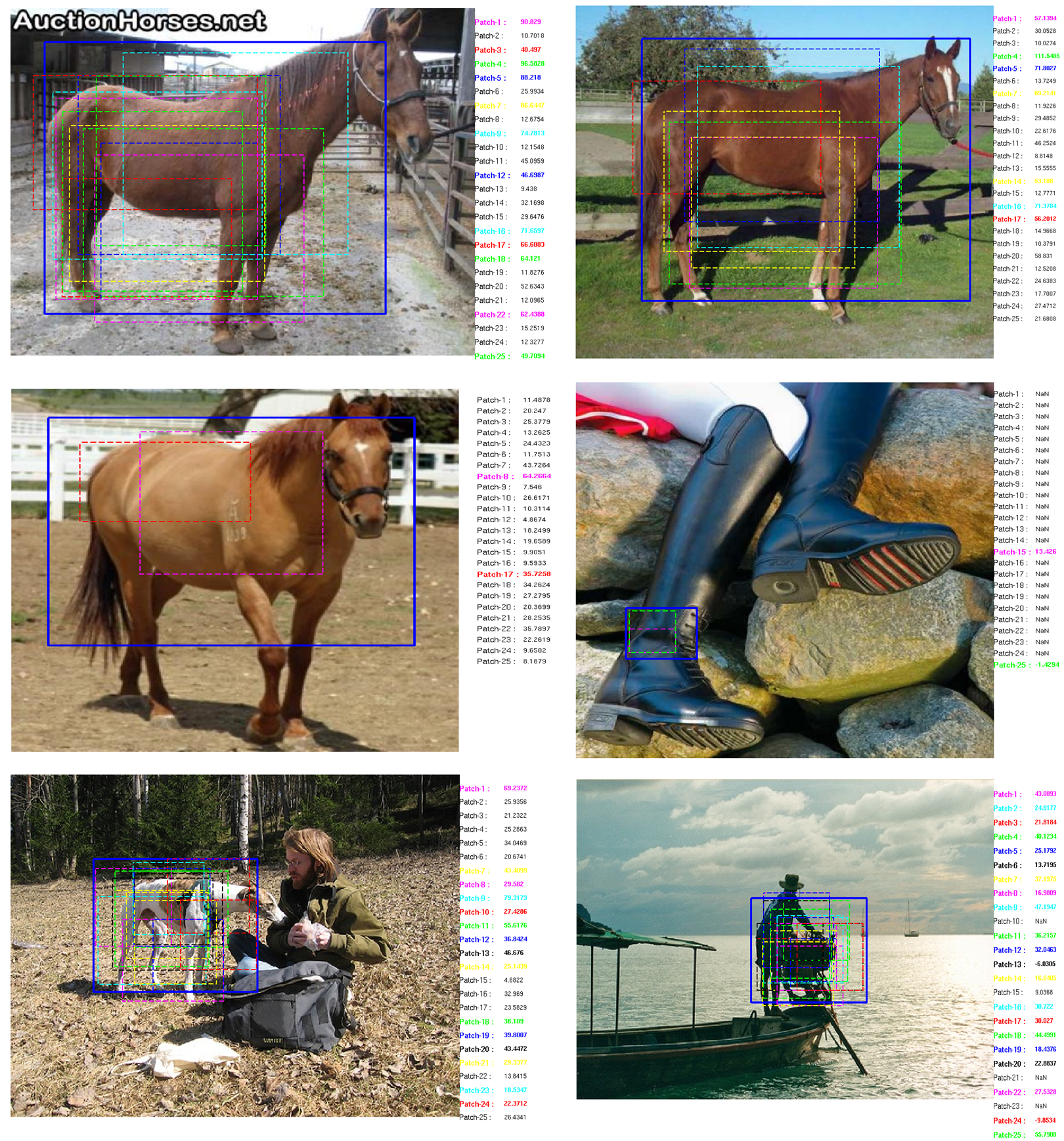} \\
\end{center}
   \caption{Patch calibration: {\bf(first row) }corrent subcategory images, {\bf(middle row) }wrong bounding boxes, and {\bf(last row) }false patch activation.}
\label{fig:patch_rescoring}
\end{figure*}

In our experiment, we observed that there are samples which the monolithic subcategory model have high score, but the patches are not activated. 
This is mainly due to the fact that the initial subcategory model trained using noisy web images employing LEVAN. We experimentally found that the patch calibration is so beneficial even for sample selection. We specifically use the samples which not only the monolithic subcategory model has high score, but also the patch models. This criteria specifically undo the effect of semantic drifting as for web-supervision. 
As shown in Figure \ref{fig:patch_rescoring}, in first row (correct samples) not only the initial model highly-detected in images (in blue), but also patches models have high scores. In the middle row (false samples), although initial model is detected by mistake, but our patches models are not highly-detected. The left sample shows an example from similar but different subcategory and the right sample illustrates the semantic drifting due to presence of the web-supervision bias. The last row in the figure belongs to the false alarm of this strategy. We note that even for this case both shown samples are still visually similar to ``horse''.


\qsection{Subcategory Inference}  \label{sec:unsup_obj}
Since the relative position of each patch to the bounding box of the object is implicitly embedded in the patch model, we can simply use a voting-based strategy at inference time. For each test image, we run the monolithic subcategory model as well as all patch models. Given the patch detections, we predict a set of hypothesis bounding box for the object. The final bounding box of the object will eventually be produced from these predicted bounding boxes. Specifically, we probe three different methods for this purpose: \emph{Non-Maximum Suppression (NMS), median bounding box, bounding box clustering.}

\begin{figure*}
\begin{center}
\fbox{\includegraphics[width=1\linewidth]{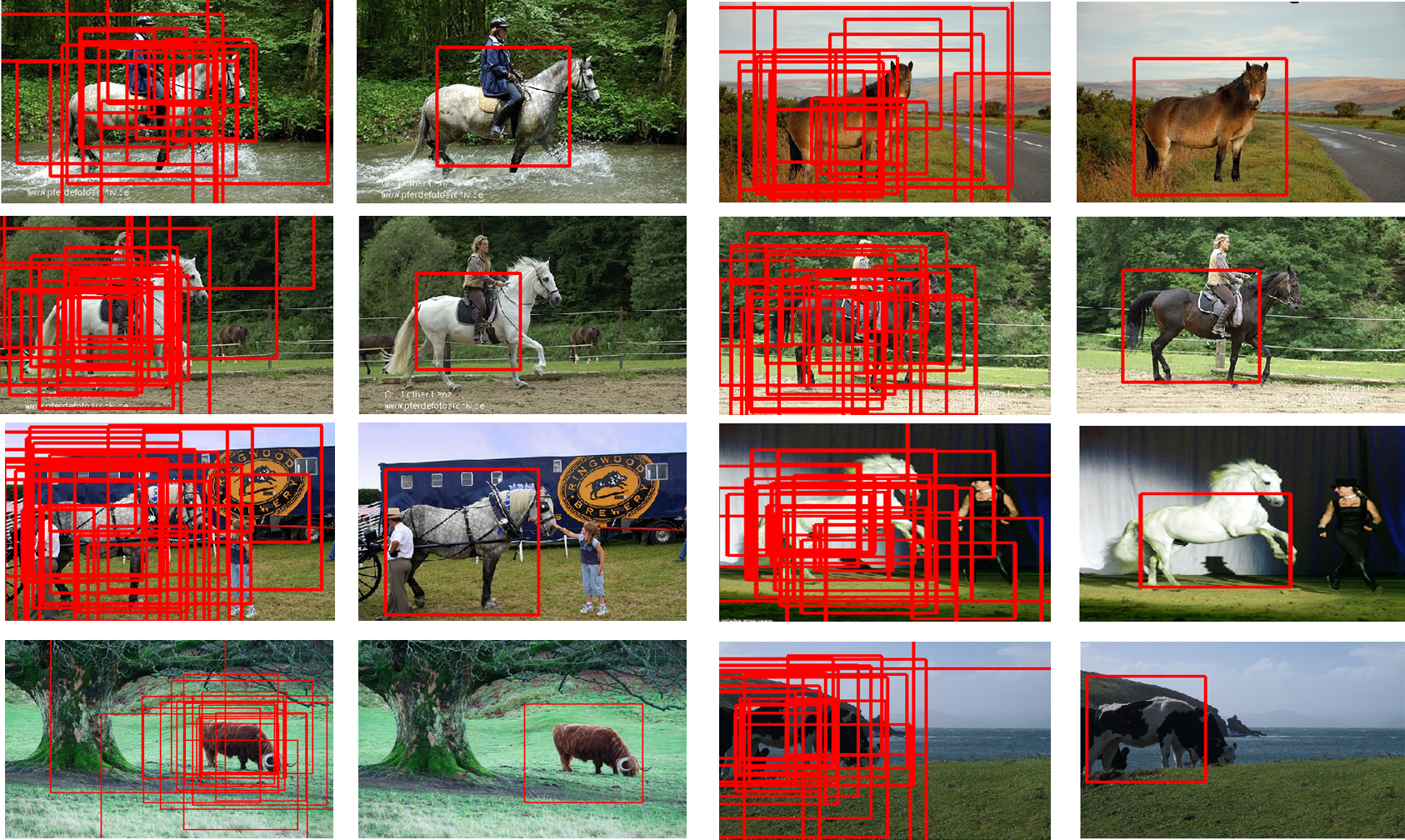}}
\end{center}
   \caption{Final bounding box from predicted bounding boxes.}
\label{fig:detectedBB}
\end{figure*}

The first two methods are done similarly to other well-established object detection methods \cite{felzenszwalb2010object}. In the case of \emph{bounding box clustering}, we employed k-means on the x-y coordinations of predicted bounding boxes to find the final hypothesis bounding boxes. In our experiments, we empirically observed that the clustering methods is outperform two other methods when the number of objects in the image (k in k-means) is known. But since this information is not given at inference time, so we used the \emph{median bounding box} method in our current implementation, which shown to be efficient and effective. (See Figure \ref{fig:detectedBB})

\ignore{To form a final category-level model from a set of subcategory models, we will use the same rank based measure introduced in \cite{santosh2014web} or \cite{ahmed2014knowing}. Our current implementation is evaluated only on individual subcategories, so we put this step as a future work of this research.}

\vspace{-.3cm}

\subsection{Experiments} \label{sec:experiments_patch}

Although we focus on the subcategory level object recognition, the framework we present in this work extends in a natural manner to training detection models for fine-grained visual recognition. To evaluate the performance of our approach, we present results for unsupervised object detection in subcategory level.\ignore{ and abnormality detection in still images.}

Our model trained using the web images downloaded from Internet (PASCAL 9990), and tested on the PASCAL 2007 in unsupervised object detection fashion. To have an accurate analysis, we confine to evaluate our method in detail on the subcategory ``Mountain Horse'' as one of the most iconic subcategories of the ``horse'' category. The other experiments on other subcategories are also confirm empirically the results of the method on this subcategory. 

\begin{figure*}
\begin{center}
\includegraphics[width=0.7\linewidth]{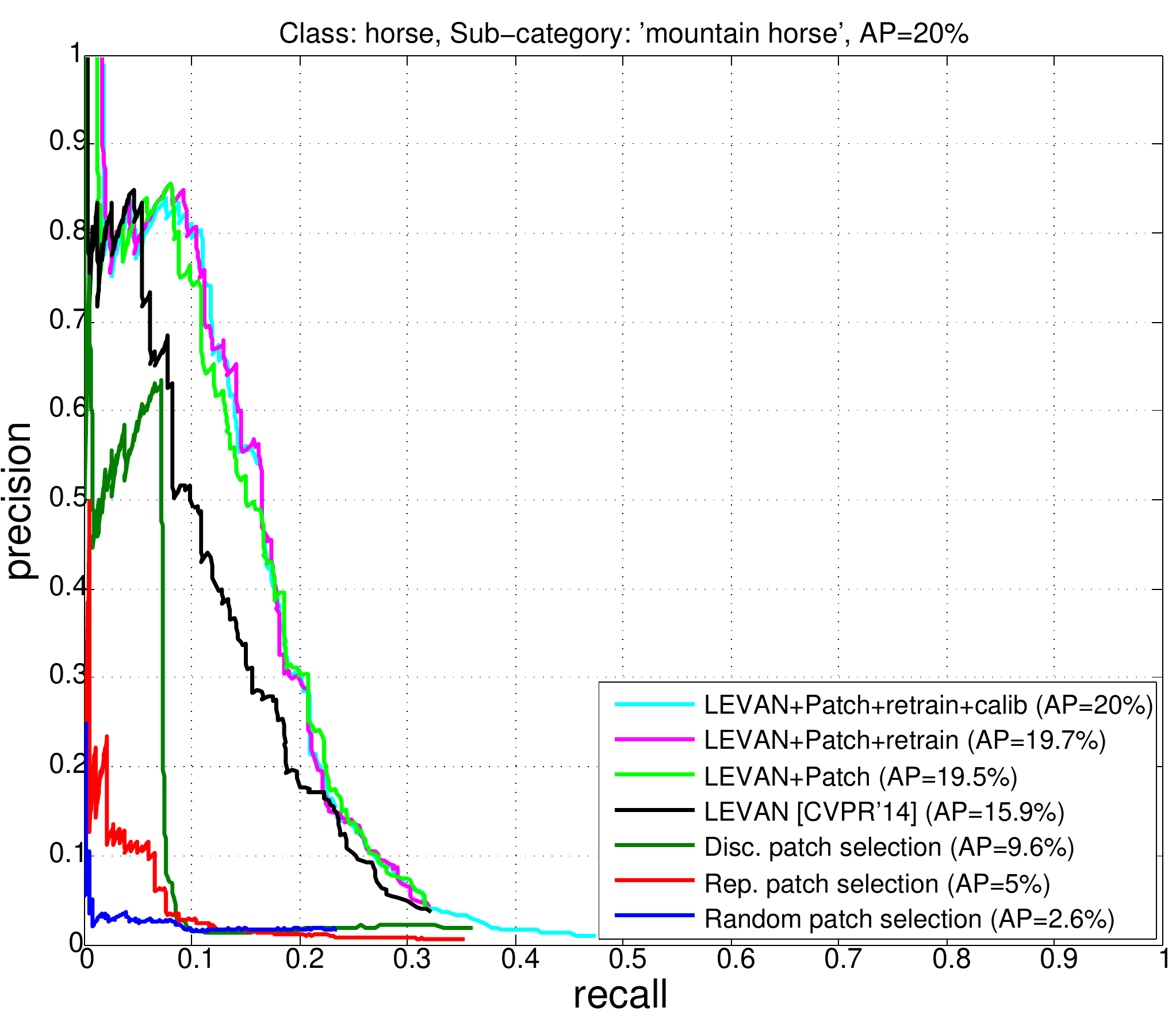}
\end{center}
   \caption{Average Precision} 
\label{fig:PR_curve}
\end{figure*}

We adopt the data, experimental setup and the initial subcategory model from LEVAN and built our model on top of it. In training step, we train our patch model inside the bounding box generated by LEVAN and for each subcategory we produce a set of patch models. For evaluation, we study the effect of each step of the method on building the final model by analyzing the performance of it on the PASCAL 2007 testset. We picked this dataset as recent state-of-the-art weakly supervised methods have been evaluated on it. In our evaluation, we ensured that none of the test images of the PASCAL 2007 testset existed in our trainset (PASCAL 9990). 

We illustrate the evolution of the experiments and the results as it showed in Figure \ref{fig:PR_curve}. Different colors show to the results obtained by using different versions of the proposed algorithm.

We first analyzed the effect of employing different patch selection strategies (as described in section \ref{sec:Patch_selection}). While in \textbf{(blue)} the \emph{random} patch selection used as the selection strategy, in \textbf{(red)} we employed \emph{representation measure} for this purpose and in \textbf{(dark green)} we added \emph{discrimination measure} as the supplementary criteria for patch selection. As shown in the figure, the former boosted the \textbf{Average Precision} (AP) for ~5\% and the latter ~4.6\%. These gains achieved respectively by leveraging web-supervision provide in PASCAL 9990 and the discriminative power captured utilizing PASCAL negatives.

\begin{figure*} 
\begin{center}
\includegraphics[width=1\linewidth]{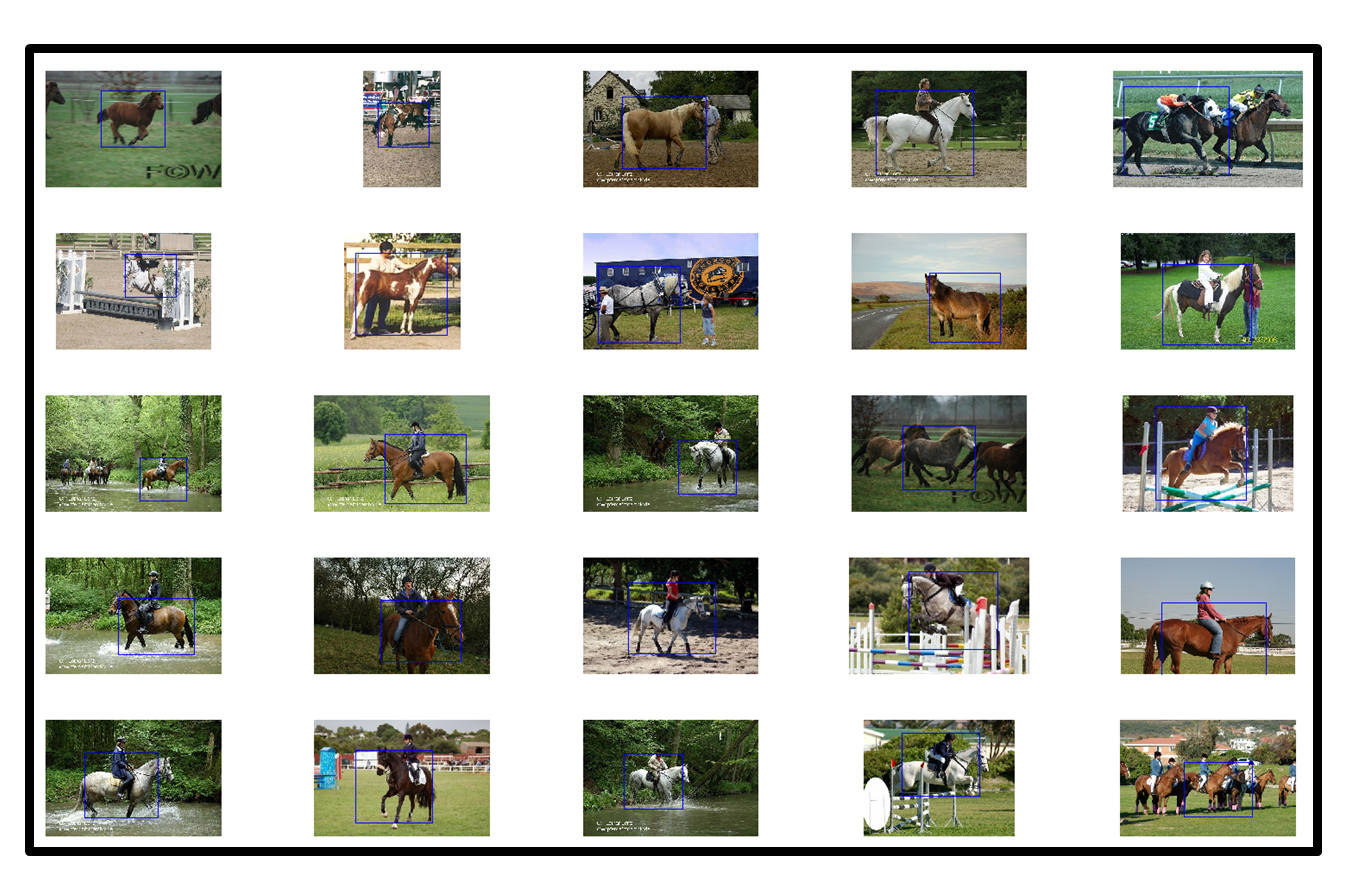}
\end{center}
   \caption{Detected object from PASCAL 2007 ('horse' category)} %
\label{fig:qulitative}
\end{figure*}

Next, we analyzed the combination of our proposed method with the state-of-the-art baseline LEVAN (shown in \textbf{black}). In \textbf{light green} we simply combine the trained patch model with the monolithic model of LEVAN. While the state-of-art \cite{santosh2014web} is 15.9\%, we could achieve to a higher performance (\textbf{19.5\%}) by adding more flexibilities to the model in terms of handling occlusions and employing object parts (see qualitative results \ref{fig:qulitative}). We also obtained the 0.2\% additional improvement by performing \emph{retraining of the patch models} using Latent-LDA which is shown in color \textbf{magenta} in the figure (section \ref{sec:Patch_Re-training}). Finally in \textbf{cyan} we showed the effect of \emph{patch calibration}. It provided us with 0.3\% extra improvement in terms of AP. We justify this as the effect of removing false web detected images and wrong subcategory assignments shown qualitatively in Figure \ref{fig:patch_rescoring} (section \ref{sec:Patch_Calibration}).

Our method uses web supervision as not even the images are explicitly supplied for training. Nonetheless, our result substantially surpasses the best performing webly-supervised method (LEVAN) with 4.1\% improvement. More precisely, with the same detection threshold with LEVAN, we got the significantly better recall showing the boost provided by our method. In addition, we reduced the detection threshold on LEVAN to generate a larger set of hypothesis initial bounding box in test time to be penalized with patch models. Figure \ref{fig:qulitative} shows some of the qualitative results for ‘horse’ category on PASCAL 2007.

\ignore{

\begin{figure*}
\begin{center}
\includegraphics[width=1\linewidth]{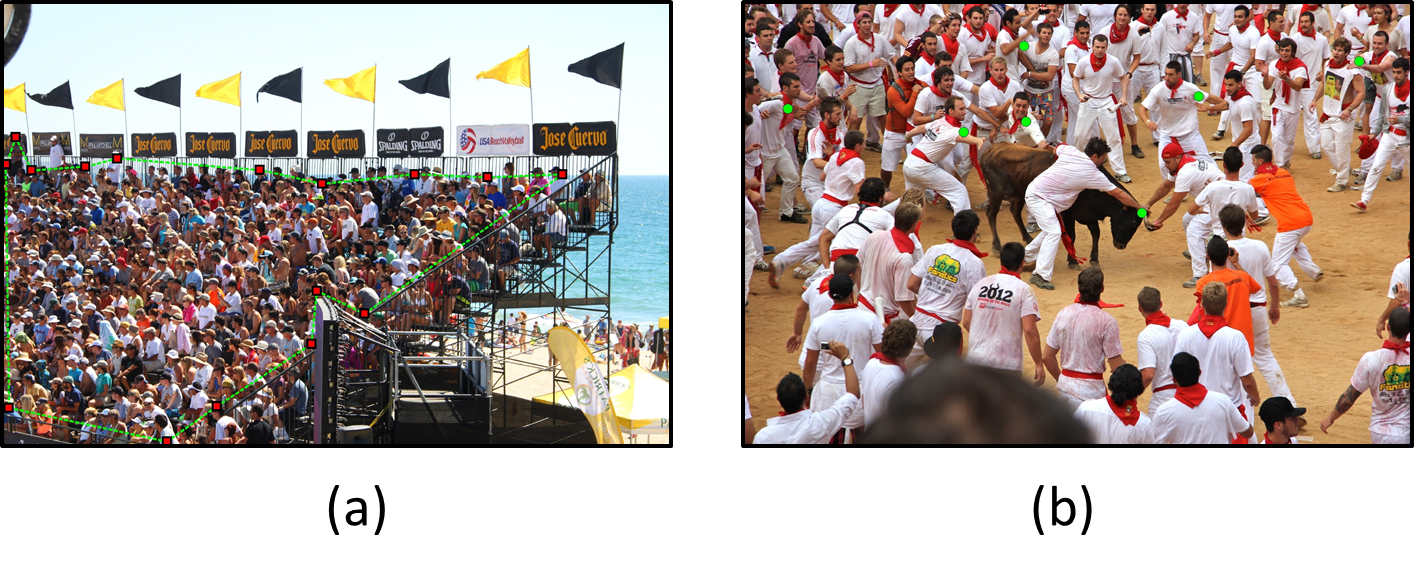}
\end{center}
   \caption{Abnormality Image Dataset  {\bf(a) }normal, {\bf(b) }abnormal.}
\label{fig:abnormality_dataset}
\end{figure*}

\qsection{Abnormality Detection in Still Images}
We used our introduced patch model for the task of abnormality detection in still images. Abnormality detection in videos has been studied extensively in computer vision community [citation]. Most of proposed methods used motion information to extract the pattern of normal/abnormal behavior in the crowd environments. We consider the more challenging task of \emph{image-based} abnormality detection, where not only the is no motion information is provided for still images (unlike videos), but also we aim at localized abnormality inside an still images with a bounding box. Further, we attempt to create a heat-map for the images showing where in the image is more probable to see an abnormal behavior. To the best of our knowledge, this is the first attempt on image-based abnormality detection in the crowd environment. For this purpose we train our patch model in a strongly-supervised setting, in which the patches capture the abnormal appearances in an crowd scenarios. The provided supervision for the method is the normal crowd region examples as well as some exemplar abnormal patches.

\noindent \textbf{Datasets- }There is no dataset available for the task of "abnormality detection in still images" in the litreture, so we collect a set of normal/abnormal images by downloading Google images using query expansion strategy and created a dataset. For query expansion we used a similar strategy like PASCAL 9990, we started by a predefined set of crowd related key-word (e.g. crowd, festival, gathering, social event ...) and download a set of images. Our dataset includes an order of thousands normal and hundreds abnormal images. Since our method is strongly supervised, we employed \emph{Amazon Mechanical Turk} users (we make use of \emph{LabelMe} instead in current implementation) to annotate the images. We asked the annotators, to draw a polygon around the region of crowd in the normal images. For the case of abnormal images, we asked the annotators to select some salient points on the images where abnormality is bounded up with. (See Figure \ref{fig:abnormality_dataset} for sample normal/abnormal images as well as provided supervision captured from annotation tool)

\noindent \textbf{Implementation details- }We simply start by extracting random patches around salient points in the abnormal images in training set. For each patch we train our previously explained patch model using Exemplar-LDA (see section \ref{sec:Initial_Patch}). For patch selection we employed the same criteria we used for the case of object detection, where \emph{representation measure} and \emph{discriminative measure} adopted for abnormal and normal patches. Differently from the object detection case, here we just used the \emph{appearance consistency} and NOT \emph{spatial consistency} since there is no regularity about the position of abnormality in the images. In other word, an abnormal patch is a suitable patch if it repeated on the salient abnormal points on abnormal images and not detected in the normal regions in the normal images. The final abnormality localization in inference time is just running abnormal patch models on the test images and generate heat map of abnormality on it. There is no quantitative evaluation performed yet for this experiment, but our qualitative results confirm the idea. Quantitative evaluation will be added soon. (see Figure \ref{fig:abnormality_qualitative})
\begin{figure*}
\begin{center}
\includegraphics[width=1\linewidth]{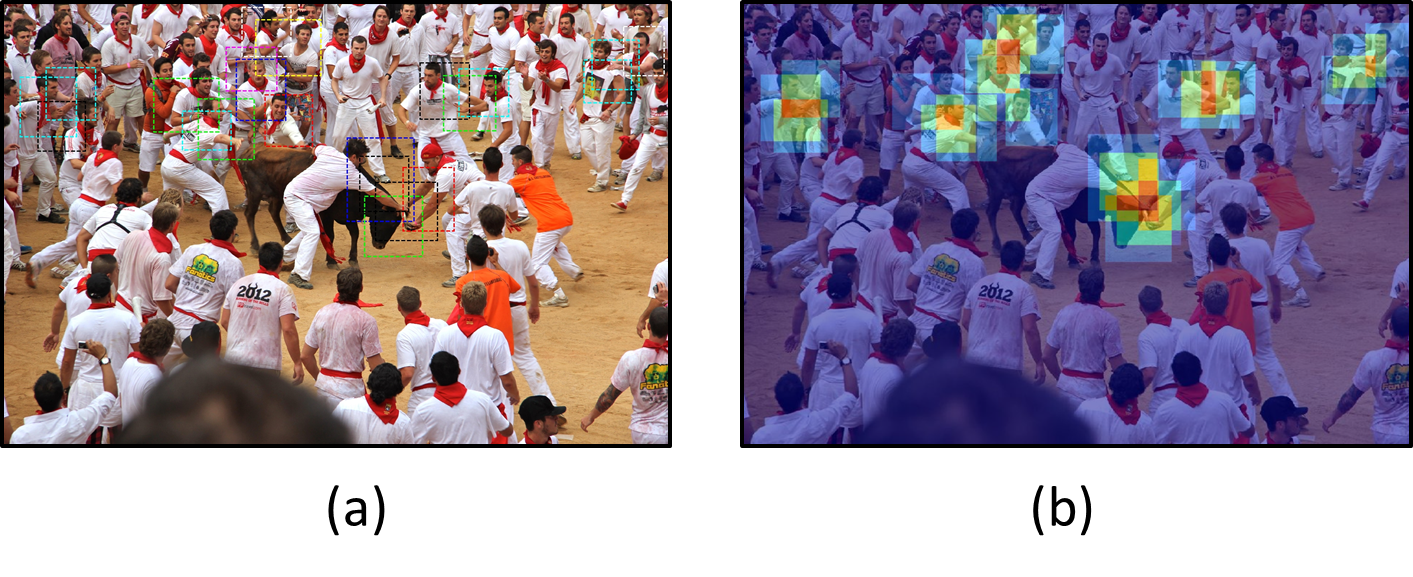}
\end{center}
   \caption{Abnormality Patch Model {\bf(a) }random patch, {\bf(b) }heat map.}
\label{fig:abnormality_qualitative}
\end{figure*}

}

\subsection{Conclusions} \label{sec:conclusions_patch}
There have been some recent efforts to build the visual recognition models by leveraging the gigantic collection of unlabeled images available on the web \cite{santosh2014web,chen2013neil,chen2014enriching}. In this section, we followed the same regime and introduced the first webly-supervised discriminative patch as an effective patch-based model for mid-level representation. An important point worth to be emphasized is that we discovered the discriminative patches in a subcategory-aware fashion due to the fact that we observed that the dataset bias (e.g. Internet bias) is less in subcategory level. Our experimental evaluation shows our subcategory-based discriminative patch can effectively be employed as mid-level image representation, so is useful for the image understanding tasks such as unsupervised object detection.\ignore{ and novel task of image-based abnormality detection in crowd.}

\newpage
\thispagestyle{empty}
\mbox{}

\part*{Part II: Mid-level Representation for Video Understanding}
\chapter{Mid-level Video Representation} \label{chap:video}
{\LARGE \qsection{Application}\emph{ Human Behavior Understanding}}\\[2cm]

In this chapter, we explore the significance of utilizing mid-level techniques for video representation. We first start by employing pure motion statistics for this purpose, then complement it by acquiring appearance information to build a richer representation. We specifically introduce a tracklet-based commotion measure and a detector-based motion representation for human behavior understanding in crowd.

\section{Introduction}
The study of human behavior has become an active research topic in the areas of intelligent surveillance, human-computer interaction, robot learning, user interface design and crowd analysis. Among these fields, group and crowd analysis have recently attracted increasing attention in computer vision community for different problems such as group/collective activity recognition \cite{choi2012unified,lan2012discriminative}, event detection \cite{tang2012learning,kwon2012unified}, and crowd behavior analysis~\cite{mehran2009abnormal,wu2014bayesian,wang2012abnormal}. 

Current state-of-the-art computer vision methods work well for problems within limited domains and for specific tasks, and human behavior understanding is not an exception. However, when such methods are applied in more general settings, they become brittle, much less effective, or even computationally intractable. In a nutshell, existing computer vision approaches are not general, and they do not scale well. Specifically, the techniques designed for the detection of individual behavioral patterns are not suitable for modeling and detecting events in (highly) populated (i.e., crowded) scenes. On the other hand, a video captured from a realistic scene of human activity involve multiple, inter-related actions, human-object and human-human interactions at the same time. Moreover, it is well noted in the sociological literature that a crowd goes beyond a set of individuals that independently display their personal behavioral patterns~\cite{wijermans2007modelling, benesch1995atlas}. In other words, the behavior of each individual in a crowd can be influenced by ``crowd factors'' (e.g. dynamics, goal, environment, event, etc.), and the individuals behave in a different way than if they were alone. This, indeed, implies that crowds reflect high-level semantic behaviors which can be exploited to model crowd behaviors~\cite{junior2010crowd}. This has encouraged the vision community to design tailored techniques for modeling and understanding behavioral patterns in crowded scenarios. A large portion of recent works is dedicated to recognize the collective activity of people in groups and model and detect abnormal behaviors in video data.

The \emph{motion} characteristics of the crowd scenes have been studied extensively for human behavior understanding. Specifically, both low-level features (e.g., optical flow~\cite{mehran2009abnormal}) and high-level features (e.g., long trajectories~\cite{hu2006system,delaitre2012scene}) have been employed for this purpose. While low-level features can effectively capture the local motion statistics, the high-level features may catch the global crowd structure. Employing an intermediate-level representation (e.g., tracklets \cite{tracklets}) can fill the gap between these two extremes, providing the capability of taking advantage of both local and global information (see more discussions in Section \ref{sec:tracklet_mid}).

Using motion-based methods for crowd representation has shown to be noticeably effective crowd behavior understanding. However, there has been recently a growing number of approaches suggesting that using only low-level motion features may be insufficient, and that improvement can be achieved through the use of \emph{appearance} information. The main reason for using the motion information in crowd analysis is the low-quality of videos captured by surveillance cameras and the real-time nature of this task. However, the motion-based representations suffer from the lack of an actual semantic interpretation since the identity and the appearance information of human parts peculiar of the interaction are typically disregarded. Employing appearance information, on the other hand, can narrow down the gap by injecting semantic values to spatio-temporal features, so leading to more powerful representations of human behavior.

Following in this chapter, we study the utilization both motion-based and appearance-based approaches for mid-level video representation. To this purpose, we first introduce a motion-based video representation, employing tracklets as intermediate-level representation\cite{mousavi2015crowd}. We specifically propose a novel unsupervised measure to capture the commotion of a crowd motion for the task of abnormality detection in crowd scene in pixel, frame and video levels (Section \ref{sec:motion}). Next, we focus on the problem of detecting/recognizing the activity of a group of people in crowded environments. For this purpose, we introduce an appearance-based video representation (i.e., temporal Poselets \cite{nabi2013temporal}), by exploiting the dynamics of a given dictionary of patch-based models (e.g., Poselets \cite{bourdev2009poselets}) in time (Section \ref{sec:app}).

\section{Motion-based Video Representation} \label{sec:motion}

In this section, we study the utilization motion statistics in crowd scenes for video representation. To this purpose, we employ tracklet \cite{tracklets} as mid-level representation for crowd motion. We specifically introduce a novel tracklet-based measure to capture the commotion of a crowd motion for the task of \emph{abnormality detection in crowd}. The unsupervised nature of the proposed measure allows to detect abnormality adaptively (i.e., context dependent) with no training cost. We validate the proposed method extensively in three different levels (e.g., pixel, frame and video) on UCSD~\cite{Authors14}, SocialForce\cite{sfm}, York~\cite{zaharescu2010anomalous}, UMN and Violence-in-Crowd ~\cite{conf/cvpr/HassnerIK12} datasets.
The results show the superiority of the proposed approach compared to the state-of-the-arts.

\subsection{Introduction}
\label{sec:intro}

Abnormal behavior detection in highly-crowded environments plays an important role in public surveillance systems, as a result of worldwide urbanization and population growth. Abnormality detection in crowd is challenging due to the fact that the movements of individuals are usually random and unpredictable, and occlusions caused by overcrowding make the task even more difficult.

Abnormal events are often defined as irregular events deviating from normal ones and vice-versa. The intrinsic ambiguity in this chicken-and-egg definition leads to convert the abnormality detection to an ill-posed problem. For example, slowly walking in a subway station is a normal behavior, but it appears as totally abnormal in the rush hours at the same place due to creating stationary obstacles.
This observation demands an \emph{in-the-box} viewpoint about the abnormality in which introducing a \emph{context-dependent} irregularity measure seems crucial. For this purpose, the abnormal behaviors in crowded scenes usually appear as crowd \emph{commotion}, so that anomaly detection is in general a problem of detection of crowd commotion \cite{sfm,energy}. This school of thought investigated a wide range of unsupervised criteria for this purpose and introduced different commotion measures to the literature. It has also been shown that the measure-based (\emph{unsupervised}) methods may outperform supervised methods, because of the subjective nature of annotations as well as the small size of training data~\cite{energy,sodemann2012review,xiang2008video}. 

\qsection{Literature review} The existing approaches for detecting abnormal events in crowds can be generally classified into two main categories: \emph{i)} object-based approaches, and \emph{ii)} holistic techniques.

\emph{Object-Based Approaches.} Object-based methods treat a crowd as a set of different objects. The extracted objects are then tracked through the video sequence and the target behavioral patterns are inferred from their motion/interaction models (e.g. based on trajectories). This class of methods relies on the detection and tracking of people and objects. Despite promising improvements to address several crowd problems ~\cite{rittscher05,rabaud06}, they are limited to low density situations and perform well when high quality videos are provided, which is not the case in real-world situations. In other words, they are not capable of handling high density crowd scenarios due to severe occlusion and clutter which make individuals/objects detecting and tracking intractable~\cite{marques2003tracking,piciarelli2008trajectory}. Some works made noticeable efforts to circumvent robustness issues. For instance, Zhao and Nevatia ~\cite{zhao2003bayesian} used 3D human models to detect persons in the observed scene as well as a probabilistic framework for tracking extracted features from the persons. In contrast, some other methods track feature points in the scene using the well-known KLT algorithm \cite{rabaud2006counting,shi1994good}. Then, trajectories are clustered using space proximity. Such a clustering step helps to obtain a one-to-one association between individuals and trajectory clusters, which is quite a strong assumption seldom verified in a crowded scenario.

\emph{Holistic Approaches.} The holistic approaches, on the other hand, do not separately detect and track each individual/object in a scene. Instead, they treat the crowd as a single entity and try to exploit low/medium level features (mainly in histogram form) extracted from the video in order to analyze the sequence as a whole. Typical features used in these approaches are spatial-temporal gradients or optical flows. Krausz and Bauckhage ~\cite{krausz2012loveparade,krausz2011analyzing} employed optical flow histograms to represent the global motion in a crowded scene. The extracted histograms of the optical flow along with some simple heuristic rules were used to detect specific dangerous crowd behaviors.
More advanced techniques exploited models derived from fluid dynamics or other physics laws in order to model a crowd as an ensemble of moving particles. Together with Social Force Models (SFM), it was possible to describe the behavior of a crowd by means of interaction of individuals ~\cite{helbing1995social}. In ~\cite{mehran2009abnormal} the SFM is used to detect global anomalies and estimate local anomalies by detecting focus regions in the current frame. For abnormality detection, Solmaz and Shah ~\cite{solmaz2012identifying} proposed a method to classify the critical points of a continuous dynamical system, which was applied for high-density crowds such as religious festivals and marathons ~\cite{Authors22}.
In addition, several approaches deal with the complexity of a dynamic scene analysis by partitioning a given video in spatial-temporal volumes. In ~\cite{kratz2009anomaly,kratz2010tracking} Kratz and Nishino extract spatial-temporal gradients from each pixel of a video. Then, the gradients of a spatio-temporal cell are modeled using Spatial-Temporal Motion Pattern Models, which are basically 3D Gaussian clusters of gradients. Hart and Storck ~\cite{LDA} used group gradients observed at training time in separate cluster centers, then they used the Kullback-Leibler distance~\cite{LDA} in order to select the training prototype with the closest gradient distribution.
Mahadevan \emph {et al}. ~\cite{mahadevan2010anomaly} model the observed movement in each spatial-temporal cell using dynamic textures, which can be seen as an extension of PCA-based representations. In each cell, all the possible dynamic textures are represented with a Mixture of Dynamic Texture (MDT) models, allowing to estimate the probability of a test patch to be anomalous.

Similar to this work in terms of capturing commotion, in \cite{energy} an energy-based model approach has been presented for abnormal detection in crowd environments. In \cite{Dynamic} force field model has been employed in a hierarchical clustering framework to cluster optical flows and detect abnormality. Lu et al. \cite{examination} introduced a correlation-based measure across spatio-temporal video segments extracted by clustering. Authors in \cite{Crowd} selected a region of interest utilizing the motion heat map and measured the abnormality as the entropy of the frames.
 
 \qsection{Overview} In this section, we introduce an unsupervised context-dependent statistical commotion measure and an efficient way to compute it, to detect and localize abnormal behaviors in crowded scenes. For this purpose, the scene of interest is modeled as moving particles turned out from a tracklet algorithm, which can be viewed as a motion field distributed densely over the foreground. The particles are grouped into a set of \emph{motion patterns} (prototypes) according to their motion magnitude and orientation, and a \emph{tracklet binary code} is established to figure out how the particles are distributed over the prototypes. Here, a novel statistical \emph{commotion measure} is computed from the binary code for each video clip to characterize the commotion degree of crowd movement. 

We observed that when people behave irregularly in an area under surveillance, the abnormality measure value will increase remarkably and suddenly in comparison to normal cases. We evaluated our method extensively on different abnormality detection experimental settings ranging from frame-level, pixel-level to video-level on standard datasets. The qualitative and quantitative results show promising performance compared with the state-of-the-art methods.

\qsection{Contributions} Most of the related works only pay attention to either short-period motion information (e.g., optical flow) or long-term observation (e.g., trajectories), we instead employed tracklet as an intermediate-level motion representation. To our knowledge, there is little research on anomaly detection considering simultaneously the magnitude and orientation of the motion. The closest recent work to us is HOT~\cite{mousavi}. However, these two works, are different in two ways: i) we are working in an unsupervised setting while HOT is a supervised method. ii) We change the tracklet assignment in HOT by employing an efficient hash-based binary motion representation.
We specifically shortening the major contributions of this work as following. \textbf{First}, we propose Motion Patterns to represent the statistics of a tracklet at each frame in terms of magnitude and orientation. \textbf{Second}, we introduce Tracklet Binary Code representation to model the movement of salient points over its corresponding tracklet in both spatial and temporal spaces. \textbf{Third}, we introduce a new unsupervised measure to evaluate the commotion of a crowd scene at pixel, frame and video levels. We showed the superiority of the new measure for abnormality detection in video.

  \begin{figure}[t]
  	\centering
   \includegraphics[scale = 0.7]{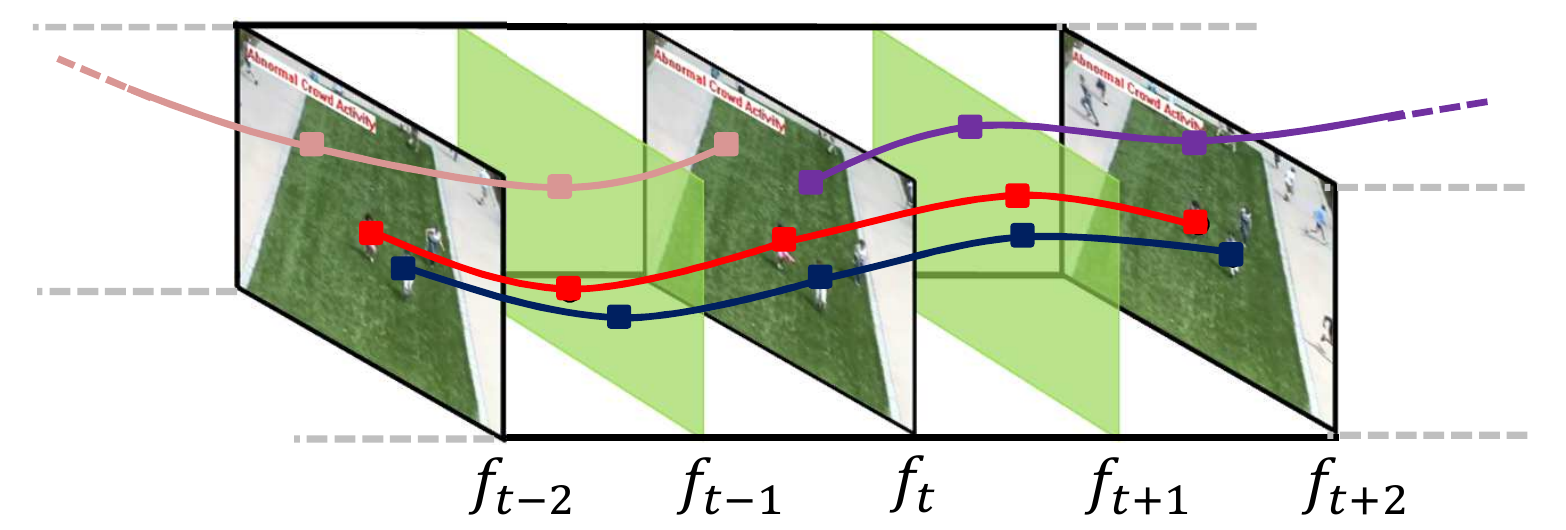}
    \caption{Four tracklets extracted from corresponding salient points tracked over five frames. }
 \label{fig:res1}
 \end{figure} 

\subsection{Tracklet: A Mid-level Representation for Crowd Motion} \label{sec:tracklet_mid}

The existing approaches for motion representation in crowds can be generally classified into two main categories: local motion based (e.g., optical flows) \cite{wang2009unsupervised,li2008global,li2008scene,hospedales2009markov,mehran2009abnormal,kim2009observe} and complete trajectories of objects \cite{makris2005learning,hu2006system} based. Both have some limitations. Without tracking objects, the information represented by local motions is limited, which weakens the models power. The crowd behavior recognized from local motions are less accurate, tend to be in short range and may fail in certain scenarios. The other type of approaches assumed that complete trajectories of objects were available and crowd video can be represented using the point trajectories. The accurate long-term observation is hard to be guaranteed due to scene clutter and tracking errors, but can effectively capture the global structure of crowd motions \cite{zhou2011random}.

Tracklets \cite{tracklets}, however, are mid-level representations between the two extremes discussed above. A tracklet is a fragment of a trajectory and is obtained by a tracker within a short period. Tracklets terminate when ambiguities caused by occlusions and scene clutters arise. They are more conservative and less likely to drift than long trajectories \cite{zhou2011random}. In another words, Tracklets represent fragments of an entire trajectory corresponding to the movement pattern of an individual point. They are generated by frame-wise association between point localization results in the neighboring frames.

More formally, a tracklet is represented as a sequence of points in the spatio-temporal space as:
\begin{equation}
\mathbf{tr}= \left ( p_{1},\cdots p_{t},\cdots p_{T} \right )\
\end{equation}
where each $p_{t}$ represents two-dimensional coordinates $\left (  x_{t}, y_{t} \right )$ of the $t^{th}$ point of the tracklet in the $t^{th}$ frame and $T$ indicates the length of each tracklet. Tracklets are formed by selecting regions (or points) of interest via a feature detector and by tracking them over a short period of time. 
Thus, we emphasize that tracklets represent a trade-off between optical flow and object's long trajectory.


\qsection{Implementation} There are two different strategies for tracklet extraction~\footnote{OpenCV code is available at http://www.ces.clemson.edu/~stb/klt/}. 

In the first strategy, \emph{video-level initialization}~\cite{wacv}, tracklets are initialized using the salient points detected in the first frame of the video, and then tracked until the tracker fails. If tracker dose not work well to detect all salient points, it consequently is not capable of extracting corresponding tracklets for abnormality detection. \\

In the second strategy, \emph{temporally dense Tracklets}, the salient points in each single frame of the video is re-initialized and track the points over $T$ frames. This strategy is not limited to the points detected at the first frame and is capable of detecting all possible salient points over a given video. In other words, no matter how long is the captured video, this strategy is able to detect the salient points of all the appearing objects/individuals over the time. This results in producing a large pool of tracklets $\mathcal{T}$ which can be used to summarize the motion-patterns observed in the scene per each frame. 

We note that the second strategy explained above is first proposed in this work and will be explained in detail in the next section.

  \begin{figure}[t]
  	\centering
   \includegraphics[scale = 0.7]{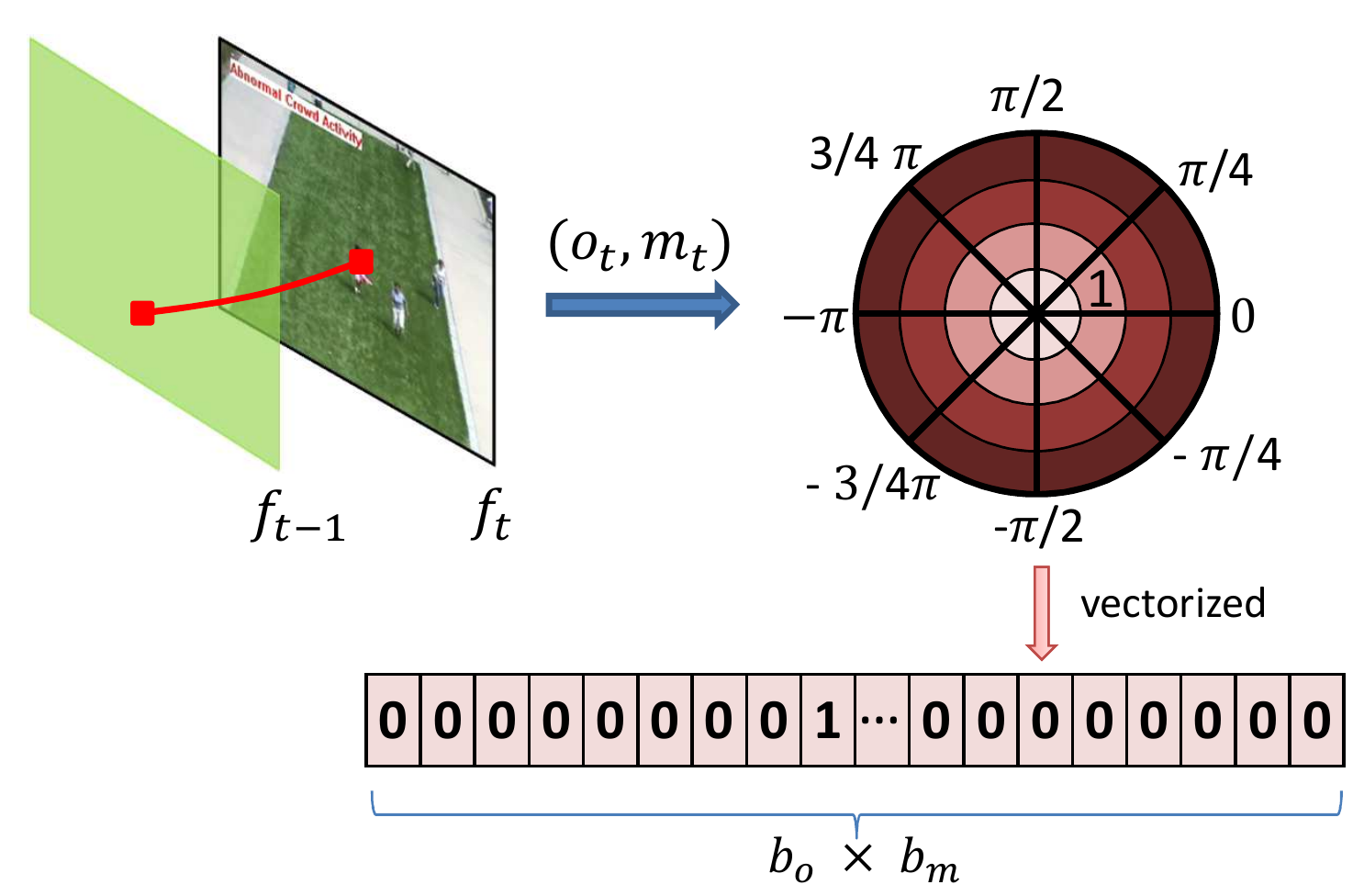}
    \caption{A polar histogram of magnitude and orientation of a salient point at the $t$-th frame of a tracklet (motion pattern).}
 \label{fig:res2}
 \end{figure}

  \begin{figure}[t]
  	\centering
   \includegraphics[scale = 0.6]{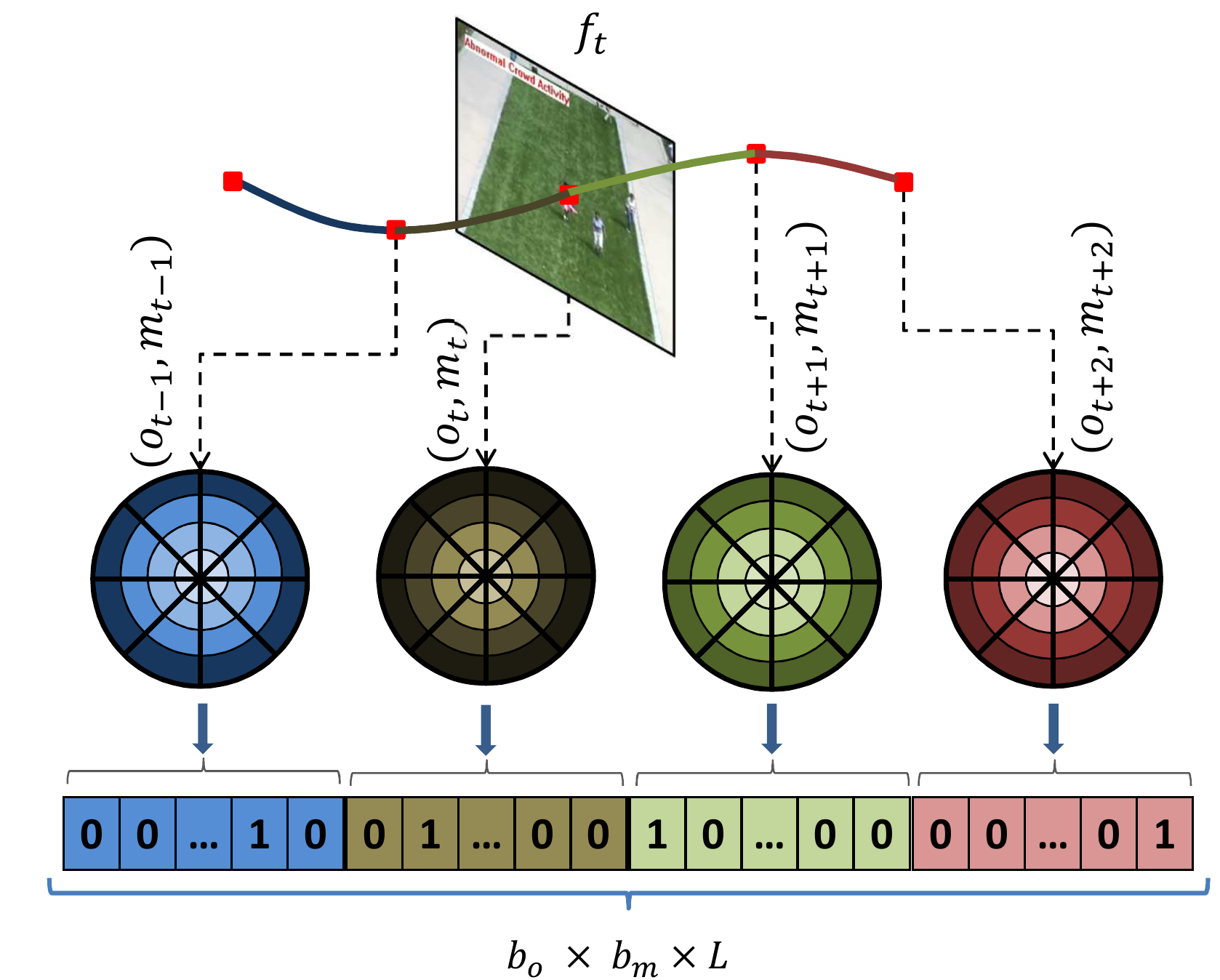}
    \caption{Tracklet binary code is constructed by concatenating a set of motion patterns corresponding salient point over $L+1$ frames.}
 \label{fig:res3}
 \end{figure}

\subsection{Tracklet-based Commotion Measure}      
               
In this section, we explain the proposed pipeline to measure abnormality of a given video $ \mathbf{v} = \{ f_t \}_{t = 1}^{T} $ with $T$ frames. 
 
\qsection{Tracklet Extraction} The first step involves extracting all tracklets of length $L+1$ in video $ \mathbf{v} $. Following the two types of tracklet extraction explained in previous section, we select the second strategy (i.e., temporally dense Tracklets). Towards this purpose, SIFT algorithm is first applied to detect salient points at each frame $f_{t}$~\cite{lowe2004distinctive}. Then a tracking technique (we employed the KLT algorithm ~\cite{Authors23}) is used to track each salient point over $ L + 1 $ frames. Tracklets whose length is less than $L+1$ are considered as noise and, thus, eliminated. The output is a set of tracklets $\mathcal{T} =  {\left \{ tr^n\right \}_{n=1}^{N}}$, where $N$ is the number of extracted tracklets and $tr^{n}$ refers to the $n$-th tracklet. Figure~\ref{fig:res1} illustrates a video example and four tracklets which are computed by tracking a set of corresponding salient points over a sequence of five frames. Following the two strategy 
 
\qsection{Motion Pattern} Each tracklet $tr^{n}$ is characterized by a set of spatial coordinates of its corresponded salient point tracked over $L+1$ frames $\{ ( x_{l}^{n} , y_{l}^{n})  \}_{l=1}^{L+1}$. These spatial coordinates are employed to compute the motion orientation and magnitude of the salient point at $l$-th frame as:
  
 \begin{eqnarray}
o_{l}^{n} & = & \arctan\frac{(y_{l}^{n}-y_{l-1}^{n})}{(x_{l}^{n}-x_{l-1}^{n})}
\label{eq:eg1} \\
m_{l}^{n} &  = &  \sqrt{\left( x_{l}^{n}-x_{l-1}^{n}\right)^2+\left( y_{l}^{n}-y_{l-1}^{n}\right)^2} 
\label{eq:eg2}
\end{eqnarray}  
where $ 2 \leq l \leq L+1$. This step computes a temporary ordered set of $L$ orientations and magnitudes of the salient point corresponded to $n$-th tracklet $ \{  ( o_{l}^{n}, m_{l}^{n} )  \}_{l=1}^{L} $ (we reset $l = 1$ for simplicity).

The motion orientations and magnitudes $ \{  ( o_{l}^{n}, m_{l}^{n} )  \}_{l=1}^{L} $ are used to form a histogram representation of the $n$-th tracklet. First, a polar histogram $h_{l}^{n}$ is computed using the orientation and magnitude of the $n$-th tracklet at frame $l$, $( o_{l}^{n}, m_{l}^{n} )$. This can be easily done by a simple hashing function in $\mathcal{O}(1)$ whose input is $( o_{l}^{n}, m_{l}^{n} )$ and returns a binary polar histogram with only one "1" value at sector  $( o_{l}^{n}, m_{l}^{n} )$ and zeros for the rest. The polar histogram then is vectorized to a vector of length $ b_{o} \times b_{m} $, where $b_{o}$ and $b_{m}$ are respectively the number of quantized bins for magnitude and orientation.  This is illustrated in Figure~\ref{fig:res2} . The color spectrum of each sector indicates the quantized bin of magnitude. Each arc represents the quantized bin of orientation. We called each vectorized $h_{l}^{n}$ a motion pattern. 


 \qsection{Tracklet Binray Code} Given  a set of orientations and magnitudes, $ \{  ( o_{l}^{n}, m_{l}^{n} )  \}_{l=1}^{L} $, we can correspondingly compute $L$ motion patterns $ \{ h_{l}^{n}  \}_{l=1}^{L}  $ for the $n$-th tracklet. Finally, all the (vectorized) motion patterns $ \{ h_{l}^{n}  \}_{l=1}^{L}  $ are concatenated to compute a tracklet histogram $ H^{n} = [h_{1}^{n},...,h_{L}^{n}]^{\top} $ of length $ b_{o} \times b_{m} \times L $ ($\top$ is transpose operator). $H$ is referred to as tracklet binary code, Figure~\ref{fig:res3}.
  
 \qsection{Commotion Measuring} To compute commotion measure, each frame $f_t$ is divided into a set of non-overlapped patches $ \{ p_{i}^{t} \} $, where $i$ indexes the $i$-th patch in the $t$-th frame. For each patch $p_i^t$, a subset of tracklet binary codes is selected from $\{ H^{n} \}_{n=1}^N$ whose corresponding tracklets spatially pass from patch $p_i^t$, and $p_i^t$ is temporally located at the middle of the selected tracklets (i.e. if the length of a tracklet is $L+1$, tracklets which start/end $L/2$ frames before/after frame $t$ passing from patch $p_i^t$ are selected). Suppose that $N_p$ tracklet motion codes are selected for patch $p_i^t$ denoted by $ \{ H^{n_p} \}_{n_p=1}^{N_p}$. Then, we statistically compute the aggregated tracklet binary code for patch $p_i^t$ as $ \mathcal{H}_i^t = \sum_{n_p = 1}^{N_p} H^{n_p}$. The aggregated histogram $ \mathcal{H}_i^t$, which contains the distribution of motion patterns of $p_i^t$, is used to compute the commotion assigned to patch $p_i^t$ as:
\begin{equation}\label{comMes}
Comm(p_i^t) = \sum_{j=1}^{ |\mathcal{H}_i^t| } w(j, j_{max})\times || \mathcal{H}_i^t(j) - \mathcal{H}_i^t(j_{max}) ||_{2}^2
\end{equation}
where $|.|$ returns the length of vector and $j_{max}$ indicates the index of maximum value in $\mathcal{H}_i^t$ (i.e. $\mathcal{H}_i^t(j_{max})$ is the maximum value in $ \mathcal{H}_i^t $). $||.||_2$ is the L2-norm. As mentioned earlier, $\mathcal{H}$ captures the motion patterns distribution of tracklets passing from a sampled patch. As a result, the maximum value of $\mathcal{H}$ indicates the dominant motion pattern over the patch of interest. The amount of commotion of a patch, therefore, can be measured by the difference (deviation) between the occurrences of the dominant motion pattern and the other motion patterns wighted by $w(j, j_{max})$.

\begin{figure*}[t]

 \begin{minipage}[b]{1.0\linewidth}
 	\centering
 	\centerline{\includegraphics [scale = .6] {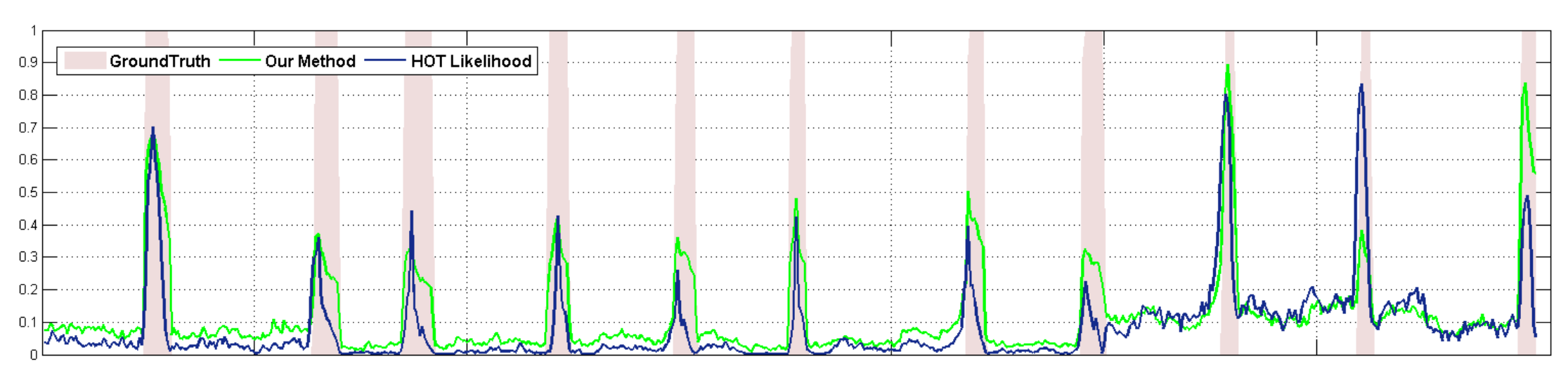}}
 \end{minipage}
     \caption{Results on UMN dataset. The blue and green signals respectively show the commotion measure computed by our approach and LDA+HOT over frames of 11 video sequences. The pink columns indicate the abnormal frames of each sequence. Each sequence starts with normal frames and ends with abnormal frames.}
 \label{fig:umn}
 \end{figure*} 

$w(j, j_{max}) \in [0,...,1]$ is a scalar weight which controls the influence of the $j$-th motion pattern on the commotion measure respect to the dominant ($ j_{max}$-th) motion pattern. The motivation behind using $w$ is to assign higher(lower) weights to pattern which are less(more) similar to the dominant pattern. The weight $w(j, j_{max})$ is defined using a two-variants Gaussian function as:		
	\begin{equation}	
	w(j, j_{max})  = \dfrac{1}{2 \pi \sigma_o \sigma_m} e^{   -   \dfrac{(\bar{o}_j - \bar{o}_{j_{max}})^2}{2 \sigma_o^2} -  \dfrac{(\bar{m}_j - \bar{m}_{j_{max}})^2}{2 \sigma_m^2}    }
	\end{equation}
	where $\bar{o}_j$ is the middle of the orientation bin that the $j$-th motion pattern belongs to. For example, if the $j$-th motion pattern falls in $[0-\pi /4]$, then $\bar{o}_j$ is $ \pi /8$. Similarly, $\bar{m}_j$ is the middle of the magnitude bin that the $j$-th motion pattern falls in (e.g. if the $j$-th motion pattern falls in $[3-6]$, then $\bar{m}_j$ is $ 9/2$). The definitions can be identically apply for $\bar{o}_{j_{max}}$ and $\bar{m}_{j_{max}}$. The values of $\sigma_o$ and $\sigma_m$ are set to $ 1/b_{o}$ and $1/b_{m}$.

  \begin{figure}[t]
  	\centering
   \includegraphics[scale = .4]{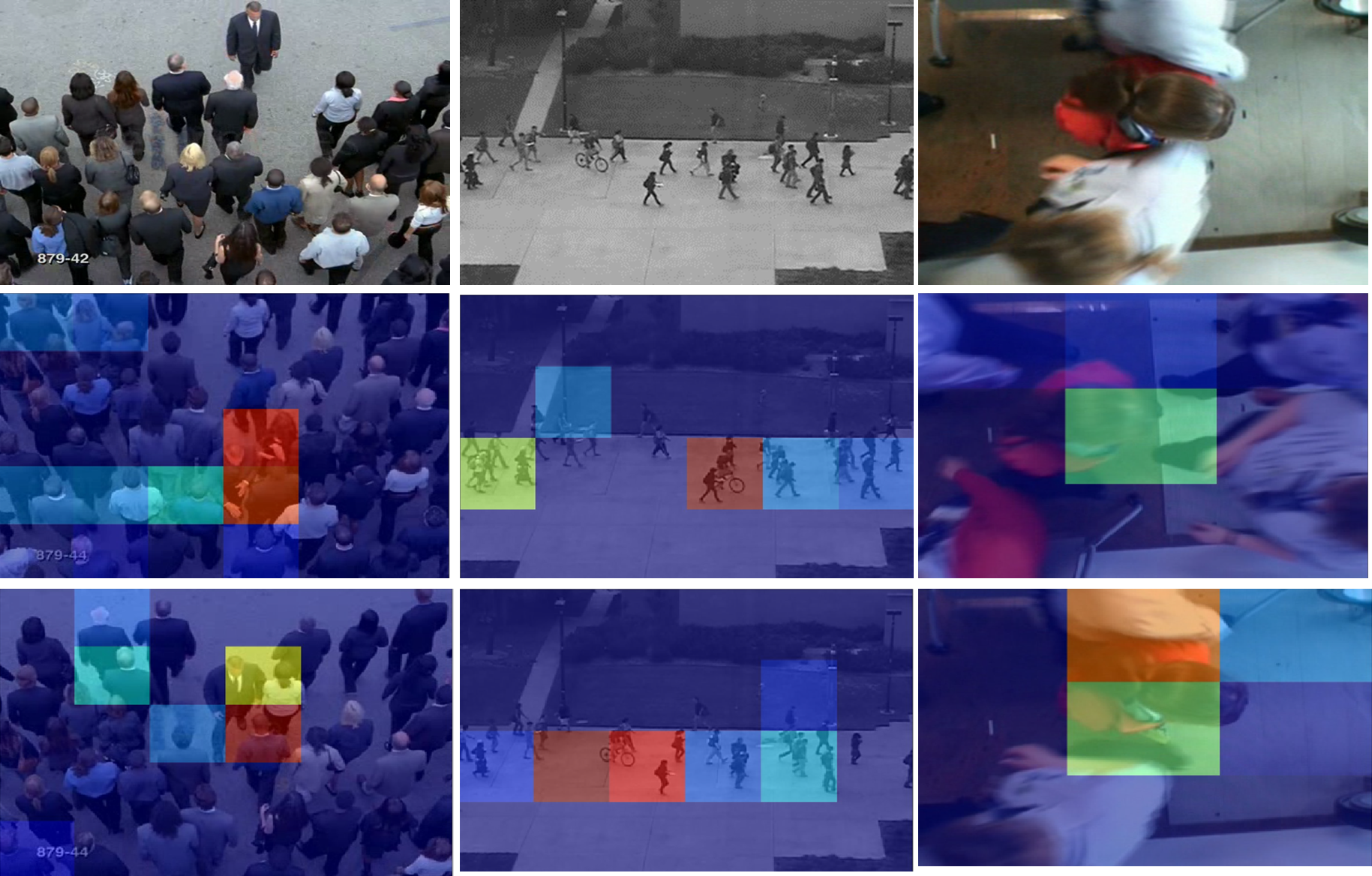}
    \caption{Qualitative results on sample sequences selected from SocialForce, UCSD and York datasets. The commotion measure of each patch is represented by a heat-map. }
 \label{fig:twod}
 \end{figure} 

\subsection{Experiments} 
In this section, we validate the proposed method in all three settings for abnormality detection including (i) \emph{pixel-level}, (ii) \emph{frame-level} and (iii) \emph{video-level}.\\

\qsection{Pixel-level} In this experiment, we evaluated our approach qualitatively on a subset of video sequences selected from standard datasets (UCSD~\cite{Authors14}, SocialForce\cite{sfm} and York~\cite{zaharescu2010anomalous}). For a given video, we first extract a set of tracklets of length $L = 10$. The magnitude and orientation bins are set to 5 and 6 respectively to form the polar histogram (motion pattern). Then each frame is divided to a set of non-overlapped patches in which for each patch the commotion measure is computed using Eq.~\ref{comMes}. Qualitative results are shown in Figure~\ref{fig:twod} in terms of heat-map respect to the locally computed commotion measure. The selected sequences characterized by different magnitude and orientation, camera view points, different type of moving objects (e.g. human and bike) over scarce-to-dense crowded scenarios. As illustrated in Figure~\ref{fig:twod} the proposed measure can be effectively exploited for abnormality localization along with a per-defined threshold. Furthermore, the commotion measure can be used as spatial-temporal interest point for video level abnormality detection (details in video level experiment).

 \begin{figure}[t]
 \begin{minipage}[b]{1.0\linewidth}
   \centering
	\includegraphics[width=0.6\linewidth]{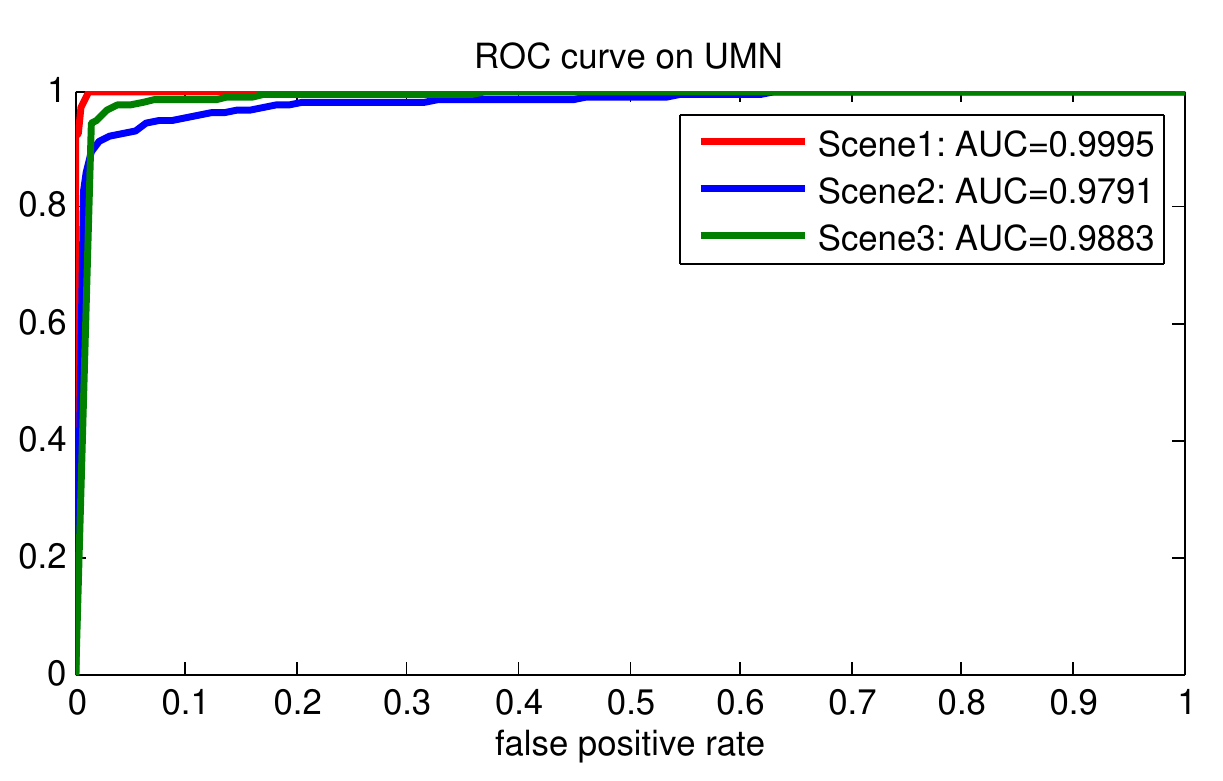}
 \end{minipage}
 \caption{AUC on 3 scenes of UMN.}
 \label{fig:umn_2}
 \end{figure}

\qsection{Frame-level} The goal of this experiment is to compute a single commotion measure for each frame of a given video. Toward this purpose, we modified the procedure of computing the commotion measure in three ways: First, each whole frame is considered as a single patch (there is not frame patching). Second a commotion measure is computed for each tracklet passing over the frame of interest. Finally, The measures computed from all the tracklets are summed up as the frame's commotion measure. We evaluated our method on UMN dataset~\footnote{http://mha.cs.umn.edu/movies/crowdactivity-all.avi.} including 11 sequences filmed in 3 different scenes. Figure~\ref{fig:umn} shows commotion measure computed for each frame and illustrated as a signal (Green). We compare our method with LDA log-likelihood on HOT, selected as a baseline measure (blue). Since the HOT approach is a supervised technique, unlike the new approach which is unsupervised, these two techniques are not directly comparable. Thus, we divided the dataset into two subsets of video sequences (e.g., A and B). We performed training and testing two times, at each time, a subset (A or B) is selected for training and the other one (B or A) for testing.  The LDA log-likelihood for each frame was considered as its commotion  measure. Obviously, both approaches perform well and assign lower(higher) commotion measures to normal(abnormal) frames. The difference, however, is that our approach is unsupervised. This is an supreme characteristic for the task of abnormal detection where in most cases there is not a clear definition of abnormality and gathering real-world training videos are intractable. We also obtained the scene-based ROC of our proposed method illustrated in Figure~\ref{fig:umn_2} and the overall Area Under ROC (AUC) in Table~\ref{tab:umn_scene} comparing with the the leading existing approaches. According to Table~\ref{tab:umn_scene}, our approach achieved the superior detection speed (in terms of frame per second) with very competitive detection performance. 

\begin{table}
\centering
		\begin{tabular}{|c|c|c|c|}
			\hline
			Method & AUC &  Speed (fps)\\
			\hline
			Optical Flow \cite{sfm}  & 0.84 & 5\\
			SFM\cite{sfm}  & 0.96 & 3\\
			Chaotic Invariants \cite{wu2010chaotic}  & 0.99& 1\\
			Sparse Reconstruction  \cite{cong2011sparse}& 0.978 & \textless 1\\
				\hline
			Proposed Scheme  & 0.9889& 5\\
			\hline
		\end{tabular}
	\caption{Performance of the proposed scheme on UMN.}
	\label{tab:umn_scene}
\end{table}

\qsection{Video-level} In this experiment, we show the effectiveness of the proposed measure when employed as spatio-temporal interest point detector along with one of the best performing descriptors (HOT). In this setting, we adopt the commotion measure to be used in a video-level abnormality detection on Violence-in-Crowd dataset~\cite{conf/cvpr/HassnerIK12}. For this purpose, we first apply a spatio-temporal grid on the video, then for each 3D cell of the grid we compute our proposed measure. For evaluation, we deployed the standard BOW representation pipeline used in most video level works \cite{mousavi,sodemann2012review}. 
We enriched the standard setting with a weight vector comes from commotion measure. We simply defined a weight vector with the same length of codebook's size. The weight of each codeword is computed as summation over the commotion measures of the 3D cells belong to its corresponding cluster. Our result outperformed all previous methods including HOT as reported in Table~\ref{tab:tableV}.


 \subsection{Conclusion} \label{sec:con_motion}

The problem of abnormality detection in crowd scenes was addressed in this section to show the effectiveness of utilizing motion-based video representations. We employed tracklet as an intermediate-level motion representation in this work. Specifically, we proposed a new measure to compute the commotion of a given video and we showed that the new measure can be effectively exploited to detect/localize abnormal events in crowded scenarios. The qualitative and quantitative results on the standard dataset show that our approach outperformed the state of the arts in terms of detection speed and performance. The future direction involves quantitatively evaluate the new approach on more challenging datasets in the pixel level. Moreover, exploring of the proposed approach for the task of action recognition would be a potential direction.

 \begin{table}
 	\begin{center}
 		\begin{tabular}{|c|c|}
 				\hline
 			Method & Accuracy  \\
 			\hline
 			Local Trinary Patterns \cite{LTP}  & 71,53\% \\
 			Histogram of oriented Gradients \cite{4587756} & 57,43\% \\
 			Histogram of oriented Optic-Flow  \cite{HOOF} & 58,53\%   \\ 
 			HNF  \cite{4587756} & 56,52\% \\
 			Violence Flows ViF \cite{conf/cvpr/HassnerIK12} & 81,30 \% \\
 			Dense Trajectories \cite{denseTraj} & 78,21 \% \\
 			HOT   \cite{mousavi} & 78,30\% \\
 			\hline
 			Our Method & \textbf{81.55}\% \\
 			\hline
 		\end{tabular}
 	\end{center}
 	\caption{Classification results on crowd violence dataset, using linear SVM in 5-folds cross validation. }
 	\label{tab:tableV}
 \end{table}      


\section{Appearance-based Video Representation} \label{sec:app}

In this section, we study acquiring appearance-based video representation for high-level human behavior understanding. The mid-level representation is provided as a result of exploiting the dynamics of a given dictionary of patch-based models (e.g., Poselets \cite{bourdev2009poselets}) in time, for \emph{detecting and recognizing the collective activity of a group of people} in crowd environments. Specifically, we present a novel semantic-based spatio-temporal descriptor which can cope with several interacting people at different scales and multiple activities in a video. Our descriptor is suitable for modelling the human motion interaction in crowded environments -- the scenario most difficult to analyse because of occlusions. In particular, we extend the Poselet detector approach by defining a descriptor based on Poselet activation patterns over time, named \emph{TPOS}. We will show that this descriptor followed by a classic classification method (i.e., bag of words + SVM) can effectively tackle complex real scenarios allowing to detect humans in the scene, to localize (in space-time) human activities,  (so performing segmentation of the video sequence),
and perform collective group activity recognition in a joint manner, reaching state-of-the-art results.

\subsection{Introduction}

Understanding human behavior is a problem whose solution has a clear impact in modern Computer Vision applications. In general, people activities convey rich information about the social interactions among individuals, the context of a scene, and can also support higher-level reasoning about the ongoing situation.
In such regard, much attention has been recently posed in the detection and recognition of specific human activities from images and videos, mainly focusing on crowded scenes. Such task has been formalized in the literature as a classification problem where a label corresponding to a collective activity has to be assigned to a specific video frame or a video clip, possibly also identifying the spatial location in the video sequence where such activities occur.

This section addresses this open issue and aims at proposing a spatio-temporal descriptor called TPOS, based on Poselets \cite{bourdev2009poselets}. This descriptor is effective in detecting in space and time \emph{multiple} activities in a \emph{single} video sequence, so providing a semantically meaningful segmentation of the footage, without resorting to elaborated high-level features or complex classification architectures.
In the literature, apart from the core classification aspects, a substantial debate has been posed over the type of features which are more discriminative for the activity detection/recognition problem.
In this context, two classes of approaches can be identified, which are related to the level of semantics embedded in the descriptor. 

On one hand, \emph{feature-based} methods adopt the classical strategy of detecting first a set of low-level spatio-temporal features,
followed by the definition of the related descriptors. These descriptors should be representative of the underlying activity and they are typically defined as a spatio-temporal extension of well-known 2D descriptors, such as 3D-SIFT \cite{scovanner20073}, extended SURF \cite{willems2008efficient}, or HOG3D \cite{klaser2008spatio}.
Among the best performing features, we can quote the Laptev's space-time interest points (STIP) \cite{laptev2005space},
the cuboid detector \cite{dollar2005behavior} and descriptor based on Gabor filters \cite{bregonzio2009recognising}.
A number of other descriptors also deserves to be mentioned like dense trajectories \cite{wang2011action}, spatial-time gradient \cite{laptev2008learning}, optical flow information \cite{dollar2005behavior}, and Local Trinary Patterns \cite{yeffet2009local}.
In \cite{le2011learning}, an unsupervised spatio-temporal feature learning scheme is proposed which only uses the intensity value of the pixels in an extended ISA (Independent Subspace Analysis) framework. An interesting comparative evaluation was presented in \cite{wang2009evaluation}, which reports a performance analysis of different combinations of feature detectors and descriptors in a common experimental protocol for a number of different datasets.

On the other hand, a different set of methods, named \emph{people-based} approaches, directly use higher-level features  which are highly task-oriented, i.e. they are tuned to deal with people. They rely on a set of video pre-processing algorithms aimed at extracting the scene context, people positions (bounding boxes, trajectories) and head orientations. For instance, the context around each individual is exploited by considering a spatio-temporal local (STL) descriptor extracted from each head pose and the local surrounding area \cite{choi2009they}. The Action Context (AC) descriptor is introduced in \cite{lan2010retrieving} in a similar way as STL, but it models action context rather than poses. In \cite{lan2010beyond,lan2012discriminative}, two new types of contextual information, individual-group interaction and individual-individual interaction are explored in a latent variable framework. The Randomized Spatio-
Temporal Volume (RSTV) model, a generalization of STL method, is introduced in \cite{choi2011learning} by considering the spatio-temporal distribution of crowd context.  Khamis et al. \cite{khamis2012flow} presented a framework by solving the problem of multiple target identity maintenance in AC descriptors. Recently, Choi and Savarese \cite{choi2012unified} put a step forward presenting a more extensive model which estimates collective activity, interactions and atomic activities simultaneously. Finally, \cite{khamis-eccv2012} introduced a model which combines tracking information and scene cues to improve action classification. 
The comprehensive surveys in \cite{aggarwal2011human,ziaeefar2015semantic} report an excellent overview of recent and past works in this domain.

Both classes of approaches have their expected pros and cons which are more evident for ``in the wild''  scenarios. In particular, people-based methods highly rely on the accuracy of sophisticated detectors that might fail in the case of dense (i.e., crowded) scenarios because of the presence of occlusions, or due the impossibility to reliably deal with the spatio-temporal correlation among the large number of targets.
However, in case of accurate detections and tracking, they have shown to obtain the best performance. This is mainly due to the significant semantic information provided by the descriptors which is tailored to the task of human sensing.
\\
Instead, feature-based approaches utilize low-level spatio-temporal representations regardless the scene complexity, and thus they are less prone to gross misinterpretations as given, for instance, by false positive person detections eventually obtained by people-based methods.
However, low-level features lack of an actual semantic interpretation since the identity and the collective human parts peculiar of the interaction are typically disregarded. In other words, such features are extracted at the whole frame level, and not necessarily correspond to high-level descriptions of a spatially and temporally localized activity, or are related to local peculiar parts of the image.
In practice, for the collective activity recognition goal, it is difficult to reason about activities of a group of people while interacting, because the local spatio-temporal features do not correspond to high-level descriptions of the activity according to a global human observation.
\vspace{0.3cm}
\subsection{Discussion and contributions}

In order to overcome these limitations, in this section we propose a method to narrow down the gap between feature- and people-based approaches, namely, trying to inject a semantic value to spatio-temporal features, so leading to more powerful discrimination of human activities while maintaining the method complexity to a manageable level.
We introduce here a novel semantic descriptor for human activity as a temporal extension of the Poselet approach \cite{bourdev2009poselets}, in which human, semantically meaningful body parts and their motion
are modelled by poselets activations in time.  In a nutshell, we devised a \emph{temporal poselet} ($\vec{TPOS}$) descriptor by analyzing the activation correlation of a bank of (poselet) detectors over time.
This provides a video representation composed by joint human detections and activity characterization using the basis of the detection activation patterns of the poselets, allowing to detect humans and characterize their activities in a joint manner using as basis the detection activation patterns of the poselets.

\begin{figure}[t]
\begin{center}
\fbox{\includegraphics[width=0.8\linewidth]{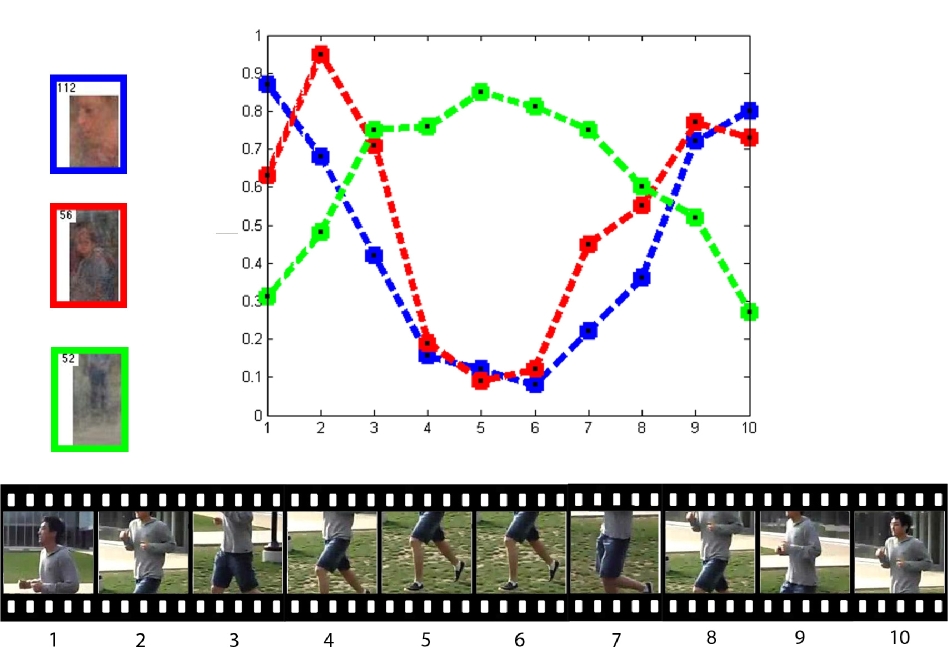}}
\end{center}
  \caption{Three types of poselet activations in time in a 10-frame video (bottom): head, torso and legs in the sequence are displayed as blue, red and green profiles, respectively, in a 10-frame sequence, showing the correlation of these types of poselets during the ``running'' activity.}
\label{fig:tpos_resume}
\end{figure}

Like\cite{maji2011action} we define poselet activations based on their detection. For this purpose we represent each patch of an image as intersection divided by union between the patch and the bounding box of a detected poselet on that image. The underlying idea of the proposed approach is that, since each poselet is activated by a particular static human pose, the collection of poselet activations in time can capture and characterize human motion. In other words, by assuming that people activities can be described as a sequence of peculiar body poses, they can be represented as a sequence of activations of a set of human body part detectors. As an example, in Figure \ref{fig:tpos_resume}, we show the activations of three different poselets (face, torso and legs) in a $10$-frame video block of a video sequence representing a running person. It is worth to note that the specific sequence of appearing/disappearing of the human legs, torso and face in this short video clip is encoded by corresponding activations of the leg, torso and face poselets, respectively. So, the activation profiles of these three poselets suggest that human activity is correlated with temporal profiles of body parts and that can be learned for a discriminative analysis.

Similar to our proposal, in certain (partial) aspects, a few recent works on activity recognition are worth to be mentioned. The closest works to us are \cite{maji2011action,nabi2012human} where poselet detector activations have been used for activity recognition in still images. The idea is similar to ours, that is to build a robust descriptor for actions considering activation of poselets.  In their work the feature level action discrimination was caught by re-training a large dictionary of 1200 action-specific poselets. This set of detectors were trained in the same dataset and tuned for specific action classes. In our case, first we deal with video sequences instead of still images, then, we aim at capturing action discrimination at a representation level and model the pose dynamics of a group of people using the basis pattern of poselet co-occurrence. So, we instead use the activation score of an outsourcing bank of detectors ($150$ general purpose poselets), as learned in the original formulation \cite{bourdev2009poselets}, and we tested on video datasets, which results in a much more challenging scenario.

Poselets, together with other features, are also used in \cite{yao2011human} for action recognition in still images. Here, action attributes and action parts are defined as verbs, objects and poselets, and are considered as bases to be combined to reconstruct, i.e., identify, an action. Poselets in this case are mainly used as in the original formulation by \cite{maji2011action} as a part of a larger set of basis blocks able to provide semantic meaning to an action.
Very recently, activities have been modelled as a sparse sequence of discriminative keyframes, that is a collection of partial key-poses of the actors in the scene \cite{Raptis:Sigal:2013}. Keyframes selection is cast as a structured learning problem, using an \emph{ad hoc} poselet-like representation trained on the same dataset used for testing (still using separate sets), and reinforcing the original poselet set using HOG features with a bag of words (BoW) component. This approach has interesting peculiarities and reaches good performance on the UT-Interaction dataset. Nevertheless, it uses specific descriptors learned from the same dataset and, despite it is claimed to be able to spatially and temporal localize actions and action parts, this is reached only in terms of the keyframes extracted, and it still has the limit to classify an activity per test video sequence.
Instead, our approach is able to deal with multiple activities in a single video sequence, while spatially and temporally localizing each collective activity performed by people, resulting in a semantically meaningful video representation.

The peculiar aspect of using activation patterns of a bank of detectors as descriptor is that it enables to perform feature detection and description in a unified framework. In our case, we define a \emph{saliency} measure as simply the density of our temporal poselet descriptor on each video blocks a sequence can be subdivided, so we select video blocks with high density of poselets activations as significant of the human activity. Our experimental evaluation shows that this measure can effectively be used for \textbf{group} detection and \textbf{collective} activity recognition in the wild.

The main peculiar aspect of our approach lies in the design of a new, yet simple, feature descriptor which results to be expressive for detecting people and to characterize their collective activity.
In particular note that our descriptor is not customized (i.e., ``tuned'') on any specific dataset (for training and testing), but the basic (poselet) detectors are used as given in \cite{bourdev2009poselets}, and tested in completely different datasets.
Moreover, our particular formulation potentially allows to deal with different activities in a single video sequence, so promoting its use for segmenting continuous video streaming providing temporal and spatial localization of the semantically meaningful events of the video sequence.
These characteristics make TPOS unique in the panorama of the spatio-temporal descriptors for human detection and collective activity recognition. It is also worth to note that single human actions are out of the scope of this work and our approach is devoted to the analysis of groups of people and crowds in time.

We tested the $\vec{TPOS}$ method on two public datasets \cite{choi2011learning,choi2012unified} and our experimental evaluation will show that it can effectively be used for group detection and collective activity recognition in the wild, outperforming the state-of-the-art baseline methods and showing the limitations of the pure feature-based methods. 

The remainder of the report is structured as follows.  The temporal poselet approach is presented in Section \ref{sec:temporal_poselet} together with its application to the people group detection. Section \ref{sec:GAR} shows the strength of the semantic representation of the temporal poselets for addressing the group activity classification problem. Experiments in Section \ref{sec:experiments} will evaluate the proposed approach on two different datasets \cite{choi2011learning,choi2012unified}. Finally Section \ref{sec:conclusions} will draw the direction for future work and further application domains.

%
%
%
%

\subsection{Temporal Poselets} \label{sec:temporal_poselet}

Our final goal is to detect the collective activity of a group of people in weakly labelled videos. In this section, we will formalize our poselet-based temporal descriptor to be used for two applications, group detection and collective activity recognition.

\qsection{Poselet-based video representation} We first derive the descriptor in 2D and then we follow with the extension in time. A generic image frame is first partitioned in a set of $N_h \times N_w$ grid cells. the overall set of image cell is represented by the set $\cal G$ such that:
$$
{\cal G} = \{ \vec g_k \}_{k=1}^{|{\cal G}|} \:\:\:\: \text{and} \:\:\:\:
{\vec g}_k = \{ \vec z_k,  w, h \}, 
$$
where $\vec z_k \in \mathbb{N}^2$ represents the coordinates of the top-left corner of the grid cell, $w$ and $h$ the constant grid cell width and height, respectively and $|{\cal G}| = N_h \times N_w$. Following this notation, a given cell grid $k$ is fully specified by $\vec g_k$.

\begin{figure}[t]
\begin{center}
\includegraphics[width=8.0cm]{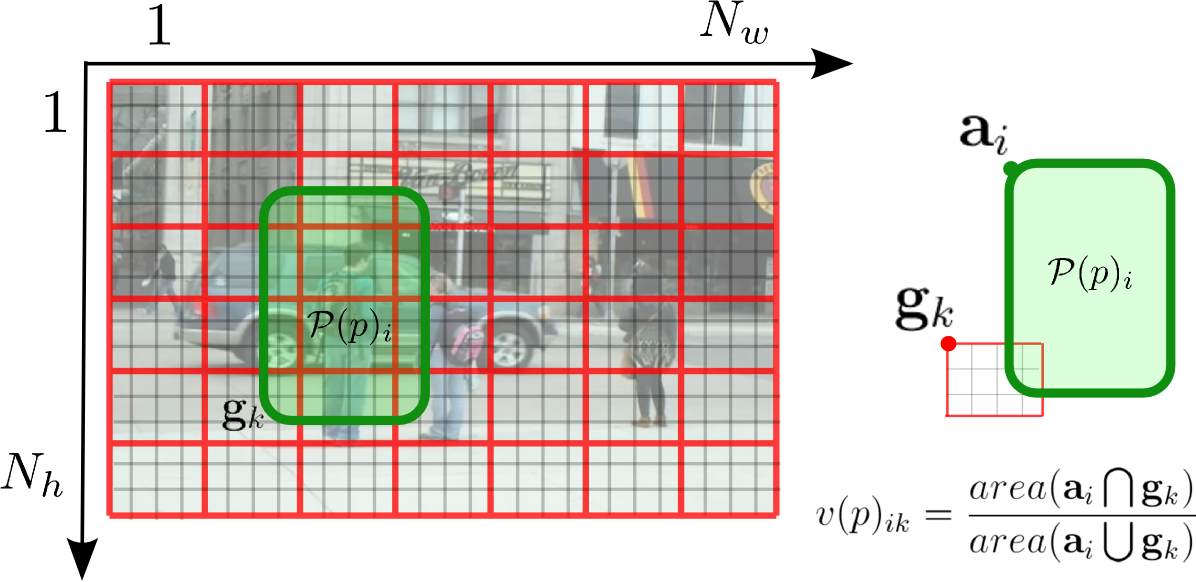} \\
\hspace{2cm} a) \hspace{3.5cm} b)
\end{center}
\caption{a) The image is partitioned using a regular grid (in red) where each cell grid is defined by ${\cal G}_j$ containing the 2D position and the cell grid size. A poselet $p$ with activation $i$ (in green) is defined in the same manner with ${\cal P}(p)_i$. Notice here that the same poselet activation may intersects and/or include several cell grids in the image. b) The activation for a cell $g_k$ is given by the intersection between the poselet ${\cal P}(p)_i$ bounding box and the grid.}
\label{fig:poselet_2D}
\end{figure}

Given an image ${\cal I}_f$ we run $P$ poselet detectors(already trained on H3D\cite{bourdev2009poselets} and PASCAL)
as a filter bank. This provides the location of the detected poselets together with their bounding box size.
In particular, for each poselet detector $p$ with $p = 1 \:\:\ldots\:\:P$, we obtain the set of poselet detection ${\cal P}$ such that:
%
$$
{\cal P}(p) = \{ \vec a_i \}_{i=1}^{|{\cal P}(p)|} = \{ \vec l_i, w_i, h_i, c_i \}_{i=1}^{|{\cal P}(p)|}
$$
where $\vec l_i \in \mathbb{N}^2$ represents the coordinates of the top-left corner of the poselet bounding box, $w_i$ and $h_i$ gives the bounding box width and height for the detection $i$ respectively and finally $c_i$ is the poselet detection score. As defined before, the activation $i$ of poselet $p$ is fully defined by ${\cal P}(p)_i = \vec a_i$. Notice here that in general a poselet detection ${\cal P}(p)_i$ may include several cell grids of the set $\cal G$ (see Figure \ref{fig:poselet_2D}a).


\begin{figure*}[t]
\begin{center}
  \includegraphics[width=0.9\linewidth]{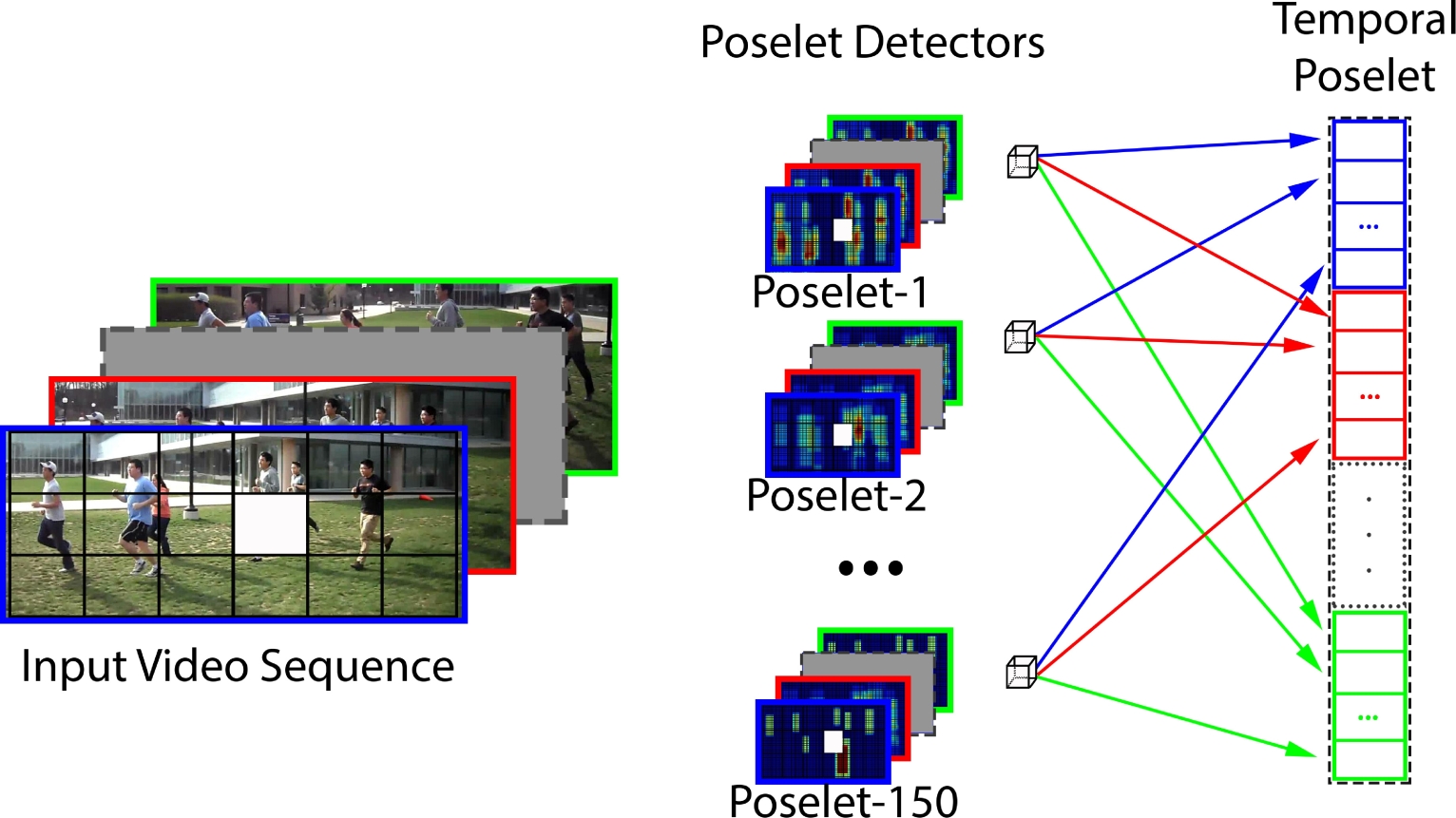}
\end{center}
   \caption{Temporal poselet for a particular video block (shown in white) where colours show poselet activation in different frames.}
\label{fig:short}
\end{figure*}

Now we need to define for each cell grid  $g_k$ of the image which type of poselets $p$ has been activated. To this end, we define a \emph{spatial poselet activation feature} $v(p)_{ik}$ as the ratio between the areas of the intersection and union of the bounding boxes, i.e.:
$$
v(p)_{ik} = \frac{area(\vec a_i \bigcap \vec g_k)}{area(\vec a_i \bigcup \vec g_k)},
$$
where the operators $\bigcap$ and $\bigcup$ define the intersection and union respectively of the image windows $i,k$ whose area is given by the function $area( \: \cdotp \:)$. This ratio indicates if $\vec a_i$, the $i^{th}$ activation of the poselet $p$ includes the image cell $\vec g_k$. By considering all the possible activations $i$ of poselet $p$ we can define the \emph{spatial poselet feature} over all the possible activations as:
$$
v(p)_{k} = \sum_{i=1}^{|{\cal P}(p)|} c_i \: v(p)_{ik}
$$
that provides an indication of the persistence of a particular poselet weighted by the score $c_i$ in a given grid of the image. We finally characterize the overall poselet activation by defining the \emph{spatial poselets vector} $\vec v_k \in \Re^P$ for the image cell $j$ as:
$$
\vec v_{k} = \left[ \begin{array}{cccc}
                  v(1)_{k}, & v(2)_{k}, & \ldots, & v(P)_{k}
                  \end{array} \right]^\top.
$$
This vector has an entry for each poselet type which reflects the degree to which a particular poselet is activated in that area.

In order to consider the dynamics of the scene, we extend the previously defined spatial descriptor with a temporal component. This is achieved by further dividing the video sequence in a set of ($N_h \times N_w \times N_t)$ video blocks  where the length of each block in time is normally $T=10$ frames.
Such frame length was optimised in \cite{le2011learning} and \cite{wang2009evaluation}.
Given each image frame in the $T$-frame video clip, we can define for each cell grid $\vec g_k$ over time $t$ (i.e., the video block) a set of $T$ \emph{spatial poselets vectors} $\vec v_{k,t}$ for $t = 1 \ldots T$. Thus we define the \emph{temporal poselet} descriptor $\vec{TPOS}_{k}$ for the video block $k$ as the concatenation of all the \emph{spatial poselet vectors} such that:
\begin{equation}
\vec{TPOS}_{k} = \left[ \begin{array}{cccc}
                  \vec v_{k,1}^\top, & \vec v_{k,2}^\top, & \ldots, & \vec v_{k,T}^\top
                  \end{array} \right]^\top.
\label{eq:tposelet}
\end{equation}
The vector $\vec{TPOS}_{k} \in \Re^{TP}$  is including the activations of all the poselets in a particular video-block in space and time and captures human motion regularities in the crowd by analyzing the statistics of their poses given the poselet activations. It also embeds the information of the activations of different poselets in time and not only in space (see Figure \ref{fig:short} for a graphical description). Intuitively, we can assume human motion as a sequence of its poses in time, so the sequential activation of poselets can capture the motion of the group of people. If such motion is the regular motion of every single person as well as the motion due to human interactions, temporal poselets activation vectors capture both these two statistics.

This semantic description of the scene helps us toward having a better understanding about the functionality of each region based on the implicit human pose in that region.

\qsection{Video representation and group detection} \label{sec:group_det}
The temporal poselet vectors defined in Eq. (\ref{eq:tposelet}) are the basic building blocks for creating a people-based representation of a video sequence. In practice, the descriptor here defined is a powerful cue for detecting human groupings in unconstrained scenes. However, as the poselet detectors are subject to false positives, some activations might be noisy or not consistent with the human activity. For this reason we define a saliency measure that discards video blocks with few activations.
In practice we define the saliency of a temporal poselet as the sum of the elements of $\vec{TPOS}_{k}$ giving:
$$
s_k = \left\| \vec{TPOS}_{k} \right\|_1.
$$
This measure is an indication of the overall activations of a specific video block and it will be also used for higher-level tasks such as group activity recognition in order to obtain
fewer examples for training and testing. We select this measure because, it exploits the strength of the spatial activations as well as the poselet scores.

Such saliency can also be used directly to provide a powerful cue for people detection in unconstrained scenes. Given $s_k$ for all the cell grids, it is possible to graphically visualise the saliency measure as in Figure \ref{fig:T-poselet}. The resulting map, overlayed over the images in Figure \ref{fig:T-poselet}c, is called Activation Map, and it shows that activations are more predominant where the major density of people are present. It is also interesting to show a comparison between the information implicit in the $\vec{TPOS}$  descriptor with respect to a general purpose descriptor such as STIP \cite{laptev2005space}. For instance, common  spatio-temporal descriptors are responsive to motion, regardless if the cause of the movement is given by a car or a pedestrian as visible in the second row of Figure \ref{fig:T-poselet}d. Also note that in the video clips in the first and third rows of the figure the camera is shaking, and this create a relevant amount of noise for a standard spatio-temporal descriptor. The Activation Map given by the temporal poselets instead only provides the location of the human motion.
For similar reasons, $\vec{TPOS}$ does not manage well single person activity as poselet detector responses are sensitive to the density of the scene. Actually, $\vec{TPOS}$  being a high-dimensional descriptor is very sparse if the scene is scarcely populated. Such sparsity can be found in other high-level descriptors like those presented in Object-bank \cite{li2010object} and Action-bank \cite{SaCoCVPR2012} works.


\subsection{Temporal poselets for group activity recognition} \label{sec:GAR}

Temporal poselets can be used for group activity recognition reaching a higher classification performance than standard spatio-temporal descriptors. In such case, we have to characterize an activity in a video stream given the temporal poselets and relate such information to the set of class activity labels. The standard procedure using temporal descriptors such as SURF3D and HOG3D follow the line of classical bag of words (BoW) approaches. Here we present an adaptation using the temporal poselets that actually achieves a finer representation power than standard methods.

First, we perform the usual split into training and testing sets using a dataset of videos showing different group activities. Then, our approach extracts all the temporal poselet vectors by dividing all the training video sequences in sub-sets of $T$ frames thus obtaining $F$ video clips in total. 
Since the overall number of temporal poselets can be arbitrarily high, we remove temporal poselets with a saliency value lower than a prefixed threshold (i.e. we keep a generic $\vec{TPOS}_{k}$ only if $s_k>s_{th}$). After this initial pruning stage we obtain $N$ temporal poselets that are then used to create a codebook with $k$ clusters (in general $k$ is in the order of the hundred and $N$ about $10^5$/$10^6$). A K-means clustering method using the cosine distance as a similarity measure is adopted to compute this dictionary obtaining for each of the $N$ temporal poselets an assignment to each of the $k$ clusters. This dictionary is consist of the frequent activation patterns of the T-Poselets in video blocks, while each word shows a motion unit corresponding to a particular human motion or human interaction. The intuition is that every complex group motion can be represented using the frequent patterns of human poses who are interacting together.

Now, for the $F$ labelled video clips we compute a histogram representing the frequency of the $k$ temporal poselet words in each clip.
This task at training phase is simplified by the fact that we already know the assignment of each temporal poselet to each of the k-cluster centers.
However, notice that at inference time we will perform a nearest neighbor assignment.
This stage provides a set of $F$ histogram $\vec h_f \in \Re^k$ with $f = 1 \: \ldots \: F$ which represents the frequent correlated Poselets in space and time for the activity classes considered. Representing the crowd motion using bag of the basis temporal poselets, provides more flexibility for representing very complex crowd motions. The histograms for each video clip and their related activity labels are then fed to a SVM classifier. At inference time we create a Bag-of-Word representation for each video clip by assigning each video block to the nearest cluster by using cosine distance. Finally we use trained SVM for classifying the activity of people in the input video clip.

\subsection{Experimental Results} \label{sec:experiments}

In this section we present the datasets used for evaluation, the baseline methods and our results for group detection and collective activity recognition.

\qsection{Dataset description}
We use several released versions of the Collective Activity Dataset (CAD), introduced first in \cite{choi2009they} for evaluating collective activities. The dataset is suitable for our task because of the presence of multiple people activities in the natural unconstrained setting, while most of the classical activity datasets (i.e., CAVIAR, IXMAS, or UIUC), are not adequate for our purpose, since they either consider only the activity of a single person or few number of people~\cite{khamis-eccv2012}. We test our descriptor on second version of the Collective Activity Dataset (CAD2) \cite{choi2011learning} and the recently released third version (CAD3) \cite{choi2012unified}. CAD2 contains $75$ video sequences captured with hand held cameras in realistic conditions including background clutter and mutual occlusions of people. We have activities classified with the following $6$ categories: crossing, waiting, queueing, talking, dancing, and jogging. Instead CAD3 presents 33 video clips with $6$ collective activities: gathering, talking, dismissal, walking together, chasing, queueing. The annotation of the datasets used in our approach are given by people bounding boxes and their trajectories in time. Noticeably, CAD3 has sequences with almost no camera motion with respect to CAD2 where the camera is subject to relevant motion.


\begin{figure*}[t]
\begin{center}
\fbox{\includegraphics[width=1\linewidth]{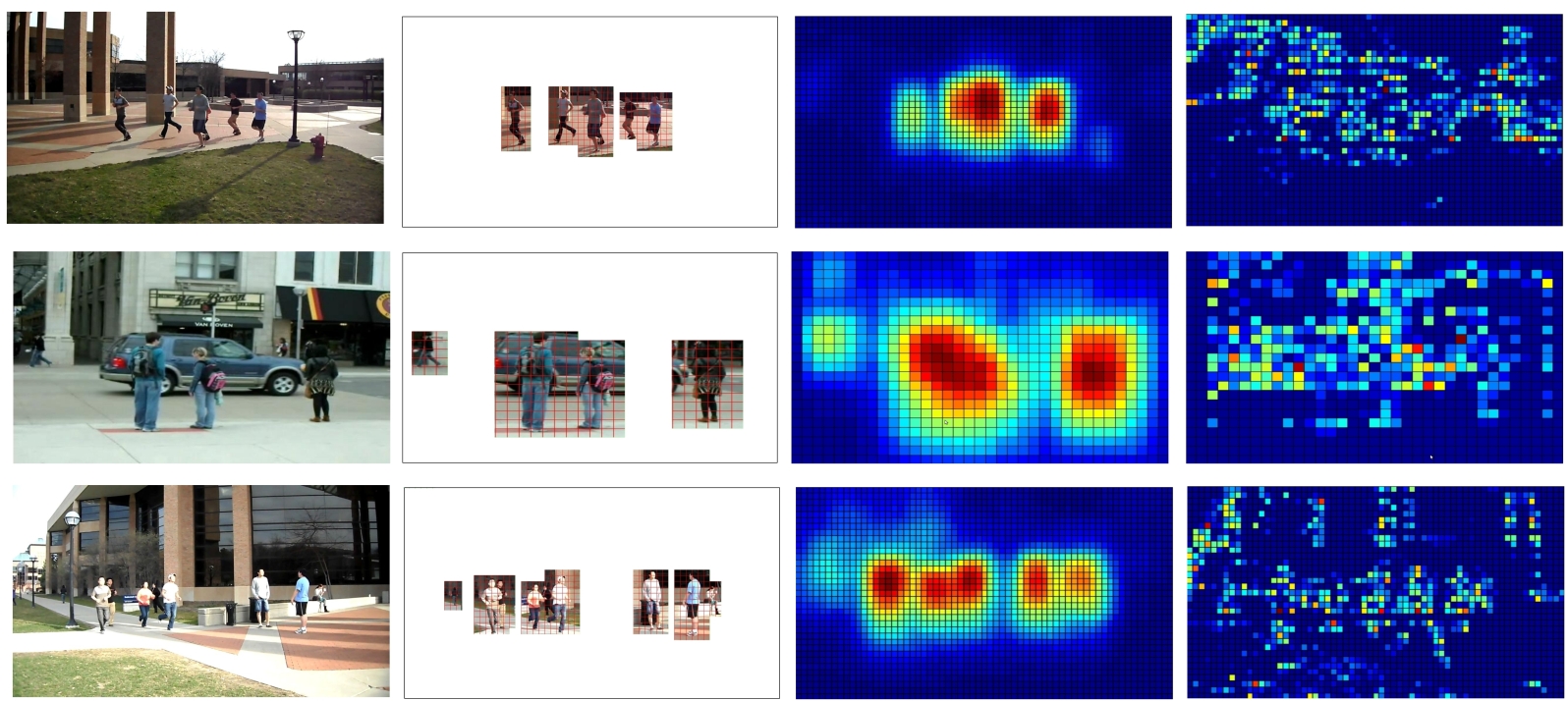}} \\
\vspace{.2cm}
a) \hspace{3cm} b) \hspace{3cm} c) \hspace{3cm} d)
\vspace{-.3cm}
\end{center}
   \caption{Temporal poselet for group detection: (a) represents a sample from $10$-frame different video clips. (b) displays the ground truth positions for the people in the scene (c) shows the color coded activation for the Activation Map using temporal poselets where red represents the stronger activations  (d) shows the color coded activation for the Activation Map using  STIP \cite{laptev2005space}.}
\label{fig:T-poselet}
\end{figure*}

\qsection{Implementation of the baseline methods}
Our final goal is to evaluate the increment of performance given the proposed $\vec{TPOS}$  descriptor. For this reason, we implement two baseline methods with classical low-level temporal descriptors, and we report the increment of performance for collective activity recognition in two datasets (CAD2 and CAD3).
When comparing with feature-based approaches we employ a similar pipeline and protocol as described in \cite{wang2009evaluation}:
we extract local features, we perform vector quantization by using K-means, and we finally classify using ${\chi}^2$ kernel SVMs. As presented in Section \ref{sec:GAR} our method changes the feature extraction stage: we replaced state-of-art descriptors with the new introduced temporal poselet features.
%

\emph{Activity Detection.} In more details, we first run space-time interest points (STIP) \cite{laptev2005space} on each frame for each video sequence of the training videos
Then, we divide each clip to fixed-size video blocks by applying a 3D grid (of size $20 \times 20 \times 10$ pixels) as described in Section \ref{sec:GAR}. Subsequently, we count the frequency of spatio-temporal interest points belonging to each block. This provides a map for each video clip similar to the Activation Map for temporal poselets which is shown in Figure \ref{fig:T-poselet}(d). We consider this approach as our baseline method for group detection.

\emph{Activity Recognition.} Now, to create a baseline for feature based collective activity recognition,
we then select a subset of the video blocks in which the number of spatio-temporal interest points inside them is higher than a prefixed threshold. Notice here that we empirically selected the saliency threshold $s_{th}$=150 in our method and $s_{th}$=120 for the baseline method since these thresholds gave the best performance for each algorithms.\\
At each selected video block we extract a HOG3D descriptor followed by the K-means clustering on around $700,000$ video blocks extracted from training data. We set the size of our codebook to $100$, and then represent every video clip using bag of these visual words.
This parameter K was optimized across the dataset carrying out several experiments with different parameter values finding no big differences in performance. In particular, we initially select K=4000 (same as \cite{wang2009evaluation}), and we reduced to K=1000 gradually down to 100, finding no significant improvement in terms of average accuracy. We finally kept K=100 as the best compromise between accuracy and computational cost.
Finally we trained multi-class SVM with ${\chi}^2$  kernel on a BoW representation of our video clips. In the inference phase, for each input video clip we again create a BoW representation by assigning each video block to the nearest cluster by using cosine distance. Finally, we use trained SVM for classifying the activity of people for every input video clip.
Note here that the BoW and classification procedure has the same set of parameters for both the baseline and the TPOS approach.
%
%

\begin{figure}[t]
\begin{center}
   \includegraphics[width=0.7\linewidth]{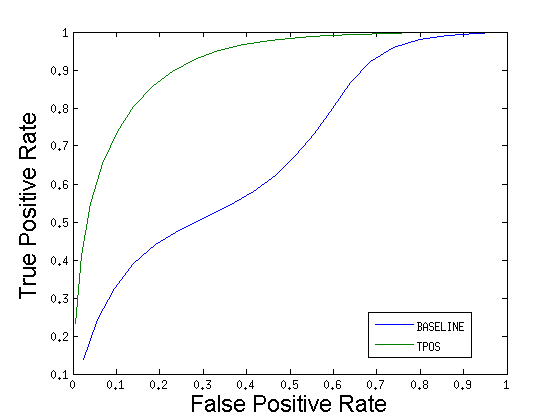}\\
\end{center}
   \caption{Group detection result: $\vec{TPOS}$ (green) vs. Baseline(blue)}
\label{fig:roc}
\end{figure}

\qsection{Results and discussion}
\begin{figure*}[t]
\begin{center}
\includegraphics[width=1\linewidth]{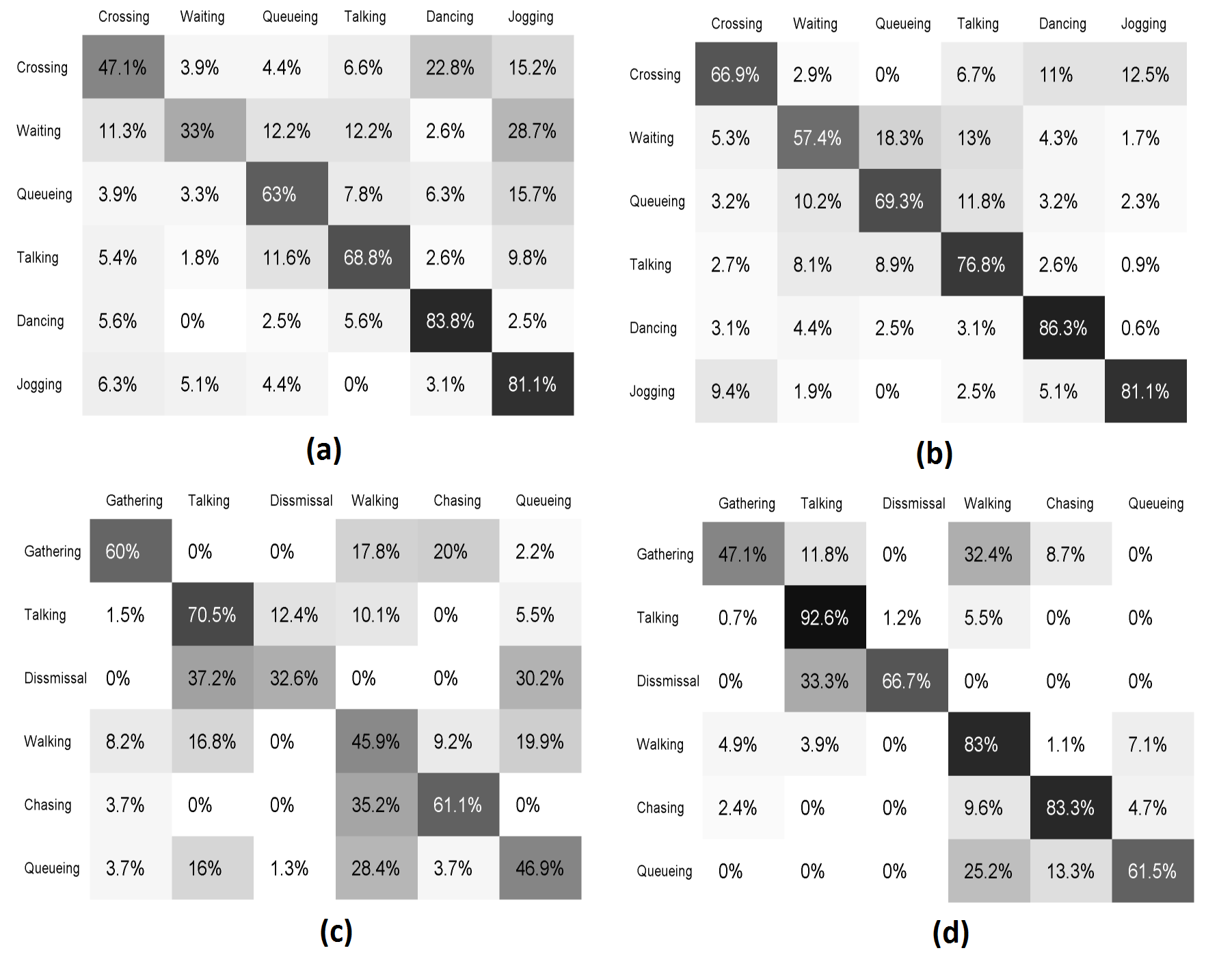} \\
\vspace{.1cm}
\end{center}
   \caption{Confusion Matrix: (a) CAD2- Baseline method. (b) CAD2- $\vec{TPOS}$ method  (c) CAD3- Baseline method (d) CAD3- $\vec{TPOS}$.}
\label{fig:confusion_matrix}
\end{figure*}
The main focus of the experiments is to quantitatively evaluate group activity detection and recognition using CAD2 and CAD3. Nevertheless, 
we also show qualitative results about the robustness of our approach to camera motion and dynamic background (see Figure  \ref{fig:T-poselet}). 
 Collective action detection is also evaluated with quantitative results in terms of ROC curve (see figure \ref{fig:roc}) against the saliency threshold parameter.
%
For quantitative evaluation of group detection, we first generate the ground truth information for multiple people in videos using single people bounding boxes. This ground truth information is provided in terms of human/non-human label for each cell of the 3D grid. More precisely, for each video sequence we generate a 3D binary matrix with the same dimension of the Activation Map. This matrix consists of 1s where people are present in that video block and 0 where there is no person inside (see Figure  \ref{fig:T-poselet}(b)). This ground truth matrix allows us to compare with the baseline method in terms of detecting video blocks as people. Since the density of people in each video sequence is shown as an Activation Map, varying the saliency threshold will result in different video blocks selected as people. We then compute the FP, TP, TN and FN for each threshold value by comparing the assigned label (human or non-human) and the ground truth label for each cell in the grid. This is given by counting the number of same/different labels of the cells in Activation Map in comparison with their corresponding cells in ground truth.
The overall results shows $24.50\%$ improvement in terms of the area under the ROC curve.

Our activity recognition results are shown in Figure \ref{fig:confusion_matrix} in the form of confusion matrices for the $6$-class datasets (CAD2 and CAD3). We also report the performance of our method compared to the baseline approach. In summary, we outperform the baseline in average by $10.10\%$ in CAD2 and $19.50\%$ in CAD3 (see Table \ref{tab:results}).
It is also interesting to notice the higher confusion values among the  waiting, queueing and talking classes. This shows that our descriptor has the tendency to confound human poses related to these three classes because they are very similar in the pose of the people taking part on these activities (i.e. mainly standing). On the other hand, when the intensity of motion activity is higher in activities like in crossing and jogging, we have more accurate results. This can be related to the implicit dynamics extracted by the temporal poselet descriptor.

Moreover, such behaviour does not emerge in the confusion matrix of the baseline approach because it captures the low-level statistics of the pixels motion and gradient. Thus, such descriptor is more sensitive to the appearance of the scene and of the background. In fact, temporal poselets are not likely to be activated in such parts of the image and they are directly related to the human content of the scene. It is also worth to note that the baseline method has worse results in CAD3 because of the lack of overall motion in the sequence since this dataset is a fixed camera scenario. Our method instead, since it is responsive to human parts can anyway provide reasonable results even if people are not moving too much. 

Table \ref{tab:results} also shows the results of our method along with our baseline and the other recently published people-based approaches like \cite{khamis-eccv2012,choi2011learning} for CAD2 and \cite{lan2010beyond,choi2012unified} for CAD3. As mentioned before, the use of higher level features increases the performance of their systems. This is not a fair comparison because we are different in terms of provided supervision. 
Actually, \cite{khamis-eccv2012} employs additional information in training time including bounding boxes of all people, the associated actions and their identities. Instead, the RSTV model \cite{choi2011learning} beside using this information, it also adds additional trajectory information during training, including the location and the pose of every person as well. However, notice that we outperform the feature-level baseline approach significantly, and we slightly outperform the RSTV method when it is not optimized adopting an MRF on top. Obviously, we perform worse than Khamis et al. \cite{khamis-eccv2012} because of the considerations made previously, which we included here for the sake of completeness of evaluation. In CAD3, our results are comparable with \cite{lan2010beyond}(as reported from \cite{choi2012unified}), and there is only $6.8\%$ difference in average accuracy with respect to \cite{choi2012unified}. 
Both these methods are people-based and are based on complex models, the former using contextual information in a latent SVM framework, and the latter by combining belief propagation with a version of branch and bound algorithm equipped with integer programming.


%

\vspace{-.3cm}

\subsection{Conclusions} \label{sec:con_app}

\vspace{-.2cm}

\begin{figure*} 
\begin{center}
\fbox{\includegraphics[width=1\linewidth]{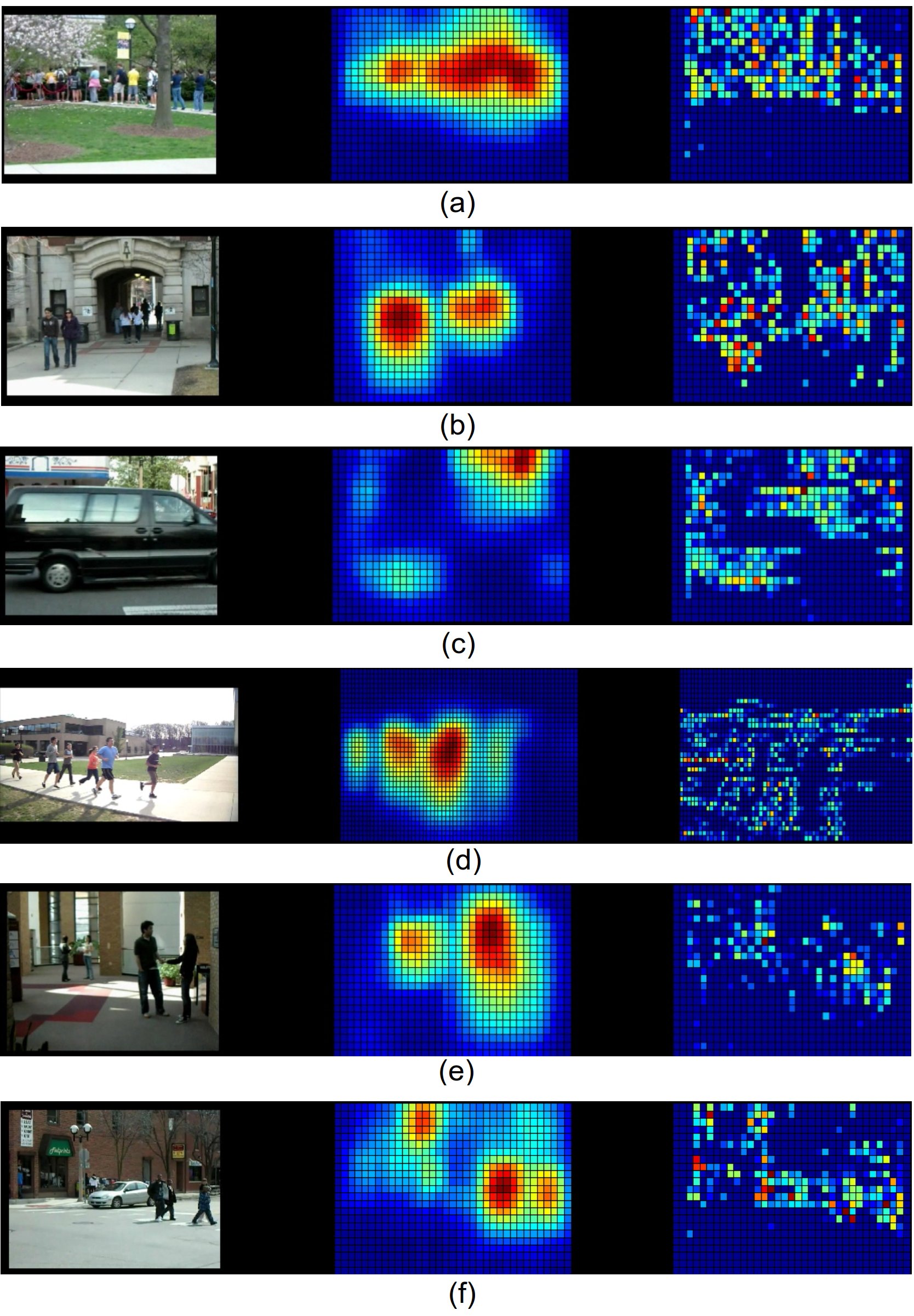}} \\
\end{center}
   \caption{Different scenarios: {\bf(a,b) }shaking camera, {\bf(c) }moving objects, {\bf(d) }change of camera viewpoint, {\bf(e) }scale change and {\bf(f) }a failure case of TPOS.}
\label{fig:T-poselet_all}
\end{figure*}

After focusing on pure motion statistics for video representation in Section \ref{sec:motion}, in this section, we complement it by appearance information in order to build a richer representation. We have evaluated our method by using the temporal poselet descriptor for group detection and activity recognition in the wild. (see more results in Figure \ref{fig:T-poselet_all})
We used the same experimental protocols described in Wang \emph{et.al.}  \cite{wang2009evaluation} and we applied it on two datasets.
The results show significant improvements in detection as well as recognition task in comparison with the baseline methods.
Our representation based on temporal poselets can locally discover the presence of collective activities and recognise the type of action in a single video sequence.
This aspect is very important for tasks in video-surveillance for crowded environments where there is a serious need to localize possibly anomalous activities.
Although there is no well-defined benchmark for the task of multiple activity localization in the wild, however our qualitative results shows that the temporal poselets can also be used for this new task.
Moreover, this approach could also be used for video summarization, by extracting the video clips mostly representative of the peculiar collective activity in a long sequence.
In particular, future work will be directed towards the space-time segmentation of different activities in a single video sequence.



\begin{table}
\begin{center}
\small
\begin{tabular}{c|c|c|c|c|c|c|l}
\cline{2-7}
 & Base & $\vec{TPOS}$ & RSTV & \cite{khamis-eccv2012}  &  \cite{lan2010beyond} & \cite{choi2012unified} \\ \cline{2-7} \hline\hline

\multicolumn{1}{ |c| }{CAD2} & 62.8~\% & 72.9~\% & 71.7~\% &85.7~\% & - & -   \\ \hline
\multicolumn{1}{ |c| }{CAD3} & 52.8~\% & 72.3~\% & - & - & 74.3 & 79.2\%    \\ \hline
\end{tabular}
\vspace{.1cm}
\caption{Average Classification Accuracy.}
\label{tab:results}
\vspace{-.5cm}
\end{center}
\end{table}

\newpage
\thispagestyle{empty}
\mbox{}

\chapter{Concluding Remarks} \label{chap:con}

\section{Summary and Discussion}

In this thesis, we investigated mid-level representation for visual recognition in two ways: image understanding, and video understanding.

In the image understanding part, we first studied the concept of ``subcategory'' for object recognition and investigated how this concept can tackle the dataset bias problem. We, next, proposed a method to discover and train a set of mid-level patches and used this set to form mid-level representation for images. However, we mine and train these patches in visual subcategories instead of object categories. As a result of this modification, we could employ the powers of ``webly-supervision'' to discover and train mid-level discriminative patches. As an another novelty, we also observed that since examples in each subcategory are visually more consistent, deformable patches can be substituted with fixed-position patches in the subcategory level. Our experiments on unsupervised object detection showed that employing richer patch-models can provide significant benefits for mid-level representations.

In the video understanding part, we considered motion-based and appearance-based methods for human behavior understanding. For the case of motion-based mid-level video representation, we showed that employing tracklet as an intermediate representation along with a simple measure to compute the commotion of a given video can be effectively exploited to detect/localize abnormal events in crowded scenarios. The qualitative and quantitative results on the standard dataset show that our approach outperformed the state of the arts in terms of detection speed and performance. For the appearance-based motion representation, we extend the Poselet detector approach by defining a mid-level representation based on Poselet activation patterns over time, and empirically showed that our introduced mid-level representation based on temporal poselets can locally discover the presence of collective activities and recognize the type of action in a single video sequence.

\section{Future Perspectives}

Even if the experiments prove the efficiency of the proposed methods for image and video understanding, there is still room to improved and boost the performance. We extensively discussed about the possible future works related to each visual recognition task in the corresponding chapters (image understanding: Sec. \ref{sec:conc}, Sec. \ref{sec:conclusions_patch} and video understanding: Sec. \ref{sec:con_motion}, Sec. \ref{sec:con_app}), but it is also shortened below:

In the image representation part, both directions of webly-supervised visual recognition and subcategory-based object modeling are interesting to explore. For the former, doing higher-level tasks such as behavior understanding, and for the former following list of research problems are potential feature works:
\begin{itemize}
\item Selecting the number of subcategories adaptively for different object categories.
\item Employing patch-based models for part-based subcategory discovery.
\item Training patch models discriminatively in a cross-subcategory  fashion and use it for subcategory discovery.
\item Inferencing final category detection using multiple unsupervised subcategory detections.
\item Apply subcategory-based models for richer image description utilizing meta-data transfer techniques~\cite{malisiewicz2011ensemble} (e.g. segmentation, visual phrase, 3D layout).
\end{itemize}

In video representation part, for the case of motion-based methods exploring of the proposed approach for the task of action recognition would be a potential direction. For appearance-based methods it would be interesting to employ a richer set of patch models instead of a given set of poselet detectors for motion representation. One example of such patch-based models could be the webly-supervised discriminative patches that discovered/trained similar to our proposed method in section \ref{sec:web_patch}.

\newpage
\thispagestyle{empty}
\mbox{}

\addcontentsline{toc}{chapter}{\scshape{Bibliography}}
{\small
\bibliographystyle{ieee}
\bibliography{bib_short}
}

\newpage
\thispagestyle{empty}
\mbox{}

\end{document}